\pdfoutput=1
\def\PREPRINT{}
\documentclass[10pt]{article}

\ifdefined\PREPRINT
  \usepackage[preprint]{tmlr}
\else
  \usepackage{tmlr}
\fi

\usepackage{amsmath,amsfonts,bm}

\def\eqref#1{equation~\ref{#1}}

\def\1{\bm{1}}

\DeclareMathAlphabet{\mathsfit}{\encodingdefault}{\sfdefault}{m}{sl}
\SetMathAlphabet{\mathsfit}{bold}{\encodingdefault}{\sfdefault}{bx}{n}

\usepackage[T1]{fontenc}
\usepackage{graphicx}
\usepackage{booktabs}
\usepackage{xcolor}
\usepackage{array}
\usepackage{tabularx}
\usepackage{enumitem}
\usepackage{placeins}
\usepackage{url}
\usepackage[hidelinks]{hyperref}
\hypersetup{pdftitle={Large Reported CHB-MIT Adaptation Gains Disappear with Pretrained Encoders and Matched Probes},pdfauthor={},pdfsubject={},pdfkeywords={}}
\graphicspath{{figures/}}

\newcommand{\Pending}[1]{\textcolor{red}{\textbf{[PENDING]}}}%
\newcommand{\AuthorA}{Joseph Sankoorikal Johny}
\newcommand{\AuthorB}{Antony J Sankoorikal}
\newcommand{\CodeURL}{\url{https://github.com/johnyj02/eeg-drift}}

\newcommand{\arm}[1]{\texttt{#1}}
\newcommand{\lparm}{\arm{lp}}
\newcommand{\lastarm}{\arm{last}}
\newcommand{\allarm}{\arm{all}}
\newcommand{\lpr}{\arm{lp\_randinit}}
\newcommand{\lastr}{\arm{last\_randinit}}
\newcommand{\lastnoise}{\arm{last\_noise}}

\newcommand{\relbase}{https://github.com/c3n-group/efficient-eeg-adapt/blob/d056cdf94a6a680917218deae1f7815143ded27c/}
\newcommand{\relcode}[2]{\href{\relbase #1}{\texttt{#2}}}
\newcommand{\ours}{\ifdefined\PREPRINT our repository (\CodeURL)\else the anonymized supplementary material\fi}

\setlist{itemsep=1pt,topsep=2pt,parsep=1pt}

\title{Large Reported CHB-MIT Adaptation Gains Disappear with\\ Pretrained Encoders and Matched Probes}

\ifdefined\PREPRINT
\author{\name \AuthorA \hfill\\
        \addr Independent Researcher, Seattle, WA, USA
        \AND
        \name \AuthorB \hfill\\
        \addr Independent Researcher, Gurugram, India}
\fi

\def\month{MM}
\def\year{YYYY}
\def\openreview{\url{https://openreview.net/forum?id=XXXX}}

\begin{document}

\maketitle

\begin{abstract}
\citet{dani2026efficient} (DL) report that adapting the final encoder block of BIOT and CBraMod with
self-supervised learning (SSL) improves CHB-MIT seizure-detection ROC-AUC by 36.2 and 29.5 points over
linear probing. We investigate these results with their released code and an independent pipeline. The release loads pretrained weights for adaptation only when an otherwise unset configuration key
is supplied, and its CBraMod probe reads the reconstruction head with dropout active. With pretrained
initialisation and a probe on the encoder output in evaluation mode, the fixed-recipe adaptation gain on DL's split is
$-0.6$ points for BIOT (95\% interval $[-3.8, +2.6]$) and $+0.01$ for CBraMod ($[-0.02, +0.04]$) across five
seeds. Across DL's split and five subject rotations on CHB-MIT and Siena Scalp EEG, with equally tuned linear
probes (30 configurations, up to 500 epochs, validation-based early stopping), mean adaptation effects range
from $-1.1$ to $+5.7$ ROC-AUC points over 18 comparisons (related settings and feature variants). Twelve of 18 support equivalence within $\pm3$ points under
dependence-adjusted inference (8 of 12 rotation and 4 of 6 fixed-split comparisons); six are unresolved. After correcting rotation intervals for dependence, no positive effect has a 95\% interval excluding zero. The $\pm3$ margin is an analysis threshold, not a clinically
validated one, and smaller margins support fewer equivalences. Two fixed-probe gains near
$+21$ points shrink under the same tuning, with smaller benefits unresolved. We also audit a CHB-MIT protocol shared by at least 14 papers.
No tested configuration reproduces DL Table~1 as a whole. These findings concern the tested configurations; the encoder-level probe and subsequent
checks were chosen post hoc.
\end{abstract}

\section{Introduction}
\label{sec:intro}

EEG foundation models such as BIOT \citep{yang2023biot}, LaBraM \citep{jiang2024labram} and CBraMod
\citep{wang2025cbramod} are pretrained on large unlabelled EEG corpora and transferred to clinical tasks.
\citet{dani2026efficient} (hereafter DL) propose a low-cost adaptation step
between pretraining and probing: update only the final encoder block with the model's own SSL objective
on unlabelled windows of the target dataset, then train a linear probe, reporting results on CHB-MIT
\citep{shoeb2009thesis,goldberger2000physionet}, TUAB and TUEV. Our primary analysis concerns the reported
adaptation gains on CHB-MIT; Section~\ref{sec:c2c3} examines their data-amount and patient-count claims,
quoted from their abstract, as exploratory analyses. These three claims are our reproduction targets.
\begin{itemize}
  \item \textbf{C1 (headline gain).} SSL adaptation (SA) improves on linear probing (LP). The largest
  gain in DL's results, +36.2 ROC-AUC, is BIOT on CHB-MIT (LP 46.93, SA 83.08, DL Table~1); for CBraMod DL report LP
  54.14 and SA 83.67 (+29.5). DL's abstract states the gain as ``up to 20$\times$ AUCPR'' (BIOT 1.53 to
  31.52). ROC-AUC is our primary metric, matching the validation selection criterion of the
  release's main trainer (Section~\ref{sec:probes}); precision--recall performance is in
  Section~\ref{sec:controls} and Appendix~\ref{app:d29}.
  \item \textbf{C2 (data amount).} ``Under a fixed compute budget, peak performance requires only 20--50\%
  of available unlabeled data.''
  \item \textbf{C3 (patient count).} ``Critically, when total window count is fixed, performance remains
  invariant to patient count.''
\end{itemize}
Our primary purpose is to reproduce a published result. DL followed a CHB-MIT protocol that at least 13
other papers also use (Section~\ref{sec:a3}).

Our audit identifies two implementation choices that materially affect the
comparison. With the released defaults the adaptation step updates a randomly initialised encoder, whereas
linear probing uses the pretrained weights: the SSL step loads the checkpoint only if a configuration key
is set that no shipped file sets. The CBraMod probe also reads reconstruction outputs with dropout active,
and the inherited CHB-MIT protocol tests on two patients (Section~\ref{sec:repro}).
Starting adaptation from the pretrained weights and probing the encoder output in eval mode (the
\emph{encoder-level probe}, Section~\ref{sec:probes}) removes the large gains on DL's CHB-MIT split: the SA
gain is $-0.6$ for BIOT and $+0.01$ for CBraMod (intervals in Table~\ref{tab:original}). Under a revised
protocol with five subject rotations on CHB-MIT and Siena and the fixed probe recipe, gains of the reported size are rejected and the
CBraMod gain is equivalent to zero within $\pm3$. Equal-budget probe tuning reduces the raw-linear BIOT
gains of +1.9 to +9.2 seen under the fixed training recipe. The expanded search covers 18 related comparisons
(Section~\ref{sec:controls}; Tables~\ref{tab:d28} and~\ref{tab:d28b}).

\paragraph{Contributions.} (1)~A reproduction attempt using DL's released code at the audited commit and an
independently implemented pipeline that matches it (Section~\ref{sec:setup}). (2)~Code evidence for two
issues in the release and an account of the inherited protocol, with controlled comparisons of each
setting (Section~\ref{sec:repro}, Table~\ref{tab:artifacts}). (3)~A revised
protocol on CHB-MIT and Siena Scalp EEG
\citep{detti2020siena,detti2020sienadata}, with equally tuned probes and random-feature and noise-input
controls (Section~\ref{sec:main}). (4)~A census of the inherited CHB-MIT protocol
(Section~\ref{sec:a3}, Appendix~\ref{app:census}).

\paragraph{Related work.} EEG foundation-model benchmarks compare many models and adaptation strategies
across many datasets under one harmonised pipeline \citep{xiong2025eegfmbench,kastrati2025eegbench,chen2026eegarena};
they ask which model transfers best, and run every model inside their own pipeline.
We instead audit the implementation underlying one published adaptation method, running the
authors' released code at a fixed commit and changing one setting at a time, and make no ranking of models.
Data-leakage studies show that EEG test sets
sharing subjects with training sets inflate accuracy, including for seizures
\citep{brookshire2024leakage,shafiezadeh2023methodological}, part of a wider pattern in machine-learning-based
science \citep{kapoor2023leakage}. The inherited CHB-MIT protocol is not of this kind: its test patients
are not in training, and its problems are a two-patient test set, a boundary-based label rule and duplicated
test windows (Section~\ref{sec:a3}).
Random-weight and noise-input controls test whether gains require
pretrained representations or EEG-specific adaptation data; such controls are standard in epidemiology
\citep{lipsitch2010negative} and have been proposed for clinical EEG foundation-model benchmarks
\citep{zare2026negative}. We apply both: a probe on randomly initialised
weights next to every adapted arm \citep{saxe2011random}, and SSL adaptation on Gaussian noise, which alone
can shape useful features \citep{baradad2021noise}. Variance that a single split hides
\citep{bouthillier2021accounting} motivates the subject rotations and the
corrected intervals \citep{nadeau2003inference}; event-level scoring follows SzCORE
\citep{dan2025szcore}.

\section{Setup}
\label{sec:setup}

\paragraph{DL's released code and our reimplementation.} We audited DL's repository
\texttt{c3n-group/efficient-eeg-adapt} at commit \texttt{d056cdf}, hereafter \emph{the release}. We ran it
with command-line overrides and, for the probe ablation in Section~\ref{sec:a2}, two one-line patches. We
also reimplemented it without copying any of its files (licences: Appendix~\ref{app:checklist}; BIOT and
CBraMod model code is vendored from their own repositories). A differential test runs both from
the same start state on the same windows, for both models, heads and initialisations. They agree in all 8
cells to within $10^{-5}$ on losses, adapted and probe weights, logits and metrics computed from the same
score representation (largest difference $8.5\times10^{-8}$ over 33 assertions). Appendix~\ref{app:e21}
describes the test and a scoring difference caused by float32 sigmoid saturation.

\paragraph{Data.} \textbf{CHB-MIT} 1.0.0 (\citealp{guttag2010chbmit}): 686 EDF files, 24 cases.
The shared preprocessing (the release's copy of BIOT's \texttt{process2.py},
Section~\ref{sec:a3}), run unmodified, gives 326,993 evaluation windows with
252 positives among 16,964 test windows (1.485\%) and 81,993 SSL windows, agreeing with our stores in
window identifiers, split assignments and labels.
\textbf{Siena Scalp EEG} (CC BY 4.0): 14 subjects, 47 seizures, mapped to BIOT's 16 bipolar
pairs. The primary path is the CHB-MIT one; a
secondary path filters at 0.3--75~Hz with a 50~Hz notch (``Siena filtered'', fixed before any Siena
run). Data checks are in Appendix~\ref{app:datachecks}.

\paragraph{Models, SSL protocol and arms.} BIOT uses \texttt{EEG-SHHS+PREST-18-channels.ckpt} and CBraMod the
Hugging Face \texttt{pretrained\_weights.pth} (hashes: Appendix~\ref{app:checklist}); both models receive
exactly the windows of the shared preprocessing. BIOT's checkpoint holds no SSL prediction head,
so that head starts from random weights in both arms (Appendix~\ref{app:e21}). The parameters the released
trainer unfreezes match DL's rounded counts (Table~\ref{tab:repro-summary}, Appendix~\ref{app:released}).
SSL adaptation follows the released path for DL Table~1: batch 256 (DL's text gives 128;
Appendix~\ref{app:questions}), 3 epochs over
$N=4{,}174$ SSL windows, AdamW (lr $3\times10^{-5}$, weight decay $5\times10^{-2}$), cosine decay to
$10^{-6}$, clipping at 1.0, keeping the epoch with the lowest training loss (51 optimizer steps). No rule
we tested yields DL's $N_\mathrm{total}=4{,}174$ from the 81,993 windows, so we use a
seeded uniform subsample and test other draws (Appendix~\ref{app:robust}).
Table~\ref{tab:arms} defines the six arms: \lparm, \lastarm{} and \allarm{} from the pretrained weights,
\lastr{} and \lpr{} from a random initialisation (the shipped path), and \lastnoise, which adapts the
pretrained final block on Gaussian noise matched to the SSL sample's per-channel mean and standard
deviation.
DL describe their SA rows as \lastarm; as shipped, the released code computes \lastr{} (Section~\ref{sec:a1}).

\paragraph{Probes.}
\label{sec:probes}
All probes train a head on a frozen backbone (the \emph{fixed recipe}: AdamW, lr $4\times10^{-4}$, 10
epochs, batch 256 unless stated); only the fine-tuning reference trains the encoder. The head is the
raw head \texttt{linear} (as DL describe), the shipped MLP head \texttt{mlp\_fullft}
(\relcode{src/conf/config.yaml\#L30}{config.yaml:30}: a 64.4M-parameter MLP for CBraMod) or the
standardized head \texttt{linear\_z}; Appendix~\ref{app:glossary} defines all three.
The \emph{tuned probes} train linear heads on cached frozen features, raw and train-standardized,
using the same validation-selected search for every arm: 30 learning-rate/weight-decay configurations,
up to 500 epochs and early stopping (Appendix~\ref{app:d28}). Selection diagnostics and the
remaining search limits are reported with the results (Section~\ref{sec:controls}). The original
16-configuration, 200-epoch search
is retained as a sensitivity analysis in the supplement.

\paragraph{Representation source and backbone mode.} Test metrics are read at the epoch with the best
validation ROC-AUC, as in the released main trainer for binary tasks
(\relcode{src/trainer.py\#L578}{trainer.py:578}); the released data-amount and patient-count trainers select
by balanced accuracy (\relcode{src/ablation_diversityN/trainer_diversityN.py\#L502}{trainer\_diversityN.py:502};
effect: Appendix~\ref{app:e25}).
The \emph{released-code probe} (the probe of DL's release) reads the encoder output for BIOT and
\texttt{proj\_out} for CBraMod, with the backbone in train mode (dropout on). The \emph{encoder-level
probe} reads the encoder output for both models with the backbone in eval mode. DL describe
a linear classifier on the encoder's representations (DL Section~2.1); the eval-mode backbone and, for
CBraMod, the encoder representation source are our post hoc choice. The encoder-level probe is used for
every result in Sections~\ref{sec:main} and~\ref{sec:c2c3} except released-code-probe values cited for
comparison. The probe searches, logistic-regression check, larger-dose reassessment and fixed-compute
controls of C2/C3 are post hoc analyses, with each protocol specified before its own runs
(dates and commits: Appendix~\ref{app:stats}).

\paragraph{Statistical reporting.}
\label{sec:rules}
Metrics are $\times100$: ROC-AUC throughout; balanced accuracy (bal), average precision (AP) and PR-AUC
(trapezoidal, the released definition) where noted. The interval, equivalence and power analyses were
retrospective. We report paired differences between arms, for
example the SA gain $\lastarm-\lparm$, in ROC-AUC points with 95\% intervals. For the subject rotations we
average seeds within each rotation and compute a $t$-interval over the five rotation means (df 4). These
intervals reflect variation across held-out subject groups; they do not treat individual windows or seeds
as independent patient samples. Because the training folds of different rotations overlap, we also report
Nadeau--Bengio corrected intervals \citep{nadeau2003inference}, and name the method wherever the two differ
(every interval: Appendix~\ref{app:stats}). On DL's split the five seeds share one fixed split, so
intervals there describe seed variation conditional on the same patients; arm levels are
mean~$\pm$~SEM (for 2-seed cells, $\pm$ is the half-range). Equivalence within $\pm3$ ROC-AUC points is
assessed by TOST at $\alpha=0.05$: the corresponding 90\% interval must lie entirely within the margin.
The displayed 95\% intervals describe effect uncertainty; ``equivalent within $\pm3$'' is used only where TOST rejects. For AP
and trapezoidal PR-AUC we report effects and intervals without applying the ROC-AUC margin. Bootstrap
intervals under-cover with five units and are descriptive (Figure~\ref{fig:forest},
Appendix~\ref{app:stats}).
We report a prespecified \emph{consistency criterion} descriptively: a paired
difference $A-B$ over five units meets it when the absolute mean exceeds 3 points and at least 4 of the 5
units have the same sign as the mean, the form our reading scripts implement. The tuned-probe protocol adds
a prespecified gain criterion: the 95\% $t$-interval excludes zero and the mean exceeds three points in the same
direction; a comparison meeting neither that criterion nor the equivalence test is ``not resolved''.
Meeting either criterion is not evidence that the population effect exceeds three points; our conclusions
rest on the intervals. The choice to interpret this criterion descriptively was post hoc
(Appendix~\ref{app:hygiene}). The criterion reaches 80\% power only for differences of about 3
to 6 points or more, and its false-positive rate reaches 0.352 for the most variable BIOT contrasts
(Appendix~\ref{app:stats}). Table~\ref{tab:vocab} maps every classification to its analysis-script output
and to the appendix tables' letters and timing tags.
Jobs ran on one NVIDIA A10G (24~GB) each, with the exceptions listed in Appendix~\ref{app:checklist};
the project cost about \$4,856 at cloud list prices (Appendix~\ref{app:compute}).

\section{Released-code evaluation}
\label{sec:repro}
\label{sec:artifacts}

Table~\ref{tab:original} sets DL's numbers next to the released code and our runs.

\begin{table}[t]
\caption{DL Table~1 (CHB-MIT, DL's split) against the released code and our runs; ``DL Table~$n$'' means
table $n$ of \citet{dani2026efficient}. ROC-AUC $\times100$, mean~$\pm$~SEM over 5 seeds (seed uncertainty
on one fixed split); SA$-$LP is the difference of the two means, except in the last column, which gives the
mean paired gain with its 95\% $t$-interval. Released code: shipped defaults (MLP head,
\texttt{foundation\_dir} unset), best / last epoch. Ours: linear head (as DL describe), best-validation
epoch; the random start is the released SSL path. Probes: Section~\ref{sec:probes}.}
\label{tab:original}
\centering\footnotesize
\setlength{\tabcolsep}{3pt}
\begin{tabular}{llcccc}
\toprule
 & & & \multicolumn{2}{c}{ours, released-code probe} & ours, encoder-level probe\\
\cmidrule(lr){4-5}\cmidrule(lr){6-6}
cell & DL & released code (best / last) & random start & pretrained start & pretrained start\\
\midrule
BIOT LP & 46.93$\pm$0.30 & 52.32$\pm$0.10 / 52.32$\pm$0.10 & \multicolumn{2}{c}{52.05$\pm$0.23} & 55.0$\pm$0.4\\
BIOT SA & 83.08$\pm$1.89 & 86.48$\pm$1.89 / 88.58$\pm$0.21 & 90.15$\pm$0.15 & 52.55$\pm$1.13 & 54.4$\pm$1.1\\
BIOT SA$-$LP & +36.2 & +34.2 / +36.3 & +38.1 & +0.5 & $-$0.6 [$-$3.8, +2.6]\\
\midrule
CBraMod LP & 54.14$\pm$1.39 & 84.18$\pm$0.96 / 82.41$\pm$0.09 & \multicolumn{2}{c}{65.36$\pm$0.52} & 92.00$\pm$0.04\\
CBraMod SA & 83.67$\pm$1.36 & 89.97$\pm$0.79 / 88.36$\pm$0.52 & 86.83$\pm$0.48 & 74.04$\pm$0.33 & 92.02$\pm$0.05\\
CBraMod SA$-$LP & +29.5 & +5.8 / +5.9 & +21.5 & +8.7 & +0.01 [$-$0.02, +0.04]\\
\bottomrule
\end{tabular}
\end{table}

\paragraph{Documented commands.}
\label{sec:e4}
Run verbatim on CPU, all documented entry points failed before any SSL step. The main and layer-ablation
commands failed when the trainer accessed the undefined configuration key
\texttt{cfg.method.spt\_anchor\_l2sp\_lambda} (\relcode{src/trainer.py\#L154}{trainer.py:154}), which no
shipped configuration defines; the two diversity commands encountered configuration errors earlier
(Appendix~\ref{app:e2}). Setting \texttt{+method.spt\_anchor\_l2sp\_lambda=0} (our choice; other values:
Appendix~\ref{app:e31}) allowed every corrected command to run and unfreeze DL's parameter counts.

\paragraph{The released code on DL Table~1 (checkpoint-loading audit).} We ran the four CHB-MIT cells of
DL Table~1 with the released defaults, \texttt{foundation\_dir} unset as shipped, five seeds, plus the LP cell with \texttt{model.use\_pretrained\_weights=false} as a
random-features control, reading each at the epoch the trainer keeps (\emph{best}) and at the final epoch
(\emph{last}; Appendix~\ref{app:e4}). By a rule fixed in advance, the released code matches our
reimplementation within a tolerance (the larger of 2 ROC points and twice the pooled SEM) in all 8 cells, and the released SA lies within 2.05 points of the random-features LP for
both models, as expected if the released SSL step adapts a randomly initialised encoder
(Section~\ref{sec:a1}). Rerunning the same seed changed bal / PR / ROC by 3.19 / 0.26 / 0.71, so the
released code is not bit-reproducible.

\paragraph{No configuration we ran reproduces DL Table~1.} In our reimplementation (a 2-seed screen and a
5-seed confirmation, Appendix~\ref{app:replication}), no configuration matches all four cells: random-init
BIOT SA lies more than 3 DL-SEM above DL's value and pretrained BIOT SA far below it, while random-init
CBraMod SA is consistent with DL's. A factorial in the released code over
head, batch size, \texttt{foundation\_dir} set or unset, and SSL store
(Appendix~\ref{app:e5}) brings neither LP cell within the comparison tolerance. DL's BIOT SA falls within
it only with the defaults, at the best epoch, and only once the tolerance is enlarged for same-seed
variability in balanced accuracy; DL's CBraMod SA falls within it at both epoch rules, but only with
non-default configurations (Table~\ref{tab:repro-summary}).
The size of the BIOT gain matches descriptively: with the shipped head, random-init SA minus pretrained LP
is $+36.0$ against DL's +36.2, although both of our cells sit 5.4 to 5.5 points above DL's
(Table~\ref{tab:wave1}).

\subsection{Issue 1: the released SSL step adapts a randomly initialised encoder by default}
\label{sec:a1}

\paragraph{Code evidence (release at \texttt{d056cdf}).} The BIOT and CBraMod SSL models
(\relcode{src/trainer.py\#L364-L368}{trainer.py:364--368}, \relcode{src/trainer.py\#L407-L411}{407--411}) load
the checkpoint only if \texttt{cfg.model.foundation\_dir} is set (default: empty string). The string
\texttt{foundation\_dir} occurs in no configuration file, README or launcher of the release, only in the
trainers that read it. After adaptation,
\relcode{src/trainer.py\#L306}{trainer.py:306} (BIOT) and \relcode{src/trainer.py\#L309}{309} (CBraMod) copy
the SSL model's encoder into the probe's backbone, overwriting the pretrained weights that
\relcode{src/model_factory.py\#L73-L75}{model\_factory.py:73--75} and
\relcode{src/model_factory.py\#L26-L28}{26--28} had loaded. So, with \texttt{foundation\_dir} unset as in every shipped file, the LP arm probes the
pretrained encoder and
the SA arm a random encoder whose last block was SSL-adapted. The layer-ablation trainer behind DL
Table~2 calls the same parent method
(\relcode{src/ablation_layer/trainer_layer.py\#L40}{trainer\_layer.py:40}). The same path, with no dataset
branch, runs TUAB and TUEV (\texttt{tuab.yaml} and \texttt{tuev.yaml} do not set \texttt{foundation\_dir}).
Our arm \lastr{} runs exactly this path; it equals the released modules (differential test) and runs
(Section~\ref{sec:e4}).

\paragraph{Checkpoint validation.} Pretrained BIOT \lastarm{} with the released-code
probe ranged from 51.99~$\pm$~1.30 to 64.48~$\pm$~0.60 across our replication runs (5 seeds each), 18.6 to
31.1 points below DL's SA (83.08). As a check that the loaded checkpoint supports the expected downstream
behaviour, fine-tuning it on the labels (one 10-epoch recipe, post hoc;
Appendix~\ref{app:ft}) gives AUROC 0.8753 on DL's split, against 0.8752 reported by
\citet{yang2023biot} with a different recipe (descriptive).

\subsection{Issue 2: the released CBraMod probe reads the reconstruction head with dropout on}
\label{sec:a2}

\paragraph{Code evidence.} Pretrained CBraMod ends in \texttt{proj\_out}, a linear layer that maps each
encoder token back to an input patch for the masked-reconstruction objective. The official CBraMod
downstream model for CHB-MIT replaces it with \texttt{nn.Identity()}
(\texttt{CBraMod/models/model\_for\_chb.py:18}) and probes the encoder output. In DL's release the line is
commented out (\texttt{\#backbone.proj\_out = nn.Identity()},
\relcode{src/model_factory.py\#L29}{model\_factory.py:29}), so the probe reads the reconstruction head;
and the probe loop calls \texttt{self.model.train()} at the start of every epoch
(\relcode{src/trainer.py\#L597}{trainer.py:597}), so the frozen backbone runs with dropout on. Neither line
depends on the dataset. With the line
active, the released SA run fails at \relcode{src/trainer.py\#L309}{trainer.py:309}, so DL's
CBraMod SA cells, if computed with the release, read \texttt{proj\_out} unless two of its lines were
changed (Appendix~\ref{app:e2}).
The pretrained \texttt{proj\_out} is nearly singular, and train-mode dropout introduces variation 2.3
(encoder) to 2.8 (\texttt{proj\_out}) times the pretrained output's variation across windows, against 1.1
times for a random initialisation (Appendix~\ref{app:cbracheck}); adapting the last block also re-trains
\texttt{proj\_out}, the layer this probe reads.

\paragraph{Probe ablation.} We evaluated all four combinations of representation source (\texttt{proj\_out}
or encoder output) and backbone mode (training or evaluation) in our pipeline and, with the two one-line
patches, in the released code
(Table~\ref{tab:tapmode}). On LP the two pipelines agree in every cell (largest gap 1.01). Neither factor
alone explains the LP level: the representation source moves it by +15.24 and the mode by +11.40, with a
strong interaction (Figure~\ref{fig:ablation} gives the SA gains).

\subsection{The inherited CHB-MIT protocol}
\label{sec:a3}

\paragraph{Code evidence.} BIOT's preprocessing script \texttt{datasets/CHB-MIT/process2.py} (BIOT at
\texttt{d138e32}) sets three things; the same lines appear in CBraMod's copy
(\texttt{preprocessing/CHB-MIT/process2.py} at \texttt{b9e9610}) and in the release
(\relcode{data_preprocessing/CHB-MIT/process2.py}{data\_preprocessing/CHB-MIT/process2.py}).
\begin{itemize}
  \item \textbf{Split} (\relcode{data_preprocessing/CHB-MIT/process2.py\#L17-L18}{lines 17--18}): test chb23
  and chb24, validation chb21 (the same subject as the training case chb01) and chb22.
  \item \textbf{Label rule} (\relcode{data_preprocessing/CHB-MIT/process2.py\#L116-L117}{lines 116--117}): a
  10~s grid window is positive only if a seizure start or end index lies strictly inside it, so a window
  that is ictal throughout is negative (example: Figure~\ref{fig:timeline}, Appendix~\ref{app:figs}). Read
  as an onset-and-offset detector, which the rule may be meant to be, it defines a different task from
  seizure detection.
  \item \textbf{Duplicate seizure windows}
  (\relcode{data_preprocessing/CHB-MIT/process2.py\#L131-L135}{lines 131--135}): for every seizure, extra
  10~s windows start every 5~s from 1~s before onset to 1~s after offset, all positive, and are written to
  the patient's split (lines 155--160), test included.
\end{itemize}
The shared preprocessing also drops chb12, chb13 and chb17; the release filters SSL windows and divides
them by 700, while evaluation windows are unfiltered and divided by 100, although DL's text describes 0.3--75~Hz
band-pass and 60~Hz notch filtering. Repeating evaluation with the stated filtering did not change the
descriptive classifications (Appendix~\ref{app:e25}).

\paragraph{Effects on the test set.} Of the 252 test positives on DL's split,
206 (82\%) are duplicate windows; of the 114 truly ictal test grid windows (overlap rule), the inherited
start/end rule labels 70 (61\%) negative, and 2 of its 46 grid positives end exactly at
seizure onset and hold no ictal sample (\relcode{data_preprocessing/CHB-MIT/process1.py\#L50}{process1.py:50} uses
seconds~$\times$~256~$-$~1).
The two test patients differ strongly (frozen pretrained BIOT 41.2 on chb23 and 64.0 on
chb24, random BIOT features 97.0 and 86.1; encoder-level probe, linear head, post hoc), and BIOT's
random-initialisation advantage of +35.8 on DL's split disappears across five rotations ($-2.8$ at one seed
per rotation with the start/end labels; +0.2 with overlap labels and three seeds). The label rule and the
duplicates do not account for it:
they roughly cancel in aggregate PR-AUC (Appendix~\ref{app:rescore}), and training the probe on overlap
labels leaves the advantage at +30.9 (Appendix~\ref{app:e18}).

\paragraph{Protocol adoption.} A citation search and a second verification pass (Appendix~\ref{app:census},
which defines the evidence classes) found, of 19 candidate papers, 14 that use it, BIOT, CBraMod and DL
among them: 8 by code and 6 by text only; 5 more are probable. We claim nothing about their numbers beyond
their use of the protocol.

\section{Pretrained weights and an encoder-level probe}
\label{sec:main}

\paragraph{The revised protocol.} A 10~s window is positive if and only if it overlaps a seizure interval by
at least one sample, with true [onset, offset) bounds; validation and test use grid windows only, and train
keeps grid and duplicate windows, relabelled. Splits are five subject rotations; on
CHB-MIT chb01 and chb21 are always in the same fold, with 4--5 test subjects and 21--41 seizures per test
fold, and the SSL pool of a rotation excludes its test subjects. We retain the shared evaluation
preprocessing, the released input scaling (except ``Siena filtered'') and the SSL protocol of
Section~\ref{sec:setup}, with probe batch 256. Evaluation uses the encoder-level probe with the linear
head, and each rotation uses three model seeds (Appendix~\ref{app:checklist}).

The design was fixed before any revised-protocol run except the probe and the seed count, both settled after
one-seed runs had been read but before their own runs. For CBraMod the $S_r$ run is an earlier run reused,
a post hoc choice; all 12 fixed-recipe consistency classifications remain unchanged without it
(Appendix~\ref{app:hygiene}).

\subsection{DL's split}
\label{sec:p1e}

\begin{figure}[t]
\centering
\IfFileExists{figures/fig_ablation.pdf}{\includegraphics[width=\linewidth,height=0.3\textheight,keepaspectratio]{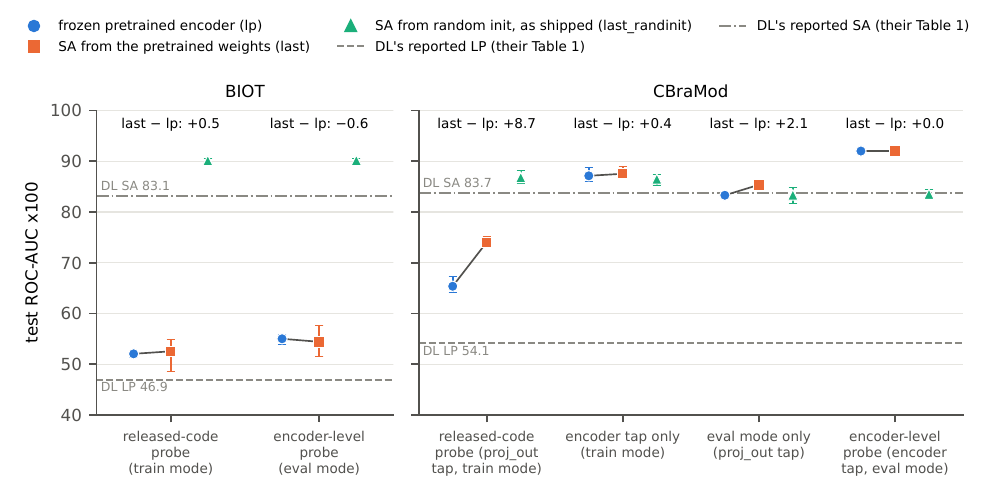}}{\fbox{\Pending{fig-ablation}}}
\caption{Initialisation $\times$ probe-setting ablation on DL's split (CHB-MIT, start/end labels, DL's SSL
protocol, linear head, 5 seeds; ROC-AUC $\times100$): frozen pretrained encoder (\lparm), adaptation from
the pretrained weights (\lastarm) and from random weights (\lastr, the shipped SSL path). Marker: mean over
seeds; whisker: range (seed variation on one fixed split). The number above each group is the SA gain
$\lastarm-\lparm$; dashed and dash-dotted lines: DL's reported LP and SA. Probe settings: the released-code
probe, for CBraMod each of its two changes alone, and the encoder-level probe. Values:
Tables~\ref{tab:original}, \ref{tab:tapmode} and~\ref{tab:p1e}.}
\label{fig:ablation}
\end{figure}

We keep DL's split, the start/end labels, the released input scaling and DL's SSL protocol, start adaptation
from the pretrained weights (\lastarm, unlike the shipped configurations) and vary the probe
(Figure~\ref{fig:ablation}). These settings do not recover the reported gains: with the encoder-level probe
the SA gain is $-0.6$ for BIOT (shipped MLP head $-1.2$, standardized probe
$+0.2$) and $+0.01$ for CBraMod (intervals: Tables~\ref{tab:original} and~\ref{tab:p1e}). The CBraMod gain
is equivalent to zero within $\pm3$ (TOST $p<0.001$); for BIOT equivalence is not shown (TOST $p=0.053$;
with tuned raw-feature probes it is, Section~\ref{sec:controls}).
Initialisation and probe choice affect the two models differently. For BIOT, whose released-code probe
already reads the encoder, the pretrained start alone removes the gain (shipped head: \lastarm{} 51.99
against \lparm{} 52.44). For CBraMod with the linear head the pretrained start leaves
$+8.67$ with the released-code probe (at $N=4{,}174$; +18 to +22 with the whole SSL store,
Table~\ref{tab:e5}); changing only the representation source or only the backbone mode
leaves +0.40 or +2.06, and changing both gives the +0.01 above.
Absolute performance helps interpret these differences. Frozen CBraMod read by the encoder-level probe
scores 92.00, above DL's \emph{adapted} 83.67, while frozen BIOT scores 55.0 and a probe on random BIOT
weights 90.1: on these two test patients random features score 35 points above pretrained BIOT (12.2 with
tuned probes; Section~\ref{sec:a3}; curves in Figure~\ref{fig:rocpr}).

\subsection{Results under the revised protocol}
\label{sec:d24a}

\begin{figure}[t]
\centering
\IfFileExists{figures/fig_effects.pdf}{\includegraphics[width=\linewidth,height=0.36\textheight,keepaspectratio]{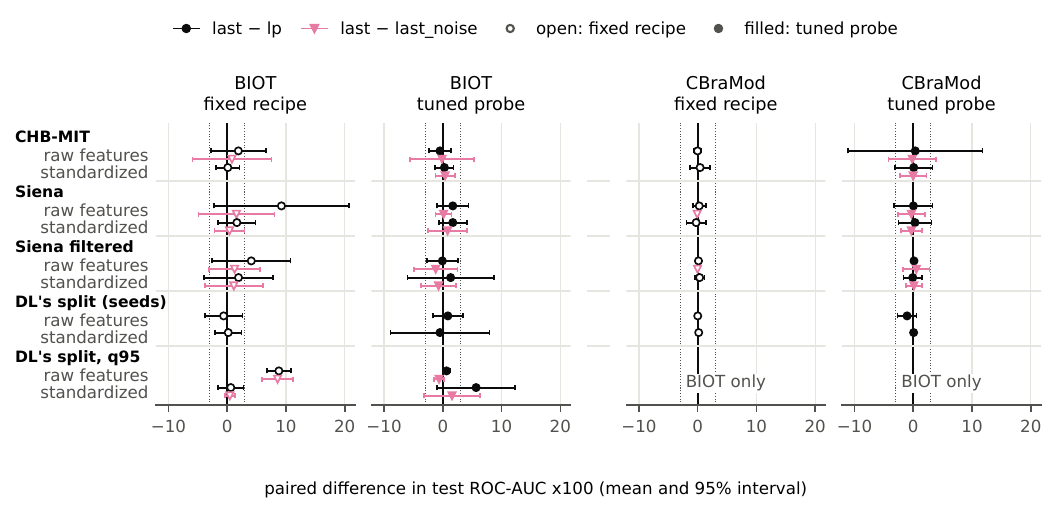}}{\fbox{\Pending{fig-effects}}}
\caption{Paired adaptation effects on the same frozen representations: fixed probe recipe (open markers)
against tuned probes (filled), ROC-AUC $\times100$, positive = the first arm higher. Per model, two panels
share a scale for $\lastarm-\lparm$ (circle, the SA gain) and $\lastarm-\lastnoise$ (triangle, the noise
control). Marker and whisker: mean and 95\% interval, Nadeau--Bengio corrected over five subject rotations
or ordinary $t$ over five seeds on DL's fixed split (seed variation conditional on the same patients;
Section~\ref{sec:rules}). Rows: raw and standardized features per setting. Zero: thick line; $\pm3$:
dotted; a missing marker means the arm was not run. Absolute levels, including random-feature controls:
Figure~\ref{fig:absolute}. All effects and intervals:
Tables~\ref{tab:d28}, \ref{tab:d28b} and~\ref{tab:d24a}.}
\label{fig:effects}
\end{figure}

\paragraph{SA against LP.} For CBraMod the SA gain (Figure~\ref{fig:effects}) is $-0.0$, $+0.3$ and $+0.1$
on CHB-MIT, Siena and filtered Siena, equivalent to zero within $\pm3$ in all three (TOST $p<0.001$). For
BIOT it is $+1.9$, $+9.2$ and $+4.1$; the $t$-intervals exclude zero on the two Siena paths, but with the
Nadeau--Bengio correction all three include zero. The BIOT CHB-MIT gain is neither shown to
differ from zero nor shown equivalent to zero within $\pm3$ (TOST $p=0.175$). Against DL's reported gains
(Section~\ref{sec:intro}), one-sided $t$-tests give $p\le2.9\times10^{-6}$ on CHB-MIT for both models and
both protocols, and $p\le2.1\times10^{-4}$ on Siena. The descriptive consistency criterion
is met only for BIOT on the two Siena paths (every interval: Table~\ref{tab:intervals}).

\paragraph{Pretrained against random initialisation and random features.} Pretrained CBraMod scores above
both a randomly initialised encoder and a probe on random features in the mean in all three settings
($\lparm-\lpr$ on CHB-MIT, $+5.5$, is representative). Its $t$-intervals exclude zero in 5 of
these 6 contrasts but the Nadeau--Bengio intervals in only 2, so the advantage is sensitive to the
correction for overlapping training folds (every effect and interval:
Table~\ref{tab:intervals}).
For BIOT both contrasts change sign between rotations, so they depend on which subjects are held out. In
every setting the shipped random-initialisation path stays within 1.1 points of a probe on random features:
at this budget, adapting the last block of a random encoder barely changes what a linear probe reads.

\subsection{Controls}
\label{sec:controls}
Details and one-seed controls: Appendix~\ref{app:robust}; every control: Table~\ref{tab:robust}.

\paragraph{Noise control.} The \lastnoise{} arm (Section~\ref{sec:setup}) was fixed in advance for the main
runs. For BIOT, $\lastarm-\lastnoise$ is $+0.8$, $+1.6$ and $+1.3$: no difference between EEG and noise
adaptation is demonstrated, and equivalence is not established either (TOST within $\pm3$: $p$ from 0.069
to 0.186); the two CBraMod cells are equivalent to zero ($p<10^{-4}$). On Siena, adaptation on noise
gives +7.6 of the +9.2 BIOT gain ($\lastnoise-\lparm$). Post hoc, we added noise controls for the remaining
BIOT configurations meeting the descriptive consistency criterion. Among the 20 comparisons specified in the decision log, 18
did not demonstrate an EEG-versus-noise difference under the consistency criterion; the other two are the
q95 cases discussed below (Appendix~\ref{app:noise}). Of the 18 one-seed dose comparisons, two met the
EEG-versus-noise criterion, each driven by one rotation; these readings are descriptive.

\paragraph{Standardized probe.} Random BIOT features are 3.5 to 11 times larger than pretrained ones
(Appendix~\ref{app:e15}). With standardized features (fixed recipe, three seeds
per rotation, co-primary with the raw head, fixed before its runs), no $\lastarm-\lparm$ meets the
consistency criterion: all six means lie within $\pm2$ points of zero (Appendix~\ref{app:d27b}); for BIOT on
the Siena paths equivalence within $\pm3$ is not shown.

\begin{table}[t]
\caption{BIOT with the raw encoder-level linear probe: SA gain $\lastarm-\lparm$ in ROC-AUC $\times100$
under the fixed probe recipe and under the tuned probe. The frozen representations are identical across the three
columns, as are the units (five rotation means, or five seeds on one fixed split for DL's split); only the
probe training differs. The last two columns are the two equal-budget searches: the original search, frozen before its runs
(16 configurations, at most 200 epochs) and the expanded one that replaces it (30 configurations, at most
500 epochs), which contains the first. Bracket: 95\% $t$-interval over the five units (df 4)
(fixed-recipe intervals: Tables~\ref{tab:d24a} and~\ref{tab:e19}; all 18 comparisons of the expanded
search, with Nadeau--Bengio intervals for the subject rotations: Tables~\ref{tab:d28}
and~\ref{tab:d28b}).}
\label{tab:tuned}
\centering\footnotesize
\begin{tabular}{lccc}
\toprule
setting (BIOT, raw features) & fixed recipe & 16 configs, 200 ep. & 30 configs, 500 ep.\\
\midrule
CHB-MIT rotations        & $+1.9$ & $-0.5$ [$-$1.0, $-$0.0] & $-0.5$ [$-$1.6, $+$0.6]\\
Siena rotations          & $+9.2$ & $+0.2$ [$-$1.4, $+$1.8] & $+1.7$ [$+$0.1, $+$3.3]\\
Siena filtered rotations & $+4.1$ & $+0.7$ [$-$1.6, $+$3.0] & $-0.1$ [$-$1.7, $+$1.5]\\
DL's split, q95 input    & $+8.8$ & $+0.9$ [$-$0.1, $+$1.8] & $+0.6$ [$-$0.006, $+$1.3]\\
\bottomrule
\end{tabular}
\end{table}

\begin{figure}[t]
\centering
\IfFileExists{figures/fig_absolute.pdf}{\includegraphics[width=\linewidth,height=0.3\textheight,keepaspectratio]{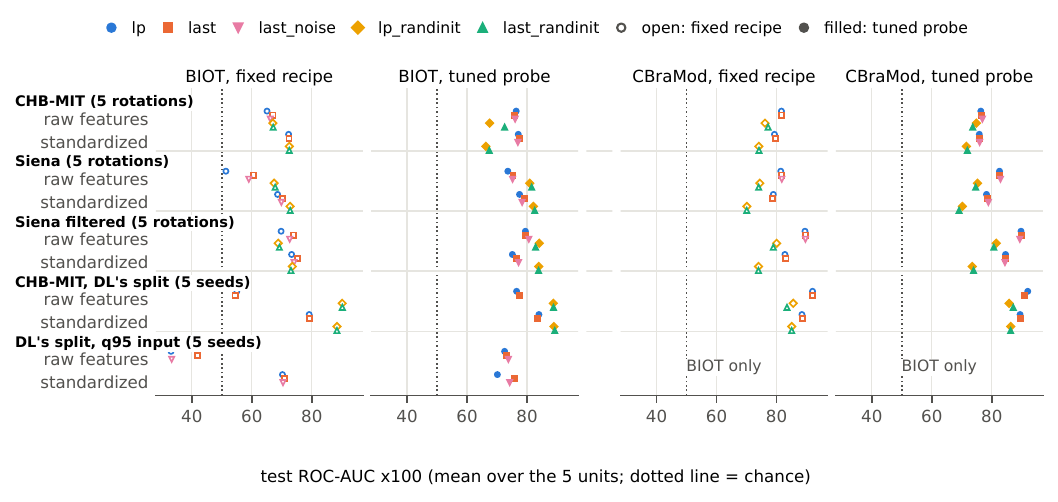}}{\fbox{\Pending{fig-absolute}}}
\caption{Absolute test ROC-AUC of every arm, fixed probe recipe (open markers) against tuned probes (filled),
so that a difference between two poor arms is visible (ROC-AUC $\times100$). Two panels per model on one
shared scale; within a row, top to bottom, \lparm{} (circle), \lastarm{} (square), \lastnoise{} (triangle
down), \lpr{} (diamond) and \lastr{} (triangle up). Marker: mean over the 5 units; dotted line at 50:
chance. Rows, settings, units and the missing-marker convention as in Figure~\ref{fig:effects}. Values: Table~S38.}
\label{fig:absolute}
\end{figure}

\paragraph{Tuned probes.} The fixed recipe leaves the raw BIOT head under-trained
(Figure~\ref{fig:curvesrot}), so every arm of the main comparisons was re-read with the tuned probes
(Section~\ref{sec:probes}, Appendix~\ref{app:d28}).
Table~\ref{tab:tuned} compares adaptation effects before and after probe
tuning: the BIOT raw-probe gains are substantially reduced under the expanded search,
with effects from $-0.5$ to $+1.7$ across these settings (every pair: Figure~\ref{fig:effects}). The absolute scores rose
substantially, frozen BIOT on Siena from 51.3 to 73.6 and every q95 arm on DL's split from below 42 to
above 72 (Figure~\ref{fig:absolute}), so the fixed-recipe differences sat between poorly performing arms.

Across 18 model--setting--feature comparisons the tuned $\lastarm-\lparm$ differences range from $-1.1$ to
$+5.7$ ROC-AUC points, none of them meeting the study's prespecified gain criterion, which requires a
95\% $t$-interval excluding zero \emph{and} a mean above 3 points. Four settings were evaluated for both
models and the q95 setting for BIOT only, each with raw and standardized features:
$(4\times2+1)\times2=18$ related comparisons, not independent replications. Two nonetheless have
$t$-intervals excluding zero, both Siena BIOT: raw, $+1.7$ $[+0.1,+3.3]$, and standardized, $+1.7$
$[+0.3,+3.1]$. Their Nadeau--Bengio intervals, $[-1.0,+4.4]$ and $[-0.7,+4.1]$, include zero, and both means
are below 3, so neither meets that criterion. No positive effect has a 95\% interval excluding zero under
dependence-adjusted inference. Equivalence within $\pm3$ points is supported in 14 of the 18 under ordinary
$t$-based inference and in 12 when the rotation comparisons use dependence-adjusted inference: four across
seeds on DL's fixed split, where the correction does not apply, and eight across subject rotations.
These counts summarise per-comparison tests; they do not establish simultaneous equivalence across all 18
comparisons.

The four unresolved under ordinary inference are standardized BIOT on DL's split ($-0.5$ $[-8.9,+7.9]$),
standardized BIOT with the q95 input ($+5.7$ $[-1.0,+12.3]$), raw CBraMod on CHB-MIT ($+0.3$
$[-6.7,+7.3]$) and standardized BIOT on filtered Siena ($+1.3$ $[-3.1,+5.8]$); the two Siena BIOT
comparisons above join them under the corrected intervals (Tables~\ref{tab:d28} and~\ref{tab:d28b}). The
$\pm3$ margin is an analysis threshold, not a clinically validated equivalence margin, and the count depends
on it: at $\pm2$ the equivalent comparisons fall to 10 under ordinary inference and 6 when adjusted, and at
$\pm1$ to 4 and 2. This margin sweep is post hoc; the registered margin is 3.
The wider search reduces the boundary problem without removing it: 864 of the 1,010 selected fits still sit
at a grid edge (85.5\%, against 944 of 1,010 under the narrower search), 428 chose the new largest learning
rate $10^{-2}$, and all selected configurations stopped before 500 epochs. Of 30,300 candidates, 491 reached that cap.
Early stopping does not establish optimization convergence; the protocol specified one expansion.

\paragraph{Converged linear probes.} To assess sensitivity to probe optimization, we fit L2 logistic
regression on the same frozen features in two BIOT settings: DL's split with q95 input and the Siena
rotations. Numerical convergence is checked on the convex training objective. The regularization $C$ was selected on validation only, over
$10^{-4}$ to $10^{2}$, and every selected model was refit at a tighter tolerance. All 840 candidate fits met
the convergence criterion, a gradient infinity-norm below $10^{-5}$ on the objective the solver minimises,
and all 120 selected models passed the validation-AUC stability check (Appendix~\ref{app:d33}).
Table~\ref{tab:d33} gives the eight contrasts. None demonstrates an improvement. Five support equivalence
within $\pm3$ points under the applicable inference, ordinary $t$ on the fixed split and
dependence-adjusted on the rotations; three Siena comparisons are unresolved, and the widest still permits
$+10.8$ points. Converged probes therefore reduce the concern that these findings come from insufficient
probe optimization, but they do not establish convergence of the searches above.

\begin{table}[t]
\caption{BIOT with an L2 logistic-regression probe on the same frozen features: adaptation effects in
ROC-AUC $\times100$. Units are five seeds on DL's fixed split or five rotation-level differences on Siena.
Siena uses paired differences averaged over three seeds per rotation; LP is a shared deterministic baseline. Bracket: 95\% interval, ordinary
$t$ on the fixed split and dependence-adjusted on the rotations; verdict from TOST at margin 3,
$\alpha=0.05$. All fits converged numerically. Full diagnostics: Appendix~\ref{app:d33}.}
\label{tab:d33}
\centering\footnotesize
\begin{tabular}{llcrl}
\toprule
setting & features & contrast & mean [95\%] & verdict\\
\midrule
DL's split, q95 & raw          & \lastarm$-$\lparm      & $+0.19$ [$-$1.3, $+$1.7] & equivalent\\
DL's split, q95 & raw          & \lastarm$-$\lastnoise  & $-0.90$ [$-$2.4, $+$0.6] & equivalent\\
DL's split, q95 & standardized & \lastarm$-$\lparm      & $+0.61$ [$-$0.7, $+$1.9] & equivalent\\
DL's split, q95 & standardized & \lastarm$-$\lastnoise  & $+0.49$ [$-$2.6, $+$3.6] & equivalent\\
Siena rotations & standardized & \lastarm$-$\lastnoise  & $+0.46$ [$-$0.6, $+$1.5] & equivalent\\
Siena rotations & standardized & \lastarm$-$\lparm      & $+0.38$ [$-$4.8, $+$5.6] & not resolved\\
Siena rotations & raw          & \lastarm$-$\lastnoise  & $+1.82$ [$-$6.3, $+$9.9] & not resolved\\
Siena rotations & raw          & \lastarm$-$\lparm      & $+2.87$ [$-$5.1, $+$10.8] & not resolved\\
\bottomrule
\end{tabular}
\end{table}

None of the 14 EEG-minus-noise comparisons in the expanded probe search meets the gain criterion.
Equivalence within $\pm3$ is supported in 13 under ordinary
$t$-based inference, and in eight under dependence-adjusted inference.

Pretrained against randomly initialised features (how the sign was read: Appendix~\ref{app:d28}), CBraMod
retains its sign in all 16 comparisons, seven meeting the gain criterion, although no rotation comparison does
so with the Nadeau--Bengio interval. For BIOT 7 of 16 signs change, 6 from fixed-recipe means within
$\pm2.2$ points; the random-initialisation advantage on DL's split shrinks ($\lastr-\lastarm$
$+35.8\to+11.4$ raw); and on the CHB-MIT rotations the standardized tuned
probe puts pretrained features ahead, with a Nadeau--Bengio interval that includes zero.

\paragraph{Precision--recall performance.} We report average precision (AP) and the release's trapezoidal
PR-AUC at the validation ROC-AUC epoch, without applying the ROC-AUC equivalence margin. With the fixed
raw head, Siena BIOT improves AP over LP by +0.55 $[+0.14,+0.96]$ and over noise by +0.28 $[+0.05,+0.52]$
(ordinary intervals). With expanded probe tuning, the corresponding estimates are +0.78 $[-0.67,+2.24]$ and
+0.12 $[-1.02,+1.27]$; dependence-adjusted intervals include zero. Fixed-compute C2/C3 results are
reported in Section~\ref{sec:c2c3}; full metric readings and baseline levels are in
Appendix~\ref{app:d29} and Tables~S35 and~S43.

\paragraph{Input normalisation.} BIOT was pretrained with per-channel q95 normalisation, while the release
divides by 100. With q95 input and the fixed recipe, 6 of 12 BIOT classifications change and frozen BIOT
gets \emph{worse}; two q95 cells meet the descriptive consistency criterion (filtered Siena +3.7, DL's split +8.8;
Table~\ref{tab:e19}). On DL's split the fixed raw heads gave the only prespecified ROC comparisons in which
EEG adaptation exceeds noise adaptation ($\lastarm-\lastnoise$ $+8.6$ $[+5.9,+11.2]$ raw head, $+9.1$
shipped MLP head), between the below-chance arms of Figure~\ref{fig:absolute}, whose heads were
under-trained. The expanded search covers linear probes only. The tuned EEG-minus-noise effect is $-0.6$ $[-1.5,+0.2]$ raw
(equivalent within $\pm3$) and +1.6 $[-3.2,+6.3]$ standardized (unresolved). The standardized EEG-minus-LP
estimate is +5.7 $[-1.0,+12.3]$, also unresolved; the raw estimate is +0.6 and equivalent
(Tables~\ref{tab:d28} and~\ref{tab:d28b}).

\paragraph{Second subject partition and leave-one-rotation-out (fixed recipe).} On a second subject partition
13 of 16 classifications replicate; the other 3 are BIOT CHB-MIT with an under-trained raw head, an SA gain
of +3.6 among them. Dropping one rotation changes 18 of 80 classifications, all BIOT
(Appendices~\ref{app:e32}, \ref{app:d26} and~\ref{app:loro}).

\paragraph{Adaptation settings and other checks.} The shipped MLP gives fixed-recipe BIOT gains of
+1.4, +11.9 and +4.1 and remains outside the tuned analysis. Two larger-dose Siena recipes give fixed raw
probe gains of +21.1 and +21.5. With matched probe tuning, these become +2.4 and +3.1;
all eight EEG-minus-LP/noise ROC intervals include zero, while smaller benefits remain unresolved
(Appendix~\ref{app:e29}). Extended stores, alternative SSL draws and event scoring have their individual
intervals and exceptions in Appendix~\ref{app:robust}.

\paragraph{Full-encoder adaptation.}
\label{sec:mainresult}
CBraMod's fixed-recipe SA gain lies between $-1.5$ and $+2.0$ in every configuration we ran at matched
probe settings. Adapting every parameter, a capacity reference with no noise control and not covered by the
tuned probes (Section~\ref{sec:lim}), is $+9.8$ (fixed recipe) $[+2.0,+17.6]$ above \lastarm{} for BIOT on
the CHB-MIT rotations (Nadeau--Bengio $[-3.0,+22.6]$).

\section{The data-amount and patient-count claims (exploratory)}
\label{sec:c2c3}

C2 and C3 are exploratory: the noise controls do not establish that the ROC-AUC plateau requires
EEG-specific information. We use the encoder-level probe with the fixed recipe (batch 128) and 132 SSL
optimizer steps. C2 varies the SSL window count $N\in\{896, 1664, 2048, 4174\}$; C3 varies patient count
$P$ at $N=2048$. Their design and rules were specified before their runs. Both rules require a curve
more than 3 points above LP; C2 also requires less data to perform better and C3 a range across $P$ of at
most 3 points. A post hoc descriptive reading instead asks whether C2 plateaus from $N\le2048$.
DL plot AUCPR, whereas these rules read ROC-AUC; we report PR effects separately.
Appendix~\ref{app:e30} gives the rules and $P$ values, and Table~\ref{tab:c2c3} gives the curves.

\paragraph{CHB-MIT (DL's split).} The BIOT curve is flat and slightly below the LP baseline and CBraMod sits
at the LP level (Figure~\ref{fig:c2c3}); because there is no adaptation gain on this split to begin with
(Section~\ref{sec:p1e}), neither C2 nor C3 can be assessed here.

\paragraph{Siena.} With the raw probe, BIOT adaptation lifts the ROC-AUC curve over $N$ about 12.7
points above LP (4.5 on filtered Siena), and the curve is flat across $P$ and flat within 0.13
from $N=896$. C3 is therefore supported as written and C2 is supported as a plateau, though not in the
stronger form that less data is better. With the standardized probe the apparent advantage largely
disappears, to about $+2.9$ above LP on Siena ($+0.9$ filtered), which supports neither claim. CBraMod
supports neither claim on either path (Appendix~\ref{app:d27c}).
Noise adaptation produces curves of a similar shape, and with the raw probe the noise curve itself satisfies
the C3 rule.

\paragraph{EEG against noise on these subsets.} With a \lastnoise{} arm on the same subsets,
$\lastarm-\lastnoise$ is $+3.0$ to $+4.2$ on Siena and $+0.9$ to $+1.5$ on filtered Siena with the raw
probe and $-0.4$ to $+1.2$ with the standardized probe, at every $N$ and $P$. Every 95\% $t$-interval
includes zero and TOST establishes equivalence within $\pm3$ in only 6 of the 28 subset comparisons, none
of them on Siena with the raw probe. On ROC-AUC the specificity is thus unresolved.
For BIOT with the fixed raw probe on unfiltered Siena, the EEG-minus-noise average-precision differences
are +0.35 to +0.50 points across all seven subsets, with ordinary and Nadeau--Bengio intervals excluding
zero. The corresponding intervals include zero with standardized features and on filtered Siena
(Appendix~\ref{app:d29}).
These exact fixed-compute curves were
not evaluated with tuned probes (Section~\ref{sec:lim}), and the raw-probe curves use the fixed recipe,
whose head was again under-trained here (Appendix~\ref{app:d29}).

\section{Discussion}
\label{sec:disc}

Initialisation, representation choice and probe training substantially affect apparent adaptation gains.
Correcting the initialisation and probe settings removes the large gains on DL's CHB-MIT split; equal-budget
probe tuning reduces several remaining BIOT effects. Across the 18 tuned comparisons, 12 support
equivalence within $\pm3$ points under dependence-adjusted inference and six are unresolved. The
intervals constrain effects in the tested configurations without ruling out smaller benefits or showing
that SSL adaptation is generally ineffective.

The sensitivity to probe tuning indicates that a fixed training recipe can confound representation quality
with how readily a classifier trains on those features. Small adaptation effects also do not imply that
pretraining is unhelpful: fixed-recipe frozen CBraMod scores above random features in all three settings,
although dependence-adjusted intervals support only some of these contrasts (Section~\ref{sec:d24a}).
Hand-crafted features with logistic regression score higher than BIOT with the fixed-recipe raw probe on
DL's split and the CHB-MIT rotations (descriptive; Appendix~\ref{app:e20}), underscoring the need for
adequately trained baselines.

\paragraph{Recommendations for evaluation.} Future CHB-MIT evaluations should state whether the task
concerns seizure presence or seizure boundaries, avoid duplicate overlapping windows in validation and
test, and assess held-out patients across several splits with intervals over those splits. Adaptation
comparisons should use matched initialisation and equally tuned probes, with validation-based early stopping
and the same search for every arm on raw and standardized features, next to a random-weights and a
noise-input control with an equivalence test. Reporting absolute performance alongside paired differences
helps readers assess the quality of the representations and the sensitivity of the comparison to probe
training.

\section{Limitations}
\label{sec:lim}
\label{sec:unknown}
\begin{itemize}
  \item \textbf{Generalizability.} Two models, one adaptation scheme and frozen probes, on seizure
  detection; Siena, the only second dataset, has 14 subjects. Without the TUH corpus we did not
  evaluate DL's TUAB and TUEV results, and the fine-tuning capacity reference uses one 10-epoch recipe.
  \item \textbf{Probe search.} The tuned probes cover DL's split (both input scalings) and the main
  rotations, and two larger-dose Siena BIOT recipes, but not the second subject partition, the shipped MLP head,
  other SSL doses, the fixed-compute C2/C3 curves or the \allarm{} arm, so BIOT effects there are not constrained by them, and the finite grid does
  not establish an optimum (Section~\ref{sec:probes}).
  \item \textbf{Uncertainty.} With five rotations no exact two-sided sign-flip test can reach $p<0.05$ and
  the consistency criterion has low power (Section~\ref{sec:rules}); because the rotations' training folds
  overlap, several effects that the $t$-interval separates from zero are unresolved under the
  Nadeau--Bengio correction. With the fixed recipe the input normalisation, the probe head, a second
  partition and dropping one rotation each change BIOT classifications (Section~\ref{sec:controls}).
  \item \textbf{Reproduction scope.} Our findings concern the audited release and the configurations
  tested here; the configurations used to produce the published results could not be verified. The
  released code is not bit-reproducible, and five cells were re-run on an L40S after an A10G out-of-memory
  failure, so the GPU difference is not separable from the rerun. The BIOT initialisation comparison
  changed when the number of seeds was changed, although alternative SSL draws did not change the
  adaptation-gain conclusion (Appendix~\ref{app:robust}).
  \item \textbf{Retrospective design choices and census evidence.} Besides the post hoc analyses of
  Sections~\ref{sec:setup} and~\ref{sec:main}, the rule on $\lparm-\lpr$ was fixed after related data had
  been seen, before its own runs. Six of the census's 14 confirmations rest on text only.
  Appendix~\ref{app:hygiene} documents the remaining timing qualifications.
\end{itemize}

\subsubsection*{Broader impact statement}
Evaluation choices can influence which seizure-detection models appear suitable for further development, so
reliable comparisons require clear labels, appropriate subject splits and adequately trained
probes, as the revised protocol of Section~\ref{sec:main} aims to provide. We use only public,
de-identified data (CHB-MIT, ODC-By; Siena, CC BY 4.0) and no TUH data, and evaluated no deployed
clinical system.

\subsubsection*{Reproducibility statement}
Code, decision log, result records and reading scripts: Appendix~\ref{app:checklist}.

\bibliography{refs}
\bibliographystyle{tmlr}

\appendix
\section*{Appendix reading guide}
\begin{itemize}[nosep]
  \item \textbf{Headline comparison} (DL's split, pretrained weights, encoder-level probe):
  Appendix~\ref{app:perrot}, Table~\ref{tab:p1e}. Revised-protocol results: Appendix~\ref{app:moved},
  Tables~\ref{tab:artifacts} and~\ref{tab:d24a}.
  \item \textbf{Tuned probes:} Appendix~\ref{app:d28}, Tables~\ref{tab:d28} and~\ref{tab:d28b}, with the
  probe-strength gate in Appendix~\ref{app:e15} and the standardized co-primary head in
  Appendix~\ref{app:d27b}. Converged linear probes: Appendix~\ref{app:d33}.
  \item \textbf{Statistical uncertainty:} Appendix~\ref{app:stats} (95\% $t$-intervals, the Nadeau--Bengio
  correction for overlapping training folds, TOST and power). Partition sensitivity:
  Appendices~\ref{app:e32}, \ref{app:d26} and~\ref{app:loro}.
  \item \textbf{Reproduction attempts and implementation audit:} Appendices~\ref{app:released}
  and~\ref{app:replication}.
  \item \textbf{Protocol census:} Appendix~\ref{app:census}.
  \item \textbf{Reproduction instructions:} Appendix~\ref{app:repro}.
  \item \textbf{Identifiers} (D and E labels, arms, run tags), defined once: Appendix~\ref{app:glossary}.
\end{itemize}
Supplementary tables and figures are cited as \emph{Table~S$n$} and \emph{Figure~S$n$}. Each names its
source artifact in the accompanying code and result archive; reproduction instructions are in
Appendix~\ref{app:repro}.

\section{Glossary of identifiers}
\label{app:glossary}
The arms are defined in Table~\ref{tab:arms}. D and E labels identify supplementary protocols and
analyses; Tables~S29 and~S30 map them to the descriptive names used in the paper. Analysis timing is
reported in Appendix~\ref{app:hygiene}.
Timing tags: P = rule fixed before any data it reads; P* = fixed before its runs, after related data had been
seen; H = post hoc (the control or analysis was chosen because of the data); D = descriptive (no rule). The paper reports rule outputs with the classifications of
Table~\ref{tab:vocab}; the decision log and our reading scripts print the logged output strings of its second
column, so every classification traces back to the log (for example, the scripts' ``SA helps'' is the output F on
$\lastarm-\lparm$). The output of a rule is called a \emph{classification}.
Probe names, used in this sense everywhere below: the \emph{raw head} is a linear probe on unstandardized
features; the \emph{standardized head} (\texttt{linear\_z}) is the same probe on features standardized with
training-split statistics; the \emph{shipped MLP head} (\texttt{mlp\_fullft}) is the classifier architecture of
the released configuration; \emph{tuned} probes are the expanded-search, early-stopped linear probes
(Appendix~\ref{app:d28}), and \emph{fixed-recipe} probes the single training recipe (10 epochs, lr
$4\times10^{-4}$, weight decay 0.05, cosine schedule). \emph{q95 input} is the per-channel q95 normalisation
used in BIOT's pretraining, against the released \texttt{/100} scaling.

\begin{table}[!htbp]
\caption{Rule output as logged (and printed by our reading scripts) and the classification used in this paper.
For a contrast $a-b$, $a$ is the first term. The letters are those of the appendix tables. Rule names of D12, D15/E30, D26, D27-D,
E4--E7, E15 and E29 are quoted as logged.}
\label{tab:vocab}
\centering\scriptsize
\setlength{\tabcolsep}{3pt}
\begin{tabularx}{\linewidth}{>{\raggedright\arraybackslash}p{3.3cm}>{\raggedright\arraybackslash}p{4.4cm}c>{\raggedright\arraybackslash}X}
\toprule
rule & output as logged & letter & classification in this paper\\
\midrule
consistency criterion (D18 paired rule; on seeds D24-B) & ``first term wins'' & F & consistent difference ($a>b$)\\
 & ``second term wins'' & S & consistent difference ($b>a$)\\
 & ``no difference shown'' & N & no consistent difference\\
 & other script wordings: ``SA helps'', ``random init wins'', ``pretrained wins'', ``pretrained features win'' or ``beat random'', ``random features win'', ``none'' & F, S, N & by the sign of the contrast as written; ``none'' = N\\
random-features control (D17) & ``random features explain it'' & R & no improvement demonstrated over random features\\
 & ``SSL adds to random init'', ``neither rule met'' (never printed in this paper) & -- & consistent difference ($\lastr>\lpr$); neither\\
noise rule (D19; E27, D27-A) & ``the gain needs EEG data'' & G & EEG adaptation exceeds noise adaptation (consistency criterion)\\
 & ``perturbation alone explains it'' & P & no difference demonstrated between EEG and noise adaptation\\
 & ``unclear'' & U & unclear\\
TOST, margin 3 (E10, E11, D27-E) & ``equivalent'' ($p<0.05$) & -- & equivalent within $\pm3$ (used only where TOST rejects)\\
tuned-probe check (D28) & ``gain demonstrated'', ``equivalent within $\pm3$'', ``not resolved'' & gain, equiv., n.r. & as logged (``equivalent within $\pm3$'' only where TOST rejects)\\
fixed-compute controls (D29) & ``the plateau needs EEG data''; ``no difference demonstrated'' & -- & the plateau exceeds noise adaptation at every $N$ (every $P$); as logged\\
C2/C3 rules (D15, E30; also D29) & ``supported as written'', ``not supported'', ``not testable: adaptation adds nothing at any $N$'', ``uninformative'' & sw, ns, nt, u & as logged\\
\bottomrule
\end{tabularx}
\end{table}

\begin{table}[!htbp]
\caption{Arms. DL describe their ``SA'' rows as \lastarm{} (adaptation from the pretrained weights); with
its shipped configurations, the released code computes \lastr{} (Section~\ref{sec:a1}).}
\label{tab:arms}
\centering\small
\begin{tabular}{lll}
\toprule
arm & encoder initialisation & adapted by SSL before the probe\\
\midrule
\lparm{} & pretrained & nothing (frozen)\\
\lastarm{} & pretrained & final block (+ SSL head)\\
\allarm{} & pretrained & every floating-point parameter\\
\lastr{} & random (the released path) & final block (+ SSL head)\\
\lpr{} & random (same seeded init as \lastr) & nothing (frozen)\\
\lastnoise{} & pretrained & final block, on Gaussian noise (E27, D27)\\
\bottomrule
\end{tabular}
\end{table}

\FloatBarrier
\section{Implementation ablations and detailed comparison results}
\label{app:moved}

Rule outputs in these tables use the letters of Table~\ref{tab:vocab}, which gives the classification used in
the text for each; timing tags and probe names as in Appendix~\ref{app:glossary}.

\begin{table}[!htbp]
\caption{Each item (1, 2: issues in DL's release; 3: the inherited protocol), the value it produces on DL's
split and the value with it changed (ROC-AUC $\times100$, mean~$\pm$~SEM over 5 seeds; ``a / b'' = best /
last epoch; 2-seed ranges are over configurations).}
\label{tab:artifacts}
\centering\footnotesize
\setlength{\tabcolsep}{3pt}
\begin{tabularx}{\linewidth}{>{\raggedright\arraybackslash}p{3.3cm}>{\raggedright\arraybackslash}p{3.55cm}>{\raggedright\arraybackslash}p{3.7cm}>{\raggedright\arraybackslash}X}
\toprule
item & quantity (DL value) & with the item & with the change\\
\midrule
1. \texttt{foundation\_dir} set in no shipped file: SSL adapts a random encoder & BIOT SA (DL 83.08) & 86.48~$\pm$~1.89 / 88.58~$\pm$~0.21 & set: 49.31 to 59.91 (released, 2 seeds); SA gain $-0.6\pm1.2$, N (ours, encoder-level probe)\\
 & CBraMod SA (DL 83.67) & 89.97~$\pm$~0.79 / 88.36~$\pm$~0.52 & set: 72.96 to 86.72 (released, 2 seeds, best)\\
2. CBraMod probe reads \texttt{proj\_out} with dropout on & CBraMod LP, linear head (DL 54.14) & 64.35~$\pm$~0.06 / 63.93~$\pm$~0.31 & 92.01~$\pm$~0.02 (released, both lines patched, 2 seeds); 92.00~$\pm$~0.04 (ours)\\
 & CBraMod SA gain, linear head (ours) & $+8.67\pm0.23$, F & $+0.01\pm0.01$, N\\
3a. Two-patient test split & BIOT \lastr$-$\lastarm{} (ours, encoder-level probe) & $+35.8\pm1.0$ & five rotations: +0.2, $t$ $[-16.8,+17.1]$, N\\
3b. Start/end labels and duplicate test windows & BIOT \lastr{} PR-AUC, shipped head (re-scored) & 36.9~$\pm$~0.9 & 37.4~$\pm$~1.2; the random-init advantage stays at +30.9 when the probe also trains on overlap labels (E18)\\
\bottomrule
\end{tabularx}
\end{table}

\begin{table}[!htbp]
\caption{Revised protocol, encoder-level probe, linear head, 3 seeds per rotation (D24-A): mean of the five
per-rotation paired differences (ROC-AUC $\times100$), 95\% $t$-interval over rotation means (df 4) and
rule output (consistency criterion; last column: \lastr$-$\lpr, random-features control). Outputs are P* except the
random-features column (H). Bottom: DL's split with the start/end labels (D24-B), 5 seeds, $t$-interval over
seeds (seed uncertainty only); only the SA gain has a rule there. Nadeau--Bengio intervals and TOST:
Appendix~\ref{app:stats}.}
\label{tab:d24a}
\centering\footnotesize
\setlength{\tabcolsep}{2pt}
\begin{tabular}{llcccc}
\toprule
setting & model & \lastarm$-$\lparm{} & \lastr$-$\lastarm{} & \lparm$-$\lpr{} & \lastr$-$\lpr{}\\
\midrule
CHB-MIT & BIOT & +1.9 [$-$1.0, +4.8] N & +0.2 [$-$16.8, +17.1] N & $-$2.0 [$-$17.0, +13.1] N & +0.1 [$-$0.1, +0.3] R\\
CHB-MIT & CBraMod & $-$0.0 [$-$0.4, +0.4] N & $-$4.4 [$-$7.4, $-$1.4] S & +5.5 [+2.5, +8.4] F & +1.0 [$-$0.9, +3.0] R\\
Siena & BIOT & +9.2 [+2.3, +16.2] F & +7.2 [$-$11.5, +25.9] N & $-$16.1 [$-$40.7, +8.6] S & +0.3 [$-$0.6, +1.3] R\\
Siena & CBraMod & +0.3 [$-$0.4, +0.9] N & $-$7.6 [$-$14.3, $-$0.9] S & +7.0 [$-$2.4, +16.3] F & $-$0.4 [$-$3.1, +2.3] R\\
Siena filt. & BIOT & +4.1 [+0.1, +8.2] F & $-$4.7 [$-$34.5, +25.1] S & +1.0 [$-$32.6, +34.6] N & +0.4 [$-$0.1, +0.9] R\\
Siena filt. & CBraMod & +0.1 [$-$0.0, +0.3] N & $-$10.7 [$-$17.0, $-$4.4] S & +9.4 [+3.6, +15.3] F & $-$1.1 [$-$3.7, +1.5] R\\
\midrule
DL's split & BIOT & $-$0.6 [$-$3.8, +2.6] N & +35.8 [+33.0, +38.5] & $-$35.1 [$-$35.9, $-$34.3] & +0.0 [$-$0.1, +0.1]\\
DL's split & CBraMod & +0.0 [$-$0.0, +0.0] N & $-$8.5 [$-$9.4, $-$7.6] & +6.4 [+4.4, +8.5] & $-$2.1 [$-$4.1, $-$0.1]\\
\bottomrule
\end{tabular}
\end{table}

\begin{figure}[!htbp]
\centering
\includegraphics[width=\linewidth]{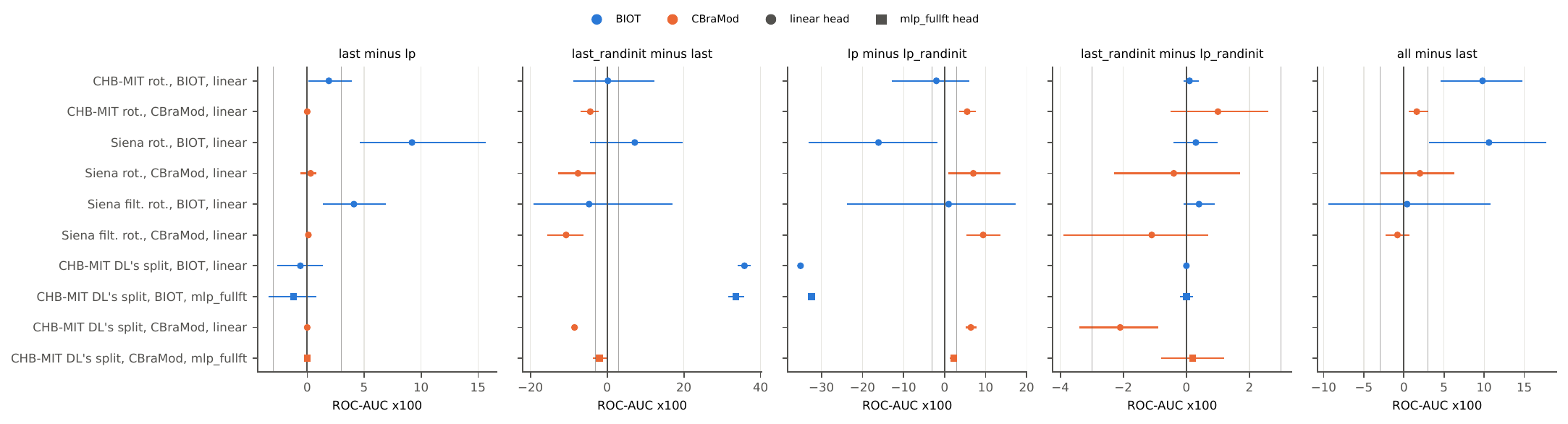}
\caption{Every D24 contrast (encoder-level probe) with its cluster-bootstrap 95\% interval (D24-D; rotations then seeds under the
revised protocol, seeds on DL's split). These intervals under-cover with five rotations and are
descriptive; Table~\ref{tab:d24a} gives the $t$-intervals. Thin lines at $\pm3$ mark the threshold of the
consistency criterion.}
\label{fig:forest}
\end{figure}

\begin{table}[!htbp]
\caption{SSL adaptation gain (\lastarm$-$\lparm, ROC-AUC $\times100$, mean over rotations) under each
control, encoder-level probe, with its rule output (letters: Table~\ref{tab:vocab}; the noise-control rows read
\lastarm$-$\lastnoise). The two G classifications are DL's split with q95 input, one with the raw linear and one
with the shipped MLP head (Appendix~\ref{app:noise}); with tuned linear probes the raw-head difference is
equivalent within $\pm3$ (Appendix~\ref{app:d28}), while the MLP head was not re-run with tuning and has only
its standardized-head control. ``a / b'' = raw or shipped head / \texttt{linear\_z}; SSL-dose rows
count cells; noise row: raw head. ``--'' = not run; $^\dagger$ descriptive.}
\label{tab:robust}
\centering\scriptsize
\setlength{\tabcolsep}{2pt}
\resizebox{\linewidth}{!}{%
\begin{tabular}{lccccccc}
\toprule
 & & \multicolumn{3}{c}{BIOT} & \multicolumn{3}{c}{CBraMod}\\
\cmidrule(lr){3-5}\cmidrule(lr){6-8}
configuration & tag & CHB-MIT & Siena & Siena filt. & CHB-MIT & Siena & Siena filt.\\
\midrule
main runs, raw linear head (D24-A) & P* & +1.9 N & +9.2 F & +4.1 F & $-$0.0 N & +0.3 N & +0.1 N\\
standardized probe, 3 seeds (D27-B) & H & +0.1 N & +1.7 N & +1.9 N & +0.4 N & $-$0.3 N & +0.3 N\\
shipped MLP head (E25) & P* & +1.4 N & +11.9 F & +4.1 F & +0.2 N & +0.1 N & +0.0 N\\
noise control, shipped MLP head (D27 addition) & H & -- & +2.0 / +0.4 P & +1.1 / +1.1 P & -- & -- & --\\
noise control, \lastarm$-$\lastnoise{} (E27) & P* & +0.8 P & +1.6 P & +1.3 P & -- & $-$0.1 P & $-$0.0 P\\
native input scaling, q95 (E19) & P* & $-$1.1 N & +2.1 N & +3.7 F & $-$0.0$^\dagger$ & +0.6$^\dagger$ & +0.2$^\dagger$\\
noise control, q95 (D27-A) & H & -- & -- & +1.4 / +0.4 P & -- & -- & --\\
2 other SSL subsamples, seed $S_r$ (E28) & P* & +2.7, +2.7 N & +6.8, +6.8 F & -- & +0.1, +0.1 N & +0.6, +0.6 N & --\\
noise control, other subsamples (D27 addition) & H & -- & +2.7, +2.7 P & -- & -- & -- & --\\
second subject partition (E32) & P* & +3.6 F & +9.0 F & -- & $-$0.0 N & +0.2 N & --\\
noise control, second partition (D27-A) & H & +1.4 / +0.3 P & +1.6 / +0.4 P & -- & -- & -- & --\\
standardized probe, second partition (D26) & H & +0.7 N & +1.7 N & -- & $-$0.0 N & $-$0.3 N & --\\
SSL dose, cells reading F (E29) & P* & 0 of 5 & 5 of 5 & -- & 0 of 5 & 0 of 5 & --\\
noise control, SSL dose (D27) & H & 3 of 4 P$^\dagger$ & 4 of 5 P$^\dagger$ & -- & -- & -- & --\\
\bottomrule
\end{tabular}}
\end{table}

\FloatBarrier
\section{Protocol census}
\label{app:census}

\paragraph{Search.} We listed every paper that Semantic Scholar shows as citing BIOT or CBraMod
(524 unique), searched the full text of the 355 available on arXiv (64 mention CHB-MIT), used OpenAlex
full-text search for journal papers, and checked GitHub repositories for copies of \texttt{process2.py}
(2026-10-31). It returned 19 candidates, and 3 papers that name the split but use other preprocessing (not
counted).
\paragraph{Verification.} Eight papers have code-level evidence of protocol use, six have textual evidence
consistent with it, and five further papers are probable users. Each candidate was checked against
its text and code on 2026-11-01, with an exact quote and URL recorded per classification. A paper is \emph{confirmed
(code)} if its code copies \texttt{process2.py}, or its repository points to CBraMod's preprocessing
or reads its output. A copied script specifies the split, label rule and duplicate windows; references to
upstream preprocessing or compatible loaders support adoption without establishing how the evaluated files
were produced. It is
\emph{confirmed (text)} if the text states the split and the 326,993-window count or says it follows CBraMod,
which implies but does not show the label rule and the duplicates. Neither class establishes the runtime
configuration of that paper's own runs. Every CHB-MIT number in the census matched
its source. The BIOT text describes a different split (``patient 1 to 19 for training, 20,21 for validation,
and 22,23 for test'', then flipping validation and test) from its code, and several texts (for example
CodeBrain and BrainRVQ) describe the windows as ``non-overlapping'', which holds for the grid windows; the 326,993 count also
includes the seizure windows that the protocol adds at a 5~s stride. Venues were checked on 2026-11-02 against the arXiv API, Crossref and
OpenReview; the bibliography gives the verified fields. The confirmed papers include BIOT (NeurIPS 2023),
CBraMod (ICLR 2025), CodeBrain and TFM-Tokenizer (ICLR 2026), CSBrain (NeurIPS 2025), EEGDM (IEEE JBHI 2026)
and DL (MICCAI 2026).

\begin{table}[!htbp]
\caption{Papers that use (confirmed) or probably use (probable) the BIOT/CBraMod CHB-MIT protocol (\texttt{process2.py}: test chb23/chb24, start/end
label rule, duplicate seizure windows in test). Evidence classes as defined above: \emph{confirmed (code)} =
code-level evidence (8 papers), \emph{confirmed (text)} = textual evidence consistent with the protocol (6),
\emph{probable} = probable user (5). Venues verified as described above; ``per arXiv comment'' =
stated by the authors, not separately confirmed.}
\label{tab:census}
\centering\footnotesize
\setlength{\tabcolsep}{3pt}
\begin{tabularx}{\linewidth}{r>{\raggedright\arraybackslash}p{3.7cm}>{\raggedright\arraybackslash}p{2.9cm}>{\raggedright\arraybackslash}p{1.9cm}>{\raggedright\arraybackslash}X}
\toprule
\# & paper & venue & evidence class & evidence\\
\midrule
1 & BIOT \citep{yang2023biot} & NeurIPS 2023 & confirmed (code) & \texttt{process2.py}; text describes another split\\
2 & CBraMod \citep{wang2025cbramod} & ICLR 2025 & confirmed (code) & differs from BIOT's only in paths\\
3 & EEGMamba \citep{wang2025eegmamba} & Neural Networks 2025 & probable & code byte-identical to CBraMod's; unclear whether the paper evaluates CHB-MIT\\
4 & CodeBrain \citep{ma2026codebrain} & ICLR 2026 & confirmed (code) & preprocessing README points to CBraMod; text states split and count\\
5 & CSBrain \citep{zhou2025csbrain} & NeurIPS 2025 & confirmed (code) & README points to CBraMod; CBraMod's loader\\
6 & TFM-Tokenizer \citep{pradeepkumar2026tfm} & ICLR 2026 & confirmed (code) & \texttt{process\_2.py} and text\\
7 & EEGDM \citep{puah2026eegdm} & IEEE JBHI 2026 (early access) & confirmed (code) & config reads the \texttt{process2.py} output; split stated\\
8 & BrainRVQ \citep{cui2026brainrvq} & arXiv 2026 & confirmed (text) & split and count stated; no public code\\
9 & DeeperBrain \citep{wang2026deeperbrain} & arXiv 2026 & confirmed (text) & dataset table\\
10 & Brain-OF \citep{guo2026brainof} & NeurIPS 2026 (per arXiv comment) & confirmed (text) & split and count stated\\
11 & MTDP \citep{li2026mtdp} & arXiv 2026 & confirmed (text) & ``adopt the fine-tuning code from CBraMod without any modification''\\
12 & TRACE \citep{ma2026trace} & arXiv 2026 & confirmed (text) & split and count stated\\
13 & CAMEL-CLIP \citep{choi2026camelclip} & arXiv 2026 & confirmed (code) & \texttt{process2.py} and text\\
14 & DL \citep{dani2026efficient} & MICCAI 2026 & confirmed (code) & \texttt{process2.py} and text\\
15 & MSBraM \citep{zhou2026msbram} & arXiv 2026 & probable & split matches, but 320,848 samples at 256~Hz\\
16 & SCOPE \citep{ma2026scope} & arXiv 2026 & probable & generic ``prior EFM settings''; 30\% of training subjects labelled\\
17 & S-CEReBrO \citep{bucagu2026scerebro} & MICCAI 2026 & probable & 16 ch, 10~s, 326,993 samples; no split stated\\
18 & CoCoT-EEG \citep{mahuas2026cocoteeg} & arXiv 2026 & confirmed (text) & split follows CBraMod; median/MAD normalisation\\
19 & CRIA \citep{liu2026cria} & Pattern Recognition 2026 & probable (weak) & baseline rows copied from BIOT; split not stated\\
\bottomrule
\end{tabularx}
\end{table}

\section{Released-code runs}
\label{app:released}

All released-code runs use the clone at \texttt{d056cdf} inside the job container, Python environment
from the release's \texttt{requirements.txt} plus the missing packages (torch 2.4.1+cu121 in every cell),
\texttt{training.wandb=False}, and only command-line overrides unless stated. Every rule was read at the
epoch the released trainer keeps (\emph{best}) and at the final epoch (\emph{last}). D13 units are
DL's SEM with a 1-point floor; the E5 noise floor widens every unit threshold by $\delta/\max(\mathrm{SEM},1)$.
Two separately written checks within this project reproduced the reader's output and confirmed its rule
outputs. The follow-up added 70 seed records, giving 267 records in 92 cells; none of the 48 rule outputs
resolved in the first part changed. A reader fix after the follow-up changed no rule or per-configuration
result (Appendix~\ref{app:hygiene}). Tables~S3 to~S8 give the readings; the complete inventory and
re-check reports accompany them in the supplementary material.
Table~\ref{tab:repro-summary} summarises what we tried to reproduce (Section~\ref{sec:repro}).

\begin{table}[!htbp]
\caption{What we tried to reproduce on CHB-MIT. ROC-AUC $\times100$, mean~$\pm$~SEM over 5 seeds on DL's
split unless stated; ``a / b'' = best / last epoch. Released = the released code at \texttt{d056cdf}
with its defaults (shipped MLP head, \texttt{foundation\_dir} unset); ours = our reimplementation,
replication runs (Appendix~\ref{app:replication}). D13 status: consistent = within 3 DL SEM on bal, PR
and ROC; partial = on two of three. The comparison tolerance is this DL-SEM distance, widened by the E5 noise
floor where stated.}
\label{tab:repro-summary}
\centering\footnotesize
\begin{tabularx}{\linewidth}{>{\raggedright\arraybackslash}p{2.9cm}>{\raggedright\arraybackslash}X>{\raggedright\arraybackslash}X>{\raggedright\arraybackslash}p{3.3cm}}
\toprule
item (DL value) & released code & ours & outcome\\
\midrule
Evaluation set (baseline 1.48\%); trainable parameters (0.92M; 1.71M; 0.44M; 0.84M) & 326,993 windows, 252 of 16,964 test windows positive (1.485\%); 920,576; 1,709,568; 442,800; 845,400 & identical stores (E1); same counts & reproduced\\
Documented SSL command; SSL windows (4,174) & stops at \texttt{trainer.py:154} (E2); 81,993 windows & no tested rule gives 4,174 & not reproduced\\
DL Table~1 BIOT LP (46.93~$\pm$~0.30) & 52.32~$\pm$~0.10 / 52.32~$\pm$~0.10 & 52.44~$\pm$~0.16 & partial (ROC off by 5.4 SEM)\\
DL Table~1 BIOT SA (83.08~$\pm$~1.89) & 86.48~$\pm$~1.89 / 88.58~$\pm$~0.21; \texttt{foundation\_dir} set: 49.31 to 59.91 (2 seeds) & random init 88.48~$\pm$~0.19; pretrained 51.99~$\pm$~1.30 to 64.48~$\pm$~0.60 & not reproduced from the checkpoint; random-init path 3.4 to 5.5 above; the released default falls within the comparison tolerance at the best epoch only once that tolerance is enlarged for same-seed variability in balanced accuracy, and not at the last epoch (E5)\\
DL Table~1 CBraMod LP (54.14~$\pm$~1.39) & 84.18~$\pm$~0.96 / 82.41~$\pm$~0.09; linear head 64.35~$\pm$~0.06 / 63.93~$\pm$~0.31 (E7) & 85.32~$\pm$~0.68; linear head 65.36~$\pm$~0.52 & not matched by any configuration we ran (closest: ours 48.46, filtered evaluation, partial)\\
DL Table~1 CBraMod SA (83.67~$\pm$~1.36) & 89.97~$\pm$~0.79 / 88.36~$\pm$~0.52 & random init 89.45~$\pm$~0.93; linear head, batch 128: 85.32~$\pm$~0.59 & consistent with random init (D13); in the released code matched only by non-default configurations, with and without \texttt{foundation\_dir} set (E5)\\
DL Table~2 BIOT layer ablation (88.67; 88.44; 88.60) & layer entry point (E6), best / last: 86.08 / 88.59; 88.45 / 88.39; 88.07 / 87.91 & random init 88.48; 88.51; 88.49 & consistent with random init (D12); the E6 layer-ablation pattern is reproduced at the last epoch, not at the best\\
C1 from the pretrained checkpoint (+36.2; +29.5) & -- & encoder-level probe: BIOT $-0.6\pm1.2$, CBraMod $+0.01\pm0.01$ & not reproduced; size rejected (Section~\ref{sec:main})\\
C2 (20--50\% of the data), C3 (patient count) & flat curves (E30c) & on CHB-MIT, C2 supported in no reading and C3 uninformative & not testable on CHB-MIT; on Siena (BIOT) C3 holds and C2 reads as a plateau with the raw head only, and a noise-input curve does the same: exploratory (Section~\ref{sec:c2c3}, Appendix~\ref{app:d29})\\
\bottomrule
\end{tabularx}
\end{table}

\FloatBarrier
\subsection{Shared preprocessing}
The shared CHB-MIT preprocessing (the release's copy), run unmodified, matches our stores on every count:
326,993 evaluation windows, 252 test positives, 206 duplicate (\texttt{-s-}) test windows, 114 truly ictal
test grid windows of which 70 are labelled 0 by the start/end rule, chb12, chb13 and chb17 absent from
evaluation, and 81,993 SSL windows (chb01--chb22). Splits: train 286,964 windows (1,894 positive),
validation 23,065 (100), test 16,964 (252). Keys: 326,993 common, 0 split or label mismatches. Values on
2,002 seeded windows: max relative difference $4.34\times10^{-7}$; SSL store: 81,993 common keys, max
difference 0. By the E1 rule (the counts of an earlier version of this analysis stand only if they match
identically and values agree within $10^{-5}$ relative) those counts stand. The same comparison per claim is
Table~S1.

\FloatBarrier
\subsection{Documented commands on CPU}
\label{app:e2}
On a 2,100-window symlink farm of the evaluation pickles of the shared preprocessing and a 4,174-key SSL store
(CPU, nothing in the clone edited), the documented entry points behave as follows.
\texttt{main.py method=step\_up} (both models) and \texttt{main\_layer.py} (last1, last2) stop with a
\texttt{ConfigAttributeError} at \texttt{trainer.py:154} as documented, and run once
\texttt{+method.spt\_anchor\_l2sp\_lambda=0} is added (SSL-trainable parameters then 920,576 for BIOT and
442,800 for CBraMod; for the layer entry point 920,576 / 1,709,568 and 442,800 / 845,400).
\texttt{main\_diversityN.py +dataset.compute\_N=896} fails Hydra composition (``an item is already at
dataset.compute\_N'') with and without the key; written as \texttt{dataset.compute\_N=896} it needs the key
(\texttt{trainer\_diversityN.py:158}) and then runs. \texttt{main\_diversityP.py} with its defaults raises a
\texttt{TypeError} at \texttt{compute\_P.py:24} (placeholder strings) and, with $N$ and $P$ set, needs the key
(\texttt{trainer\_diversityP.py:159}) and then runs. The same five entry points as a table are Table~S2.

The README documents \texttt{python main.py model=\ldots{} method=step\_up seed=\ldots}, with seeds 42, 123,
2021, 3407 and 8888.
The trainer's own ``Trainable parameters'' lines report the whole model in the SSL phase (BIOT 3,186,945;
CBraMod 69,326,201) and the probe head in the probe phase under the default \texttt{full\_ft=True} (BIOT 257,
one linear layer; CBraMod 64,402,401, a three-layer MLP). In the CPU smoke, \texttt{main.py} and
\texttt{main\_layer.py} with the key give identical final test metrics at seed 42. With
\texttt{model\_factory.py:29} uncommented, the CBraMod \texttt{step\_up} run fails after its SSL epochs
(``Unexpected key(s) in state\_dict: proj\_out.0.weight, proj\_out.0.bias''), so by the E2 rule, fixed in
advance, the commented line cannot simply have been active for DL's SA rows.

\FloatBarrier
\subsection{Differential implementation test}
\label{app:e21}
We run our modules and the released modules in one process from the same state dict, for both models,
both heads and random/pretrained initialisation. Each cell uses two SSL steps on 16 windows with identical
mask and perturbation seeds, the weight transfer, and one probe epoch on 512 windows. Losses, adapted and
head weights, logits and test metrics computed from the released score representation agree within
$10^{-5}$ in all 8 cells (largest difference $8.5\times10^{-8}$; 33 assertions).
Comparing different score representations on the same logits reveals one exception: the release's float32
sigmoid saturates into ties for CBraMod with the MLP head and a random start, giving ROC-AUC 0.5000
against 0.9601 with logit scoring. This is a scoring effect rather than an implementation difference.
The full test output is \texttt{artifacts/analysis/E21/E21.md} in the supplementary material.

\paragraph{The BIOT checkpoint has no SSL prediction head.} The BIOT checkpoint is an encoder state dict whose
57 tensors all load (none missing or extra); it holds no SSL prediction head, so in the release and in our
code that head starts from random weights in the pretrained and the random-init arm alike. Accordingly the SSL
loss at step 0 is the same for pretrained and random BIOT (0.99 against 0.98, CHB-MIT rotations), while
CBraMod, whose checkpoint includes its reconstruction head, starts at 0.029 against 64.6 (the loss curves are
Figures~S4 and~S5).

\FloatBarrier
\subsection{Released-default performance across metrics}
\label{app:e4}
E4 uses the released defaults: shipped MLP head, batch 256, \texttt{foundation\_dir} unset, released-code
probe, whole SSL store.
With the released defaults the four DL Table~1 cells read, at the best epoch, bal / PR / ROC
52.19~$\pm$~0.14 / 2.39~$\pm$~0.04 / 52.32~$\pm$~0.10 (BIOT LP, where best = last),
58.14~$\pm$~2.06 / 32.34~$\pm$~3.78 / 86.48~$\pm$~1.89 (BIOT SA),
57.80~$\pm$~0.61 / 34.74~$\pm$~1.30 / 84.18~$\pm$~0.96 (CBraMod LP) and
58.71~$\pm$~3.01 / 36.40~$\pm$~2.97 / 89.97~$\pm$~0.79 (CBraMod SA); the random-init LP controls reach
60.62~$\pm$~0.54 / 37.33~$\pm$~0.67 / 88.52~$\pm$~0.26 (BIOT) and
53.42~$\pm$~2.26 / 38.15~$\pm$~1.74 / 89.45~$\pm$~0.80 (CBraMod). In D13 units only two cells are even
partial (BIOT LP +1.5 / +0.9 / +5.4 and BIOT SA at best +5.4 / +0.7 / +1.8; CBraMod SA is partial at last,
+0.3 / +2.8 / +3.4), and CBraMod LP is off by +17.4 and +21.6 units on PR and ROC.
Every E4 cell at both epoch rules, with its D13 gaps, is Table~S3.

``Released = ours'' (E4 rule: $|\mathrm{released}-\mathrm{ours}|\le\max(2, 2\times$ pooled SEM) in ROC)
holds in all 8 cells. The smallest margin is CBraMod LP at last (1.39 against 2.00). For BIOT SA at best
(2.00 against 3.80), seed 123 kept epoch 1: this one run both creates most of the gap and widens the threshold.

\FloatBarrier
\subsection{Released configuration sensitivity}
\label{app:e5}
Screen: seeds 42 and 123; a configuration \emph{passes the screening criterion} if it is within 5 widened units
on 2 of 3 metrics at either epoch rule. A configuration \emph{falls within the comparison tolerance} (the rule
is logged as ``reaches'') if 5 seeds lie within 3 widened units on all three metrics; a configuration
that fails the screen at both epoch rules counts as outside the tolerance. Table~\ref{tab:e5} lists ROC at the best epoch;
the reader script prints all three metrics at both rules (\ours).

\begin{table}[!htbp]
\caption{E5 screen (2 seeds), released code, ROC $\times100$ at best (ROC gap in DL-SEM units; the noise floor widens the thresholds, not the gaps).
ft0/ft1 = \texttt{training.full\_ft} False/True; b128/b256 = batch; fd0/fd1 = \texttt{foundation\_dir}
unset/set; sR/s4174 = released SSL store / our 4,174-window store. The E4 default is fd0-ft1-b256.}
\label{tab:e5}
\centering\footnotesize
\setlength{\tabcolsep}{3pt}
\begin{tabular}{lll|lll}
\toprule
configuration & ROC (units) & screen & configuration & ROC (units) & screen\\
\midrule
\multicolumn{3}{l|}{\emph{BIOT LP}} & \multicolumn{3}{l}{\emph{CBraMod LP}}\\
ft0-b128 & 58.86 (+11.9) & in & ft0-b128 & 68.30 (+10.2) & in\\
ft0-b256 & 52.55 (+5.6) & in & ft0-b256 & 64.36 (+7.4) & in\\
ft1-b128 & 58.94 (+12.0) & in & ft1-b128 & 84.53 (+21.9) & out\\
E4 default & 52.46 (+5.5) & in & E4 default & 85.00 (+22.2) & out\\
\midrule
\multicolumn{3}{l|}{\emph{BIOT SA}} & \multicolumn{3}{l}{\emph{CBraMod SA}}\\
fd0-ft0-b128 & 89.14 (+3.2) & in & fd0-ft0-b128 & 85.94 (+1.7) & in\\
fd0-ft0-b256 & 89.08 (+3.2) & in & fd0-ft0-b256 & 86.45 (+2.0) & in\\
fd0-ft1-b128 & 89.20 (+3.2) & in & fd0-ft1-b128 & 89.02 (+3.9) & in\\
E4 default & 83.79 (+0.4) & in & E4 default & 89.05 (+4.0) & in\\
fd1-ft0-b128-s4174 & 56.55 ($-$14.0) & out & fd1-ft0-b128-s4174 & 76.89 ($-$5.0) & in\\
fd1-ft0-b128-sR & 57.66 ($-$13.4) & out & fd1-ft0-b128-sR & 86.52 (+2.1) & in\\
fd1-ft0-b256-s4174 & 49.31 ($-$17.9) & out & fd1-ft0-b256-s4174 & 72.96 ($-$7.9) & in\\
fd1-ft0-b256-sR & 50.67 ($-$17.1) & out & fd1-ft0-b256-sR & 86.53 (+2.1) & in\\
fd1-ft1-b128-s4174 & 57.72 ($-$13.4) & out & fd1-ft1-b128-s4174 & 85.08 (+1.0) & in\\
fd1-ft1-b128-sR & 59.91 ($-$12.3) & out & fd1-ft1-b128-sR & 84.22 (+0.4) & in\\
fd1-ft1-b256-s4174 & 50.66 ($-$17.2) & out & fd1-ft1-b256-s4174 & 86.72 (+2.2) & in\\
fd1-ft1-b256-sR & 52.53 ($-$16.2) & out & fd1-ft1-b256-sR & 83.54 ($-$0.1) & in\\
\bottomrule
\end{tabular}
\end{table}
The 19 configurations that passed the screening criterion then ran seeds 2021, 3407 and 8888 on the same instance type as the
screen (A10G). Per target and epoch rule: BIOT LP stays outside the tolerance at both epoch rules;
BIOT SA falls within it at the best epoch, only through the E4 default (balanced-accuracy gap 5.43 against a widened
threshold of 6.19, margin 0.76 units, which rests on a single same-seed balanced-accuracy difference), and not
at the last epoch; CBraMod LP stays outside it at either rule; CBraMod SA falls within it at both rules, but only in
configurations other than the E4 default. The 9 configurations that fail the screen count as outside the tolerance
(their 2-seed gaps are 12 to 25 units). The classifier head printed in every E4 and E5 log matches
\texttt{training.full\_ft} (133 of 133 seed logs; with \texttt{training.wandb=False} no W\&B mode tag is
logged, so this is the check).

The E5 table of this rule (logged as ``reaches'') --- every target at both epoch rules, with the 5-seed levels, the per-metric gaps and
the widened thresholds --- is Table~S4, and the per-configuration notes are included with it. Two of its facts are used elsewhere in this paper: the E4 default's per-seed ROC
for BIOT SA is 88.67, 78.92, 88.07, 88.21 and 88.51 (seed 123 kept probe epoch 1, balanced accuracy 50.0), and
four configurations come within 3 units of DL's CBraMod SA on all three metrics at both epoch rules even without
the noise-floor widening (fd0-ft0-b128, fd0-ft0-b256, fd1-ft0-b128-sR, fd1-ft1-b128-sR; for example
fd0-ft0-b128, with \texttt{foundation\_dir} unset, \texttt{full\_ft=False} and batch 128:
56.52~$\pm$~2.63 / 25.48~$\pm$~1.66 / 85.28~$\pm$~0.74 at best, per-seed ROC 85.25, 86.62, 82.47, 86.05,
86.02).

\FloatBarrier
\subsection{Layer-ablation entry point}
\label{app:e6}
E6 runs \texttt{main\_layer.py} with \texttt{foundation\_dir} unset for five seeds (our runs alongside:
Table~\ref{tab:fingerprint}). The all-blocks cells (\texttt{foundation\_dir} unset and set) and the three
all-blocks E31 cells ran out of GPU memory on the 24~GB A10G and were re-run unchanged on one NVIDIA L40S
48~GB, so the E6 rows mix GPUs (last block and last two blocks on A10G, all blocks on L40S); since the released
code is not bit-reproducible at a fixed seed, a GPU effect cannot be separated. The provenance of that
attribution, and the fact that the re-run overwrote the failed attempt's log, are in the released-code reading
report and the run log that ship with this submission.
Per-seed agreement of \texttt{main\_layer} last1 with \texttt{main.py}
\texttt{step\_up} within the noise floor: seeds 42, 2021, 3407 agree, 123 and 8888 do not (at best only;
seed 123: PR 0.48 against 0.26, ROC 2.25 against 0.71; seed 8888: PR 1.22 against 0.26); at the last epoch
all five seeds agree within 0.2, so ``entry points agree'' stays False. The layer-ablation pattern
(all three rows within 2 ROC of 88.67 / 88.44 / 88.60) is not reproduced at the best epoch (last1 86.08, 2.59 from
88.67; its seed 123 kept probe epoch 1, ROC 76.67, while the other four seeds give 88.10 to 88.69) and is
reproduced at the last epoch (88.59 / 88.39 / 87.91). This is a reading choice: we read the 2-ROC tolerance in ROC points, as
recorded in part~1. Read as a unit threshold widened by the ROC noise floor (0.71), it would be 2.71, and the
best-epoch pattern would count as reproduced as well.

\begin{table}[!htbp]
\caption{BIOT on DL's split, ROC-AUC $\times100$, mean~$\pm$~SEM (seeds in parentheses; for 2 seeds $\pm$
is the half-range). Ours: replication runs, shipped MLP head, released-code probe (with the linear head
the random-init rows are 90.15~$\pm$~0.15, 90.15~$\pm$~0.28, 90.07~$\pm$~0.15). Released: layer-ablation
entry point (E6), MLP head, best / last.}
\label{tab:fingerprint}
\centering\footnotesize
\setlength{\tabcolsep}{3pt}
\begin{tabular}{lccccc}
\toprule
adapted & DL & ours, random & released, random & released, \texttt{foundation\_dir} set & ours, pretrained\\
\midrule
last block & 88.67 & 88.48 $\pm$ 0.19 (5) & 86.08 $\pm$ 2.36 / 88.59 $\pm$ 0.20 (5) & 52.49 $\pm$ 2.37 / 52.49 $\pm$ 2.37 (2) & 51.99 $\pm$ 1.30 (5)\\
last two & 88.44 & 88.51 $\pm$ 0.53 (2) & 88.45 $\pm$ 0.26 / 88.39 $\pm$ 0.24 (5) & 63.45 $\pm$ 0.61 / 62.52 $\pm$ 0.33 (2) & --\\
all blocks & 88.60 & 88.49 $\pm$ 0.40 (2) & 88.07 $\pm$ 0.49 / 87.91 $\pm$ 0.47 (5) & 87.37 $\pm$ 0.34 / 87.37 $\pm$ 0.34 (2) & --\\
\bottomrule
\end{tabular}
\end{table}

With \texttt{foundation\_dir} set, adapting all blocks gives bal / PR / ROC 50.10~$\pm$~0.10 /
27.46~$\pm$~0.07 / 87.37~$\pm$~0.34 at both epoch rules: ROC 1.23 below DL's 88.60, with balanced accuracy at
chance (one predicted class at the 0.5 threshold).
All three ablation rows at both epoch rules, with and without \texttt{foundation\_dir}, are Table~S5; the
per-seed values ship with it.

\FloatBarrier
\subsection{Representation source and backbone mode}
\label{app:e7}
The post hoc ablation evaluates representation source~$\times$~backbone mode in our pipeline (E17) and
the release (E7), using two one-line patches only for this ablation: uncommenting line 29 of
\texttt{model\_factory.py} and adding \texttt{self.model.backbone.eval()} after line 597 of
\texttt{trainer.py}. Table~\ref{tab:tapmode} gives both pipelines; analysis timing is in
Appendix~\ref{app:hygiene}.

\begin{table}[!htbp]
\caption{CBraMod on DL's split, linear head: representation source (the code-level probe tap: \texttt{proj\_out}
or the encoder output) $\times$ backbone mode. Ours (E17): 5 seeds
($\pm$ = SEM over seeds on one split, i.e.\ seed uncertainty), D24-B rule on the SA gain (F = consistent difference,
N = no consistent difference; Table~\ref{tab:vocab}). Released code (E7, \texttt{training.full\_ft=False}): LP at the best epoch, 2 seeds
($\pm$ = half-range), except the released configuration (5 seeds, E5 follow-up). ROC-AUC $\times100$, mean~$\pm$~SEM. The bottom row is the released configuration.}
\label{tab:tapmode}
\centering\small
\begin{tabular}{lccccc}
\toprule
source, mode & ours \lparm{} & ours \lastarm{} & ours SA gain & rule & released \lparm\\
\midrule
encoder, eval (encoder-level) & 92.00 $\pm$ 0.04 & 92.02 $\pm$ 0.05 & 0.01 $\pm$ 0.01 & N & 92.01 $\pm$ 0.02\\
\texttt{proj\_out}, eval & 83.29 $\pm$ 0.28 & 85.35 $\pm$ 0.14 & 2.06 $\pm$ 0.19 & N & 82.54 $\pm$ 0.06\\
encoder, train & 87.12 $\pm$ 0.47 & 87.53 $\pm$ 0.39 & 0.40 $\pm$ 0.29 & N & 86.12 $\pm$ 0.52\\
\texttt{proj\_out}, train (released) & 65.36 $\pm$ 0.52 & 74.04 $\pm$ 0.33 & 8.67 $\pm$ 0.23 & F & 64.35 $\pm$ 0.06\\
\bottomrule
\end{tabular}
\end{table}

The full E7 grid --- both heads $\times$ representation source $\times$ mode at both epoch rules, with the D13
gaps against DL's CBraMod LP --- is Table~S6.

Representation-source and mode effects (ROC at seeds 42/123, so the released cell enters at 85.00 / 82.48, averaged over the other factor): MLP head, best: source +3.56, mode +3.18; linear,
best: +15.61, +12.04; MLP, last: +0.60, +7.94 (``mode explains'', driven by a collapse of the encoder /
train-mode cell at last); linear, last: +15.05, +12.72. BIOT eval minus train mode: +3.78 (\texttt{full\_ft=True}, BIOT's ELU--linear head) and +3.29
(\texttt{full\_ft=False}). The logged question ``54.14 reachable in released code'' is answered no at both epoch rules. With the E5
follow-up seeds the closest cell (linear head, \texttt{proj\_out} source, train mode, the E5 configuration CBraMod
LP ft0-b256) reads 64.35 at best (+7.35 units) and 63.93 at last (+7.04) over 5 seeds; the verdict is
unchanged.

\FloatBarrier
\subsection{CBraMod probe: CPU checks}
\label{app:cbracheck}
\begin{itemize}
  \item The pretrained \texttt{proj\_out} weight is nearly singular: condition number $8.0\times10^{6}$
  (float64 SVD), 78 of its 200 singular values below 1\% of the largest, and these directions carry
  67.5\% of the encoder-output variance on 512 CHB-MIT windows. A random-init \texttt{proj\_out} has
  condition number $1.2\times10^{3}$ and 3 such directions, carrying 1.5\% of the variance.
  \item On 64 CHB-MIT evaluation windows the pretrained \texttt{proj\_out} output has standard deviation
  0.057, against 7.32 for a random-init head; the pretrained encoder output is 0.107. A linear head
  trained at lr $4\times10^{-4}$ for 10 epochs under-fits features at this scale.
  \item Dropout in train mode moves the pretrained features by 2.3 (encoder) to 2.8 (\texttt{proj\_out})
  times their spread between windows, against 1.1 times for random init.
  \item We ruled out the other causes we checked: the weights load with no missing or unexpected keys, masked reconstruction works
  with the checkpoint, the input format matches official CBraMod, and the result records are internally
  consistent.
\end{itemize}
Figure~S1 plots the spectrum and the feature scales: singular values relative to the largest, pretrained
against random init, and the standard deviation of the encoder and \texttt{proj\_out} outputs on 512 windows,
overall and across windows (pretrained \texttt{proj\_out} output 0.0566 against 7.14 for random init).

\FloatBarrier
\subsection{Released data-amount and patient-count curves}
\label{app:e30c}
The curves themselves --- ROC at both epoch rules for the four data-amount and three patient-count points,
random init and \texttt{foundation\_dir} set, 2 seeds --- are Table~S7.

Random-init curves are flat on ROC (ranges 0.08 to 1.92) and so are the pretrained ones (at most 2.91).
On balanced accuracy, the released selection metric (\texttt{main\_metric\_name = 'acc'} at
\relcode{src/ablation_diversityN/trainer_diversityN.py\#L502}{trainer\_diversityN.py:502}), the curves
are not flat at best (BIOT N 3.63, BIOT P 9.04, CBraMod P 4.98), because tied validation balanced
accuracy at 0.5 often keeps epoch 1. C2 is supported in no curve (descriptive). Other differences between
the released diversity trainers and our fixed-compute mode: LP cosine floor $10^{-6}$ (line 152; 153 in \texttt{trainer\_diversityP.py}),
\texttt{drop\_last=True} (line 173; 174), budget from the store length
(\texttt{main\_diversityN.py:316}) and the $P$-subset rule in \texttt{compute\_P.py}.

\FloatBarrier
\subsection{L2-SP regularization weight}
\label{app:e31}
The sweep itself --- three $\lambda$ values $\times$ model $\times$ \texttt{foundation\_dir} $\times$ entry
point, at both epoch rules, against $\lambda=0$ --- is Table~S8.

The criterion ``$\lambda$ matters'' is met in 8 of 30 comparisons: 8 of the 24 \texttt{step\_up} comparisons and none of the 6
layer-trainer comparisons ($|\Delta\mathrm{ROC}|\le0.88$). The BIOT cells that meet it with \texttt{foundation\_dir} unset come
from the $\lambda=0$ baseline, whose seed 123 kept epoch 1 (per-seed differences at seed 42 are
+0.28, +0.14, +0.13); at last all are within 0.66. The CBraMod cells that meet it with \texttt{foundation\_dir} set are
real and not monotone in $\lambda$ ($10^{-2}$ is worst). No ``non-finite SPT loss'' line occurs in any cell.
In the layer-trainer cells L2-SP was active (the logs print ``Using L2-SP regularization with lambda=$\lambda$''),
and the final SPT loss rises with $\lambda$ (seeds 42 / 123: 0.0022 / 0.0023 at $10^{-4}$, 0.016 / 0.017 at
$10^{-2}$, 0.245 / 0.275 at 1, against 0.0015 / 0.0016 at 0); each of the three SPT epochs saved a new best in
every re-run seed.

\section{Sensitivity analyses}
\label{app:robust}
Unless stated, results use the revised protocol, encoder-level probe, linear head and ROC-AUC $\times100$.
Intervals are ordinary 95\% $t$-intervals over five rotation means (each averaging the three D24 seeds),
or over five seeds on DL's fixed split; dependence-adjusted intervals are identified separately
(Appendix~\ref{app:stats}). Rule letters follow Table~\ref{tab:vocab}: F/S = consistent difference in
favour of the first/second term, N = no consistent difference, R = no improvement demonstrated over
random features. Absence of a demonstrated difference is not equivalence; only a rejected TOST supports
``equivalent within $\pm3$'', and that ROC-AUC margin is not applied to PR metrics.

The fixed-recipe raw head often under-trains on pretrained BIOT, selecting its final epoch in 58 of 90
E15 fits and all 30 A1 fits. These comparisons may understate what the representation supports;
Appendices~\ref{app:e15} and~\ref{app:d28} give the diagnostics and equal-budget tuning.

Separately written checks within this project confirmed the rule outputs of E17--E19, E25--E30, E32,
E33a and E33b, with E26, E26F, E30, E32 and E33a recomputed from raw records. D26, D27-A with its addition,
D27-B, D27-C, D27-D and leave-one-rotation-out outputs were also recomputed from raw records by a
separately implemented analysis of ours. The corrections to the reading reports are incorporated below.
These checks were not external replications.

\FloatBarrier
\subsection{BIOT probe strength and feature scale}
\label{app:e15}
CPU, eval-mode BIOT encoder features, seed $S_r$. Our cached raw head reproduces the 50 D24 records
within 0.02 ROC, and CPU re-adaptation of Siena \lastarm{} matches the GPU losses within $10^{-4}$. Feature
standard deviation (input /100): pretrained 1.184 / 1.196 / 0.408 (CHB-MIT / Siena / Siena filtered),
random init 4.139--4.417 / 8.212--8.757 / 4.382--4.655. Rule~1 (``BIOT conclusions are probe-limited'' if a
z-scored head or a standardized logistic regression moves \lparm$-$\lpr{} by more than 3 in at least 4 of 5
rotations): z-scored head 5/5, 4/5, 4/5 rotations, so it is met on all three datasets and triggers E26.
Rule~2 (q95 input raises frozen BIOT by more than 3 in 4 of 5): 0/5, 1/5, 0/5, not met. Caveats: 48
of 90 logistic-regression fits chose a grid-edge $C$; the raw head's best epoch is its last in 58 of 90
fits. On DL's split frozen BIOT scores 55.0 (raw head), 79.1 (z-scored) and 84.0 (logistic
regression), against 90.1, 88.4 and 89.1 for random features.

\FloatBarrier
\subsection{Standardized linear head}
\label{app:e26}
The E26 one-seed grid --- all four contrasts per model and setting with their 95\% $t$-intervals and rule
outputs, and the seed-matched raw-head control --- is Table~S9; three seeds per rotation supersede it (D27-B,
Appendix~\ref{app:d27b}).

With standardized features at seed $S_r$ no \lastarm$-$\lparm{} comparison shows a consistent difference,
while the raw head still does on Siena with identical SSL losses.
DL's split (P1Z, 5 seeds): BIOT \lastarm$-$\lparm{} $+0.2\pm0.8$ (per seed $-0.8, -1.0, -0.2, -0.5, +3.4$),
CBraMod $+0.2\pm0.1$, both N (no consistent difference). Classifications that change against D24-A (3 seeds, raw head): BIOT 4 of 9, CBraMod 2 of
9; seed-matched: BIOT 6 of 9, CBraMod 1 of 9 (the CBraMod CHB-MIT \lastr$-$\lastarm{} change is a seed-count
effect). Post hoc bootstrap intervals for \lastarm$-$\lparm: BIOT CHB-MIT [$-$0.4, +0.7], Siena [$-$0.1,
+2.1], Siena filtered [+0.1, +0.6]; CBraMod [+0.1, +0.5], [+0.3, +1.2], [$-$0.2, +0.7]; every mean is below 1
point. Integrity: 160 records, evaluation hashes equal to their D24 twins; records do
not store the data variant (filtered runs are identified by run tag and evaluation hash).
Three seeds per rotation: Appendix~\ref{app:d27b}.

\FloatBarrier
\subsection{Standardized head with three seeds}
\label{app:d27b}
D27-B adds seeds $1000+r$ and $2000+r$ to E26 (run tags S1Z, S3Z, S3FZ; 120 records each, both models, the
four D24 arms) and reads the D24 rules on the 3-seed rotation means; on DL's split it reads P1Z (5 seeds) by
D24-B on \lastarm$-$\lparm. The rule, fixed before these runs and after all earlier data had been seen
(post hoc), makes the two heads co-primary: a conclusion \emph{holds} only if the raw head (D24-A) and the
standardized head give the same classification, and is \emph{head-dependent} otherwise. Every record pairs with its
raw-head twin (same evaluation hash, test index and adaptation spec, apart from fields the older records lack),
and the epoch SSL losses are identical
in 60 of 60 adapted runs per setting (20 of 20 on DL's split), so only the head differs.

The co-primary grid --- the standardized-head mean, its 95\% $t$-interval and rule output next to the raw-head
mean and output, for all four contrasts in each of the six rotation settings and on DL's split --- is
Table~S10. Row~2 of Table~\ref{tab:robust} summarises its \lastarm$-$\lparm{} column.

The two heads give the same classification in 23 of 26 comparisons; the 3 head-dependent ones are all BIOT on
Siena: Siena \lastarm$-$\lparm{}
(+9.2 F raw against +1.7 N standardized), filtered Siena \lastarm$-$\lparm{} (+4.1 F against +1.9 N) and
filtered Siena \lastr$-$\lastarm{} ($-$4.7 S against $-$2.2 N). Strictly, D27-B registers the 24 rotation
outputs (21 agree); the two outputs on DL's split are an extension (both agree). No standardized-head
\lastarm$-$\lparm{} comparison shows a consistent difference. Bootstrap 95\% intervals for BIOT \lastarm$-$\lparm{} with the standardized head
(descriptive): CHB-MIT +0.1 [$-$0.7, +1.0], Siena +1.7 [+0.4, +3.3], filtered Siena +1.9 [$-$0.3, +5.4], DL's
split +0.2 [$-$0.8, +1.8]; CBraMod on DL's split +0.2 [+0.0, +0.3]. Frozen BIOT (\lparm) scores 72.3, 68.6, 73.3
and 79.1 (CHB-MIT, Siena, filtered Siena, DL's split) with the standardized head against 65.1, 51.3, 69.8 and
55.0 with the raw head.
One agreement is fragile (descriptive): BIOT Siena \lparm$-$\lpr{} ($-$4.2 S, 4 of 5 rotations negative) needs
rotation 2 negative ($-$2.8), and that comes from one run (seed 1002, \lparm{} test ROC 50.7 at validation
74.4; \lastarm{} 53.6). Without that seed rotation 2 is +5.5 and the classification is N; on the E32 partition the
same cell gives $-$1.6 N (D26). Of the 6 one-seed E26 classifications that differed from D24-A, 3 return to the D24-A
classification with three seeds (BIOT Siena \lparm$-$\lpr; CBraMod CHB-MIT \lastr$-$\lastarm{} and \lparm$-$\lpr). As in
E26, a few BIOT CHB-MIT rotation-0 runs score test ROC below 50 at validation ROC above 86 (\lparm{} at seeds 42
and 1000: 49.9, 44.8; \lastarm{} at 1000: 43.0).

\FloatBarrier
\subsection{Expanded probe search}
\label{app:d28}
The primary tuned analysis uses a matched search over 30 configurations and up to 500 epochs.
Its protocol and primary-analysis designation were specified before its runs, after the fixed-recipe
results were known (Appendix~\ref{app:hygiene}). The original 16-configuration sensitivity analysis is
retained in Tables~S31--S33,
its per-cell fit counts and early-stopping statement Table~S11, and its random-initialisation arms under
tuned probes Table~S12.

\paragraph{Design.} A linear head reads cached frozen encoder features with the backbone in eval mode.
Raw features and features standardized using training-split statistics receive the same search:
learning rates $\{3\times10^{-5},10^{-4},3\times10^{-4},10^{-3},3\times10^{-3},10^{-2}\}$ and weight decay
$\{0,10^{-4},10^{-2},10^{-1},1\}$; AdamW, batch 256, constant learning rate and binary cross-entropy.
A candidate stops after 20 epochs without a validation ROC-AUC improvement greater than $10^{-4}$,
with a 500-epoch maximum. Configuration and epoch are selected on validation ROC-AUC; ties select the
smaller learning rate, then smaller weight decay. Test metrics are read only after selection.

Both probe searches cover the same settings: both models on DL's split with five seeds, BIOT q95 input on that split,
and both models on five CHB-MIT, Siena and filtered-Siena rotations with three seeds per rotation.
Intervals use the five rotation means, with the Nadeau--Bengio correction for overlapping training folds;
fixed-split intervals describe seed variation conditional on the same patients. The gain criterion requires
an ordinary 95\% interval excluding zero and a mean exceeding three points in the same direction;
TOST equivalence uses $\pm3$ and $\alpha=0.05$. This criterion does not establish that the population effect
exceeds three points. Corrected inference and margin sensitivity are reported alongside the ordinary results.

\paragraph{Checks and search limits.} The 505 records contain 1,010 selected fits and 30,300 candidate fits.
All selected configurations stopped before the cap; 491 candidates reached 500 epochs. In 109 of the 1,010
searches at least one candidate reached that cap. The longest selected configuration ran 400 epochs.
Of the selected configurations, 864 lie on a grid boundary and 428 use the largest learning rate.
These diagnostics describe a finite equal-budget search, not established optimization convergence.
Figure~S6 gives the selected validation trajectories; Table~S39 gives the per-cell diagnostics.
The CPU reader validates coverage, selection and the frozen search against the records. A separate
re-computation of selected metrics reproduces all 64 ROC contrasts; its method and precision--recall
readings are in \texttt{scripts/analysis/tuned\_metrics.py} and \texttt{artifacts/analysis/D31/metrics.json}.

\begin{table}[!htbp]
\caption{Expanded-search \lastarm$-$\lparm contrasts in ROC-AUC points. Units are five seeds on DL\textquotesingle s fixed split or five rotation means of three seeds. NB corrects the rotation intervals for overlapping training folds; it does not apply on the fixed split. TOST uses $\pm3$ and $\alpha=0.05$; its $p$ and classification use NB where applicable (eq. = equivalent; n.r. = unresolved). These are per-comparison results, not simultaneous equivalence.}
\label{tab:d28}
\centering\footnotesize
\setlength{\tabcolsep}{3pt}
\begin{tabular}{lllrrrrl}
\toprule
Setting & Model & Features & Mean & 95\% $t$ & 95\% NB & TOST $p$ & Class\\
\midrule
DL's split & BIOT & raw & +0.9 & [-1.7, +3.4] & -- & 0.0399 & eq.\\
DL's split & BIOT & std & -0.5 & [-8.9, +7.9] & -- & 0.226 & n.r.\\
DL's split & CBraMod & raw & -1.1 & [-2.7, +0.6] & -- & 0.0145 & eq.\\
DL's split & CBraMod & std & +0.1 & [-0.3, +0.4] & -- & 8e-06 & eq.\\
DL's split, q95 & BIOT & raw & +0.6 & [-0.006, +1.3] & -- & 0.000239 & eq.\\
DL's split, q95 & BIOT & std & +5.7 & [-1.0, +12.3] & -- & 0.837 & n.r.\\
CHB-MIT & BIOT & raw & -0.5 & [-1.6, +0.6] & [-2.3, +1.4] & 0.00995 & eq.\\
CHB-MIT & BIOT & std & +0.3 & [-0.7, +1.2] & [-1.3, +1.8] & 0.0041 & eq.\\
CHB-MIT & CBraMod & raw & +0.3 & [-6.7, +7.3] & [-11.1, +11.8] & 0.277 & n.r.\\
CHB-MIT & CBraMod & std & +0.1 & [-1.9, +2.0] & [-3.1, +3.2] & 0.0311 & eq.\\
Siena & BIOT & raw & +1.7 & [+0.1, +3.3] & [-1.0, +4.4] & 0.124 & n.r.\\
Siena & BIOT & std & +1.7 & [+0.3, +3.1] & [-0.7, +4.1] & 0.102 & n.r.\\
Siena & CBraMod & raw & +0.0 & [-2.0, +2.0] & [-3.3, +3.3] & 0.0324 & eq.\\
Siena & CBraMod & std & +0.3 & [-1.4, +2.0] & [-2.5, +3.1] & 0.0273 & eq.\\
Filtered Siena & BIOT & raw & -0.1 & [-1.7, +1.5] & [-2.7, +2.6] & 0.019 & eq.\\
Filtered Siena & BIOT & std & +1.3 & [-3.1, +5.8] & [-6.0, +8.7] & 0.281 & n.r.\\
Filtered Siena & CBraMod & raw & +0.1 & [-0.1, +0.4] & [-0.3, +0.6] & 2.6e-05 & eq.\\
Filtered Siena & CBraMod & std & -0.1 & [-1.0, +0.8] & [-1.7, +1.5] & 0.00332 & eq.\\
\bottomrule
\end{tabular}
\end{table}

\begin{table}[!htbp]
\caption{Expanded-search \lastarm$-$\lastnoise contrasts in ROC-AUC points. Units are five seeds on DL\textquotesingle s fixed split or five rotation means of three seeds. NB corrects the rotation intervals for overlapping training folds; it does not apply on the fixed split. TOST uses $\pm3$ and $\alpha=0.05$; its $p$ and classification use NB where applicable (eq. = equivalent; n.r. = unresolved). These are per-comparison results, not simultaneous equivalence.}
\label{tab:d28b}
\centering\footnotesize
\setlength{\tabcolsep}{3pt}
\begin{tabular}{lllrrrrl}
\toprule
Setting & Model & Features & Mean & 95\% $t$ & 95\% NB & TOST $p$ & Class\\
\midrule
DL's split, q95 & BIOT & raw & -0.6 & [-1.5, +0.2] & -- & 0.000702 & eq.\\
DL's split, q95 & BIOT & std & +1.6 & [-3.2, +6.3] & -- & 0.225 & n.r.\\
CHB-MIT & BIOT & raw & -0.1 & [-3.5, +3.2] & [-5.6, +5.4] & 0.11 & n.r.\\
CHB-MIT & BIOT & std & +0.4 & [-0.6, +1.4] & [-1.2, +2.0] & 0.00581 & eq.\\
CHB-MIT & CBraMod & raw & -0.2 & [-2.7, +2.3] & [-4.2, +3.9] & 0.0628 & n.r.\\
CHB-MIT & CBraMod & std & -0.0 & [-1.4, +1.4] & [-2.3, +2.2] & 0.0107 & eq.\\
Siena & BIOT & raw & +0.1 & [-0.7, +1.0] & [-1.2, +1.5] & 0.00211 & eq.\\
Siena & BIOT & std & +0.8 & [-1.2, +2.8] & [-2.5, +4.2] & 0.0727 & n.r.\\
Siena & CBraMod & raw & -0.3 & [-1.7, +1.1] & [-2.6, +2.0] & 0.016 & eq.\\
Siena & CBraMod & std & -0.3 & [-1.4, +0.7] & [-2.1, +1.4] & 0.00701 & eq.\\
Filtered Siena & BIOT & raw & -1.2 & [-3.4, +1.0] & [-4.9, +2.5] & 0.125 & n.r.\\
Filtered Siena & BIOT & std & -0.7 & [-2.5, +1.1] & [-3.7, +2.3] & 0.0504 & n.r.\\
Filtered Siena & CBraMod & raw & +0.5 & [-0.8, +1.9] & [-1.7, +2.8] & 0.0186 & eq.\\
Filtered Siena & CBraMod & std & +0.1 & [-0.7, +0.9] & [-1.3, +1.5] & 0.00217 & eq.\\
\bottomrule
\end{tabular}
\end{table}

\paragraph{Sensitivity to the search.} Section~\ref{sec:controls} interprets the primary results above.
The narrower search gave 16 ordinary and 13 corrected equivalences, with effects from $-4.4$ to $+0.9$;
the expansion changes these to 14 and 12, with effects from $-1.1$ to $+5.7$. The narrower-search results
are retained as a sensitivity analysis, rather than pooled with the primary comparisons.

All 64 expanded-search contrasts, including pretrained-against-random comparisons, are in Tables~S36 and~S37;
absolute arm levels are Table~S38. The pretrained/random contrasts are discussed in
Section~\ref{sec:controls}.

\subsection{Converged linear probes}
\label{app:d33}
This additional analysis uses L2 logistic
regression on the cached frozen features of two BIOT settings, DL's split with the q95 input (5 seeds) and
the Siena rotations (5 rotations $\times$ 3 seeds), for \lparm, \lastarm{} and \lastnoise. Its protocol
was specified before its runs (Appendix~\ref{app:hygiene}). Within each arm the
raw and standardized probes share that arm's one extracted feature matrix, and standardization statistics
come from its training split only; all arms in a setting use identical split definitions and extraction.
Regularization $C\in\{10^{-4},10^{-3},10^{-2},10^{-1},1,10,10^{2}\}$ is selected on validation ROC-AUC
alone, ties taking the smaller $C$; scores are the linear decision values. Test metrics are evaluated
after selection and never guide tuning.

\paragraph{Numerical convergence.} The solver is L-BFGS at tolerance $10^{-6}$ with a
cap of 20{,}000 iterations. A fit counts as numerically converged when it stops below that cap with no
convergence warning \emph{and} the gradient infinity-norm of the objective the solver minimises is below
$10^{-5}$. That objective is the mean-normalised one, $\frac{1}{n}\sum_i \mathrm{BCE}_i +
\lVert w\rVert^2/(2Cn)$ with the intercept unpenalised; the summed form has the same minimiser but a
gradient $n$ times larger, which can falsely classify converged fits against the same fixed threshold. All 840 candidate fits
met the criterion. Separately, every selected model was refit at tolerance $10^{-9}$; all 120 moved
validation ROC-AUC by less than 0.1 points, which is a validation-AUC stability check and not a statement
about coefficients, scores or individual predictions.

\paragraph{Selected regularization.} 45 of the 120 selected fits sit at a boundary of the
$C$ grid. The selections spread across it: 39 at $10^{-4}$, 35 at 1, 16 at $10^{-3}$, 13 at $10^{-1}$, 9 at
$10^{-2}$, 6 at $10^{2}$ and 2 at 10. A boundary selection means validation preferred the strongest or
weakest regularization offered; it bears neither on convergence at that $C$ nor on whether a better $C$ lies
outside the grid, and the protocol did not widen it.

\paragraph{Interpretation.} The targeted BIOT findings persist with
numerically converged, validation-selected regularized linear probes on the same frozen features. These
results assess sensitivity to probe fitting; they establish neither optimal test performance nor
convergence of the earlier early-stopped searches. On DL's fixed split the five-seed intervals describe
adaptation randomness conditional on those patients and that data, not uncertainty across new patients.
\lparm{} returned identical values across the seed-labelled runs in each split because its inputs and
fitting procedure were fixed, so its repeated values are a shared baseline rather than independent replicates; the
Siena effect is therefore the rotation-level difference between the seed-averaged adapted arm and that
baseline. Results: Table~\ref{tab:d33}. Records and their source directories are listed in
\texttt{artifacts/analysis/D33/manifest.json}, which identifies these records because their Siena run tag is
shared with the earlier shipped-MLP analysis.

\subsection{Noise-input adaptation control}
The E27 rows are the top block of Table~\ref{tab:noise} (D19 output, 90\% and 95\% $t$-intervals, TOST
$p$ and the arm means), which collects every noise control in one place; the bootstrap intervals of its other
columns are in the E27 reading included in the supplementary material.

The noise run reaches a lower SSL loss than the EEG run (BIOT
CHB-MIT epoch 3: 0.0563 against 0.1385) at nearly the same displacement of the adapted block (0.0163
against 0.0173).
Noise controls for the other rule-passing BIOT gains: Appendix~\ref{app:noise}.

\FloatBarrier
\subsection{Matched noise controls}
\label{app:noise}
These post hoc controls add \lastnoise{} to every rule-passing BIOT SA cell that lacked it. Their
protocols were specified before their own runs (Appendix~\ref{app:hygiene}). Items: A1 DL's split with BIOT's
native q95 input, 5 seeds, raw linear, shipped MLP and standardized heads (P1QN); A2 filtered Siena, q95
(S3FQN); A3 the E32 partition, CHB-MIT and Siena (S1RN, S3RN); A4 the E29 CHB-MIT cells 1, 3, 4, 5 (S1LRN, one
seed per rotation); A5 the E29 Siena cells 1--5 (S3LRN, one seed); A6 Siena and filtered Siena with the shipped
MLP head and the standardized head (S3NM, S3FNM); A7 Siena under SSL draws 11 and 12 (S3DN11, S3DN12; one seed).
Rule, per head (D19 on \lastarm$-$\lastnoise): G, ``EEG adaptation exceeds noise adaptation (consistency
criterion)'' (logged as ``the gain needs EEG data''), if \lastarm{} is larger by the consistency criterion
($\mathrm{mean}>3$ and at least 4 of 5 units positive, the signed form the script implements); P, ``no difference demonstrated between EEG and noise
adaptation'' (logged as ``perturbation alone explains it''), if $|\mathrm{mean}|\le3$; otherwise U, ``unclear''. Units are rotation means over the 3 seeds, or the 5 seeds on
DL's split (seed uncertainty on one split). TOST with margin 3 and $\alpha=0.05$ ($t$, df 4): ``equivalent within
$\pm3$'' only if TOST rejects. The rule text makes A4 and A5 descriptive; read
literally it registers A7, which our reading script had treated as descriptive (its outcome is the same either
way).

\begin{table}[!htbp]
\caption{\lastarm$-$\lastnoise{} (ROC-AUC $\times100$) with its 90\% (TOST) and 95\% $t$-intervals over the 5 units
(A1: seeds on DL's split, i.e.\ seed uncertainty on one split), the D19 output (G = EEG adaptation exceeds noise
adaptation, consistency criterion; P = no difference demonstrated between EEG and noise adaptation; U = unclear;
logged outputs in Table~\ref{tab:vocab}), the TOST $p$ (margin 3) and whether equivalence within $\pm3$ is shown,
and the arm means \lparm{} / \lastarm{} / \lastnoise. BIOT unless stated. These are the registered readings
(E27 and D27; the TOST is post hoc for E27); the one-seed A4 and A5 rows are descriptive and are in the
supplementary material (see below). MLP = shipped MLP head; z = \texttt{linear\_z}.}
\label{tab:noise}
\centering\scriptsize
\setlength{\tabcolsep}{1.5pt}
\begin{tabular}{lllllllll}
\toprule
item & setting & head & \lastarm$-$\lastnoise{} [90\% $t$] & 95\% $t$ & D19 & TOST $p$ & equivalent & \lparm{} / \lastarm{} / \lastnoise\\
\midrule
\multicolumn{9}{l}{\emph{registered}}\\
E27 & CHB-MIT & linear & +0.81 [$-$2.34, +3.96] & [$-$3.29, +4.91] & P & 0.106 & no & 65.1 / 67.0 / 66.2\\
E27 & Siena & linear & +1.58 [$-$1.43, +4.60] & [$-$2.34, +5.51] & P & 0.186 & no & 51.3 / 60.6 / 59.0\\
E27 & Siena, CBraMod & linear & $-$0.07 [$-$0.11, $-$0.03] & [$-$0.13, $-$0.02] & P & $4.7\times10^{-9}$ & yes & 81.4 / 81.7 / 81.8\\
E27 & Siena filt. & linear & +1.26 [$-$0.75, +3.26] & [$-$1.36, +3.87] & P & 0.069 & no & 69.8 / 73.9 / 72.6\\
E27 & Siena filt., CBraMod & linear & $-$0.02 [$-$0.04, $-$0.00] & [$-$0.04, +0.01] & P & $2.3\times10^{-10}$ & yes & 89.5 / 89.6 / 89.6\\
A1 & DL's split, q95 & linear & +8.58 [+6.53, +10.62] & [+5.92, +11.24] & G & 0.998 & no & 33.1 / 41.9 / 33.3\\
A1 & DL's split, q95 & MLP & +9.06 [+6.76, +11.35] & [+6.07, +12.04] & G & 0.998 & no & 32.5 / 41.1 / 32.1\\
A1 & DL's split, q95 & z & +0.46 [$-$0.18, +1.10] & [$-$0.38, +1.29] & P & 0.0005 & yes & 70.2 / 70.8 / 70.4\\
A2 & Siena filt., q95 & linear & +1.43 [$-$2.01, +4.87] & [$-$3.05, +5.91] & P & 0.193 & no & 54.1 / 57.9 / 56.5\\
A2 & Siena filt., q95 & z & +0.36 [$-$0.68, +1.40] & [$-$1.00, +1.72] & P & 0.0028 & yes & 71.5 / 71.8 / 71.5\\
A3 & CHB-MIT, E32 partition & linear & +1.42 [$-$1.05, +3.89] & [$-$1.80, +4.63] & P & 0.122 & no & 61.0 / 64.6 / 63.2\\
A3 & CHB-MIT, E32 partition & z & +0.30 [$-$1.10, +1.70] & [$-$1.52, +2.12] & P & 0.0073 & yes & 72.6 / 73.3 / 73.0\\
A3 & Siena, E32 partition & linear & +1.62 [+0.19, +3.06] & [$-$0.25, +3.49] & P & 0.055 & no & 50.7 / 59.8 / 58.2\\
A3 & Siena, E32 partition & z & +0.43 [$-$0.77, +1.62] & [$-$1.13, +1.99] & P & 0.0051 & yes & 69.4 / 71.1 / 70.6\\
A6 & Siena & MLP & +1.98 [+0.09, +3.88] & [$-$0.49, +4.46] & P & 0.159 & no & 51.3 / 63.2 / 61.2\\
A6 & Siena & z & +0.40 [$-$0.79, +1.59] & [$-$1.15, +1.95] & P & 0.0048 & yes & 68.6 / 70.2 / 69.8\\
A6 & Siena filt. & MLP & +1.08 [$-$1.32, +3.47] & [$-$2.05, +4.20] & P & 0.081 & no & 70.1 / 74.2 / 73.1\\
A6 & Siena filt. & z & +1.13 [$-$1.15, +3.42] & [$-$1.84, +4.11] & P & 0.078 & no & 73.3 / 75.2 / 74.1\\
A7 & Siena, draw 11 & linear & +2.71 [+0.52, +4.91] & [$-$0.15, +5.57] & P & 0.397 & no & 52.2 / 59.1 / 56.3\\
A7 & Siena, draw 12 & linear & +2.73 [+0.54, +4.93] & [$-$0.13, +5.59] & P & 0.405 & no & 52.2 / 59.1 / 56.3\\
\bottomrule
\end{tabular}
\end{table}
\paragraph{Reading.} EEG adaptation met the descriptive superiority criterion (G) in 2 of the 20 registered
comparisons; the remaining 18 showed no demonstrated difference between EEG and noise adaptation under that
criterion (P), and none was unclear (U). Equivalence within $\pm3$ is shown by TOST in 7 comparisons: the two
CBraMod E27 rows and five of the six BIOT standardized-head rows. No BIOT comparison with a raw linear
or shipped MLP head shows equivalence, so for those rows the absence of a demonstrated difference is not
evidence that the two arms are equivalent; every registered BIOT P row has a 95\% $t$-interval that contains 0. In
particular, the Siena BIOT gains with the shipped MLP head (E25, +11.9), under other SSL draws
(E28, +6.8) and at larger SSL doses (E29, +9.9 to +21.5; its controls A5 are one-seed and descriptive, and
cell 1 is unclear) show no demonstrated difference from their noise controls, and none is shown equivalent
to them; the descriptive rows of the same reading are Table~S18, whose registered rows stay here.
\paragraph{The two A1 comparisons.} On DL's split, BIOT with q95 input and the fixed recipe exceeds its
noise control with the raw and shipped MLP heads. The expanded search covers linear probes only: the raw EEG-minus-noise
contrast is $-0.6$ $[-1.5,+0.2]$, equivalent within $\pm3$, while the standardized contrast is +1.6
$[-3.2,+6.3]$ and unresolved. The shipped MLP was not re-evaluated with tuned probes.

\paragraph{One-seed rows (A4, A5; descriptive).} Cell numbers are the E29 learning-rate $\times$ budget cells
of Appendix~\ref{app:e29} (A4: cells 1, 3, 4, 5 on CHB-MIT; A5: cells 1--5 on Siena).
A4 cell 3 with the raw head (+3.98) and cell 5 with the
standardized head (+3.95) read G, each carried by rotation 4 (+19.7 and +15.4); without rotation 4 cell 3 is
+0.05 (post hoc). With the raw head, rotation 4 gives +15.4 to +19.7 in every A4 cell and rotation 0 gives
$-$2.9 to $-$12.2. A5 cell 1 with the raw head is unclear (+3.26); A5 cell 3 with the standardized head has TOST
$p=0.022$ (not read). These runs also give the E29 cells a standardized head at one seed: \lastarm$-$\lparm{}
is +1.4, $-$0.9, +2.8 and +3.5 on CHB-MIT (cells 1, 3, 4, 5) and +3.9, +3.2, +0.5, +4.3 and +3.6 on Siena
(cells 1--5), against +5.9 to +6.8 and +9.9 to +21.5 with the raw head (descriptive; no rule was run on them).
\paragraph{Checks.} Every arm re-run under a new run tag reproduces the earlier run within the tolerance of
Appendix~\ref{app:checklist} ($10^{-6}$): P1QN against P1Q,
S3FQN against S3FQ, and the S1LRN and S3LRN \lastarm{} runs against E29, 95 of 95 values, each \lastnoise{} SSL
sample is derived from its \lastarm{} sample by a fixed hash, and the 340 \lastarm{} / \lastnoise{} / \lparm{}
triples share evaluation data, labels, test index and probe specification, so only the SSL input differs.
The one-seed A4 and A5 rows --- 9 cells $\times$ two heads, with their intervals, the logged D19 outputs and
the per-unit values --- are Table~S13. The full pairing and reproduction inventory is included with it, including the
note that in 9 A4 cells the \lastarm{} and \lastnoise{} records come from the original and the re-run job
respectively (same A10G GPU, more host memory).

\FloatBarrier
\subsection{BIOT's q95 input normalisation}
\begin{table}[!htbp]
\caption{E19, BIOT with per-channel q95 normalisation (q95) against the released /100 scaling, D24-A
arms, 3 seeds per rotation. Mean, 95\% $t$-interval over rotation means (q95; the /100 intervals are in
Table~\ref{tab:intervals} and Table~\ref{tab:d24a}) and rule output.}
\label{tab:e19}
\centering\scriptsize
\setlength{\tabcolsep}{3pt}
\begin{tabular}{llll}
\toprule
contrast & CHB-MIT q95 / (/100) & Siena q95 / (/100) & Siena filt.\ q95 / (/100)\\
\midrule
\lastarm$-$\lparm{} & $-$1.1 [$-$11.5, +9.4] N / +1.9 N & +2.1 [$-$4.2, +8.4] N / +9.2 F & +3.7 [$-$0.5, +8.0] F / +4.1 F\\
\lastr$-$\lastarm{} & +18.5 [+1.0, +36.0] F / +0.2 N & +28.0 [+9.3, +46.6] F / +7.2 N & +18.2 [$-$1.1, +37.4] F / $-$4.7 S\\
\lparm$-$\lpr{} & $-$17.7 [$-$32.9, $-$2.4] S / $-$2.0 N & $-$30.0 [$-$43.0, $-$17.0] S / $-$16.1 S & $-$21.9 [$-$41.9, $-$1.8] S / +1.0 N\\
D17 & $-$0.2 [$-$0.6, +0.2] R / +0.1 R & +0.0 [$-$0.4, +0.5] R / +0.3 R & +0.1 [$-$0.3, +0.4] R / +0.4 R\\
\lparm{} level & 48.3 / 65.1 & 45.4 / 51.3 & 54.1 / 69.8\\
\bottomrule
\end{tabular}
\end{table}
DL's split (P1Q, 5 seeds), q95 against /100: \lparm{} 33.1 / 55.0, \lastarm{} 41.9 / 54.4, \lastr{}
82.8 / 90.2, \lpr{} 82.8 / 90.1 (linear head); SA gain $+8.8\pm0.7$ (95\% $t$ over seeds [+6.8, +10.8], seed
uncertainty on one split; F by D24-B) against $-0.6\pm1.2$; MLP head $+8.6\pm0.9$ ([+6.2, +11.0]) against $-1.2\pm1.2$. Noise controls for the two q95 cells that pass
(A1, A2; Table~\ref{tab:noise}): on DL's split the raw-head gain exceeds its noise control (\lastarm$-$\lastnoise{}
+8.58 linear, +9.06 MLP, G) between arms that all score below chance with a probe that selected its final
epoch in every run (Appendix~\ref{app:noise}), and the
standardized head gives +0.46 (equivalent within $\pm3$); on filtered Siena +1.43 (raw, P, not
equivalent) and +0.36 (standardized, P, equivalent within $\pm3$).
Expanded-search q95 adaptation effects and intervals are in Tables~\ref{tab:d28} and~\ref{tab:d28b};
the tuned noise contrasts are discussed in Appendix~\ref{app:noise}.

\FloatBarrier
\subsection{Alternative SSL subsamples}
Alternative SSL draws preserved the adaptation-gain conclusions but changed the BIOT initialisation comparison
on CHB-MIT; the complete differences and intervals follow, and the per-draw reading is in the supplementary material
(\texttt{artifacts/analysis/E28/e28.md}).
Pool sizes make two draws nearly disjoint on CHB-MIT (expected overlap 5--7\%) and partly overlapping on
Siena (32--52\%). At seed $S_r$, \lastarm{} is unchanged by the draw (mean range 0.0 for both models on both
datasets), while \lastr{} moves for CBraMod (3.3 on CHB-MIT). The classifications that change against D24-A are all in
CHB-MIT \lastr$-$\lastarm{} (seed $S_r$, 95\% $t$-intervals over the five one-seed rotation values): BIOT $-5.9$
S under both draws ([$-$28.1, +16.3]; the base draw at the same seed also reads $-5.9$ S, [$-$28.1, +16.3]),
CBraMod $-2.9$ N under draw 12 ([$-$6.8, +1.1]). Read against the base draw at the same seed ($-2.9$ N,
[$-$4.9, $-$0.9]), only CBraMod under draw 11 changes ($-5.4$ S, [$-$9.4, $-$1.4]). On DL's split the base draw and the four new draws leave BIOT
\lastr{} at 90.2 (range 0.00) and CBraMod \lastr{} between 82.9 and 85.5.
Noise control for the Siena BIOT gains under draws 11 and 12 (A7, seed $S_r$): \lastarm$-$\lastnoise{} +2.71 and
+2.73, P, not equivalent (TOST $p$ 0.397, 0.405; Table~\ref{tab:noise}).

\FloatBarrier
\subsection{Second subject partition}
\label{app:e32}
The E32 grid --- all four contrasts per model and setting on the new partition next to D24-A, each with its
bootstrap and 95\% $t$-interval and rule output, and the replication column --- is Table~S14.
By the rule fixed in advance (same rule output and overlapping bootstrap intervals) 13 of 16 cells
replicate. The three that do not are all BIOT on CHB-MIT, where the new partition gives \lastarm$-$\lparm{}
+3.6 [+0.8, +8.1] F against D24-A's +1.9 N, \lastr$-$\lastarm{} +7.8 [+2.7, +13.8] F against +0.2 N, and
\lparm$-$\lpr{} $-$11.6 [$-$21.5, $-$4.5] S against $-$2.0 N (bootstrap intervals; their 95\% $t$-intervals
over rotation means, [$-$2.4, +9.6], [$-$0.6, +16.3] and [$-$25.3, +2.1], all contain 0).

Per rotation, E32 BIOT CHB-MIT \lastarm$-$\lparm{} is +1.3, +2.4, +0.1, +12.1, +2.1, again with an under-trained
raw head (best epoch = last in 30 of 30 pretrained runs).
 Under the post hoc
reading ``the original mean lies inside the new interval'', 11 of 16 replicate (Siena CBraMod \lparm$-$\lpr{}
and D17 also fail). The standardized-head reassessment is in Appendix~\ref{app:d26}, noise controls
(A3) in Table~\ref{tab:noise}, and leave-one-rotation-out results in Appendix~\ref{app:loro}.

\FloatBarrier
\subsection{Standardized head on the second partition}
\label{app:d26}
D26 repeats E32 (fold seed 20261101, the four D24 arms, seeds $[S_r, 1000+r, 2000+r]$; run tags S1RZ,
S3RZ, 120 records each) with the standardized head. Rule, written after E32 and before any D26 run (post hoc):
for each of the three E32 BIOT CHB-MIT non-replications, ``raw-head effect'' if the standardized head shows no consistent difference,
``partition effect'' if it gives the E32 raw-head classification, otherwise reported as it is; all other cells are
descriptive. Every record pairs with its E32 raw-head twin (same evaluation hash, test index and adaptation
spec; SSL losses identical in 60 of 60 adapted runs per dataset), so only the head differs.

The D26 grid --- standardized-head mean, bootstrap and 95\% $t$-interval and rule output for all four
contrasts per model and setting, next to the E32 raw-head column and the D26 rule outcome --- is Table~S15.
The rule outputs for the three registered cells are
``raw-head effect'' in all three: with the standardized head the E32 partition gives \lastarm$-$\lparm{} +0.7
[$-$1.0, +2.4] N, \lastr$-$\lastarm{} +2.6 [$-$4.5, +9.7] N and \lparm$-$\lpr{} $-$3.4 [$-$11.4, +4.6] N
(95\% $t$ over rotation means).

The three registered cells fail the paired rule in different ways: \lastarm$-$\lparm{} on its mean (+0.7;
per rotation +0.7, +1.3, $-$1.6, +1.7, +1.5), \lastr$-$\lastarm{} on its mean (+2.6; 4 of 5 rotations positive) and
\lparm$-$\lpr{} only on the sign count ($-$3.4, 3 of 5 negative). Descriptively, 7 of the 16 E32 classifications change
with the standardized head; all 16 equal E26 at seed $S_r$ on the original partition, and 13 of 16 equal D27-B.
Besides the three registered cells the standardized head removes four raw-head consistent differences on this partition (post
hoc): BIOT Siena \lastarm$-$\lparm{} and \lparm$-$\lpr, and CBraMod CHB-MIT \lastr$-$\lastarm{} and \lparm$-$\lpr.
Frozen BIOT scores 72.6 (CHB-MIT) and 69.4 (Siena) with the standardized head against 61.0 and 50.7 with the raw
head.

\FloatBarrier
\subsection{Leave-one-rotation-out sensitivity}
\label{app:loro}
Each rule output is re-read five times, each time without one rotation: the paired rule with $|\mathrm{mean}|>3$
and the same sign in at least 3 of 4 rotations, D17 with $|\mathrm{mean}|\le3$ (both fixed before any output of
this reading; an earlier statistics script, E10, had already printed leave-one-out classifications with the same rule
for the 24 D24-A verdicts, and they agree). Over the rule outputs of D24-A, E32, D27-B and D26, 18 of 80 change at least once (25 of 400
drops; D24-A 6 of 24, E32 2 of 16, D27-B 6 of 24, D26 4 of 16). All 18 are BIOT; no D17 classification, no CBraMod classification
and no standardized-head \lastarm$-$\lparm{} classification changes. The one-seed E26 reading, which D27-B supersedes
and which is not in the 80, changes 7 of 24 (3 of them CBraMod). The reading script re-derives the published
strings before it drops any rotation, and an independent re-check from the raw records confirmed every row.

The table itself --- for every cell of D24-A, E32, D27-B, D26 and the superseded one-seed E26 reading, the
five-rotation mean, its rule output and the output with each single rotation dropped --- is Table~S16.

\FloatBarrier
\subsection{Extended recording store and published-style Siena split}
A larger CHB-MIT store and a published-style Siena split left every classification unchanged: all 8
E33a rule outputs equal those of the main protocol, and every E33b contrast lies inside its five-rotation range.
The differences and intervals follow; the stores and per-rotation readings are in the supplementary material
(\texttt{artifacts/analysis/E33A/README.md}, \texttt{E33b/e33b\_pub.md}).
E33a (23-subject CHB-MIT store, 23-subject rotations, seed 20261014; bootstrap interval, then 95\% $t$-interval
over rotation means): BIOT \lastarm$-$\lparm{} +1.6 [+0.1, +2.9] [$-$0.6, +3.8] N, \lastr$-$\lastarm{}
+1.4 [$-$12.3, +15.1] N, \lparm$-$\lpr{} $-2.9$ [$-$16.9, +11.0] N, D17
[$-$0.1, +0.2] R; CBraMod $-0.0$ [$-$0.2, +0.2] N, $-4.8$ [$-$7.4, $-$2.3] S,
+3.9 [+0.4, +7.5] F, [$-$2.8, +0.9] R. The \allarm$-$\lastarm{} reference is +6.7 (BIOT) and +1.1 (CBraMod). The
store adds 78 EDFs (chb02\_16+, chb12 $\times$21, chb13 $\times$33, chb17 $\times$21, chb24\_02, chb24\_22),
all with 16 of 16 channels; three chb12 files with a referential montage were left out.
E33b (Siena published-style split: test PN13, PN14, PN16, PN17; validation PN10, PN12; 3 seeds; descriptive):
BIOT \lastarm$-$\lparm{} +7.4~$\pm$~1.3 (Siena) and +4.8~$\pm$~0.4 (filtered), CBraMod +0.4~$\pm$~0.1 and
+0.0~$\pm$~0.0; every contrast lies inside the D24 five-rotation range.

\FloatBarrier
\subsection{MLP head and filtered CHB-MIT}
\label{app:e25}
With the shipped MLP head the only consistent SA differences are BIOT on the two Siena paths,
\lastarm$-$\lparm{} +11.9 [$-$0.2, +23.9] F on Siena and +4.1 [$-$0.9, +9.0] F on filtered Siena (both
intervals contain 0); every other E25 cell gives the same classification as its linear-head twin, and D17 is R
(no improvement demonstrated over random features) in all six cells. The E25 and E26F grids --- all four contrasts per model and setting, with 95\% $t$-intervals and rule outputs,
and the E26F filtered-CHB-MIT evaluation with both heads --- are Table~S17.

DL's split with the MLP head (P1L): BIOT $-1.2$ (95\% $t$ over seeds [$-$4.6, +2.2], seed uncertainty on one split), CBraMod
+0.0 ([$-$0.4, +0.4]), both N. 25 of 26 E25 rule outputs equal the
linear ones; all 16 E26F outputs equal unfiltered CHB-MIT. Descriptively, choosing the probe epoch by
validation balanced accuracy instead of ROC-AUC changes 11 of 52 rule outputs (both heads; reading the last epoch instead changes none of 52; for example, P1L BIOT
\lastarm$-$\lparm{} is F with the MLP head and S with the linear head under that
selection). BIOT runs with the linear head select a median epoch of 10 (the last) on Siena and on
DL's split (all arms pooled).
Noise control for the two Siena BIOT MLP-head gains (A6, D27 addition): \lastarm$-$\lastnoise{} +1.98 (Siena) and
+1.08 (filtered Siena), P, not equivalent (TOST $p$ 0.159, 0.081); with the standardized head +0.40 and +1.13
(Table~\ref{tab:noise}).

\FloatBarrier
\subsection{Data-amount and patient-count curves}
\label{app:e30}
SSL follows the released fixed-compute path (132 optimizer steps at batch 128, constant lr $3\times10^{-5}$,
lowest-loss pass restored), so smaller subsets are seen more often. $N\in\{896, 1664, 2048, 4174\}$ is the
released rounding of 20, 40, 50 and 100\% of 4,174. By the D15 rule C2 is ``supported for the pretrained FM''
(foundation model) only if the best $N$ is at most 2048 and beats both $N=4174$ and the LP baseline by more
than 3 points. The post hoc plateau reading follows DL's wording (DL Figure~1: performance ``peaks at
20--50\% of data then plateaus''; DL Section~3.3: beyond that point ``additional unique data provides no
measurable benefit''). C3 uses $P\in\{3,11,21\}$ patients on CHB-MIT and $\{2,5,10\}$ on Siena; DL's
Figures~1 and~2 plot AUCPR, and DL note that their CHB-MIT patient range is limited (3--22). On CHB-MIT,
D24-C's one-seed ``not supported'' for BIOT reflects seed 42, the lowest seed.
The Siena BIOT readings hold with the raw head only, and a noise-input curve passes the same C3 rule; the D29
controls do not establish, on ROC-AUC, that the plateau requires EEG-specific information in any setting or
head, so C2 and C3 are exploratory (Appendix~\ref{app:d29}).
\begin{table}[!htbp]
\caption{C2/C3 with the encoder-level probe, pretrained \lastarm{}, ROC-AUC $\times100$. CHB-MIT: DL's split, 5
seeds (E30a primary reading; seed 42 alone = D24-C). Siena: mean over 5 rotations of 3-seed means (E30b);
the CBraMod control rows are at seed $S_r$, against CBraMod LP at seed $S_r$ (D27-C; Appendix~\ref{app:d27c}).
LP = \lparm{} at probe batch 128. $P_1$--$P_3$ = 3/11/21 SSL subjects (CHB-MIT) or 2/5/10 (Siena), at N2048; ``--'' = not run.
Noise-input and standardized-probe controls of the Siena BIOT curves: Appendix~\ref{app:d29}.}
\label{tab:c2c3}
\centering\footnotesize
\setlength{\tabcolsep}{2.5pt}
\begin{tabular}{lcccccccccc}
\toprule
setting & N896 & N1664 & N2048 & N4174 & $P_1$ & $P_2$ & $P_3$ & LP & C2 & C3\\
\midrule
CHB-MIT BIOT, 5 seeds & 64.5 & 64.4 & 64.4 & 64.6 & 63.6 & 64.3 & 64.6 & 65.09 & not testable & uninformative\\
CHB-MIT BIOT, seed 42 & 62.45 & 61.90 & 62.07 & 62.32 & 61.82 & 62.04 & 62.43 & 65.09 & not supported & uninformative\\
CHB-MIT CBraMod, seed 42 & 92.22 & 92.22 & 92.23 & 92.23 & 92.21 & 92.23 & 92.23 & 92.01 & not testable & uninformative\\
Siena BIOT & 70.5 & 70.3 & 70.3 & 70.3 & 69.8 & 70.5 & 70.4 & 57.70 & not supported & supported as written\\
Siena filt.\ BIOT & 79.0 & 79.0 & 78.9 & 79.0 & 78.8 & 79.0 & 79.0 & 74.43 & not supported & supported as written\\
Siena CBraMod, $S_r$ & 84.6 & -- & -- & 84.6 & 84.6 & -- & 84.6 & 82.57 & not testable & uninformative\\
Siena filt.\ CBraMod, $S_r$ & 90.1 & -- & -- & 90.1 & 90.1 & -- & 90.1 & 89.95 & not testable & uninformative\\
\bottomrule
\end{tabular}
\end{table}

\FloatBarrier
\subsection{CBraMod batch-size control on Siena}
\label{app:d27c}
The CBraMod control curve adds nothing at any $N$ on either Siena path; the exception is rotation 0, where one
LP run kept its first epoch. Detail, with the per-rotation reading in the archive
(\texttt{artifacts/analysis/D27CD/README.md} section C):
the E30 CBraMod control curve on Siena could not be read because no CBraMod LP at probe batch 128 existed there.
D27-C (fixed before its runs, post hoc) ran it (S3CLPB, S3FCLPB; CBraMod \lparm, 3 seeds, probe spec identical
to the control's apart from the seed, same evaluation hash in every rotation) and applies the E30 C2/C3 rules to
the control curve at seed $S_r$ against LP at seed $S_r$. Siena: C2 ``not testable: adaptation adds nothing at
any $N$'' and C3 ``uninformative'' (curve 84.56 to 84.61, +1.99 to +2.03 above LP 82.57, range 0.05); filtered
Siena the same classifications (+0.11 above LP 89.95, range 0.00). Against the 3-seed LP means (descriptive) they are
unchanged. Per rotation (descriptive), Siena rotation 0 is the exception (+6.2 above its LP: C2 ``not supported'', C3
``supported as written''); there the seed-42 LP run kept its first epoch (83.40, against 87.85 and 87.16 at the
other two seeds). An earlier descriptive E30 reading on Siena against the batch-256 LP (C3 ``supported as
written'', +3.76 above LP) does not carry over: part of that gap was the probe batch (LP batch 128 minus batch 256: +1.7 on Siena, +0.6 on filtered Siena, at seed $S_r$; descriptive).

\FloatBarrier
\subsection{Fixed-compute controls and precision--recall performance}
\label{app:d29}
D29 asks whether the E30 data-amount and patient-count patterns are specific to EEG adaptation.
Its post hoc protocol was frozen before its runs (Appendix~\ref{app:hygiene}). BIOT on Siena and filtered
Siena uses five rotations, three seeds, seven data subsets, EEG/noise adaptation, raw/standardized heads,
and matched 132-update adaptation budgets (450 records). The final-four-subset C2 reading and C3 rule
are applied as recorded, with their complete text and per-subset intervals in Tables~S19 and~S20 and
\texttt{artifacts/analysis/D29/README.md}. The existing EEG/LP curves reproduce in 240 of 240 records.
A separately written internal analysis confirmed the rule readings (\texttt{D29/recheck.md}).

\paragraph{ROC-AUC.} Neither head establishes an EEG-specific plateau: the noise rule reports no demonstrated
difference in all four setting--head cells. Raw-feature EEG-minus-noise means span +2.96 to +4.24 on Siena
and +0.90 to +1.54 on filtered Siena. All ordinary 95\% intervals include zero; only the filtered-Siena
subset with $P=2$, $N=2048$ supports equivalence within $\pm3$ under ordinary TOST. C3 is supported as written
only with raw features, and a noise-adapted curve passes the same patient-count rule. C2 is unsupported with
the raw head and not testable with the standardized head. These findings retain C2/C3 as exploratory.
Individual subset effects, intervals and rule exceptions remain in Table~S20 and the saved analysis.

\paragraph{Precision--recall.} These post hoc readings report average precision and the release's trapezoidal
PR-AUC at the ROC-selected epoch, with no practical-importance or equivalence margin (D30). The original
fixed-recipe/D28 C1 table is Table~S34; the primary D31 AP readings are Table~S43. Baseline levels and
prevalence are provided with the saved metric readings because relative changes can be large when the
baseline is near prevalence.

With the fixed raw head at the main 51-update budget, Siena BIOT has an AP gain of +0.55 $[+0.14,+0.96]$
over LP and +0.28 $[+0.05,+0.52]$ over noise (ordinary intervals). D31's corresponding raw differences are
+0.78 $[-0.67,+2.24]$ over LP and +0.12 $[-1.02,+1.27]$ over noise; corrected intervals include zero.
At D29's separate 132-update budget the fixed raw head gives EEG-minus-noise AP differences of +0.35 to
+0.50 across all seven subsets, with ordinary and corrected intervals excluding zero. These curves were
not re-evaluated with tuned probes; support at that budget does not transfer to the main budget.
The noise curve itself reaches 2.19--2.24 AP against LP 1.64, accounting for much of the total gain over LP.
The fixed-compute AP spans and their intervals are Table~S35; full readings for both precision--recall metrics
are in \texttt{artifacts/analysis/D29/pr\_auc.md}. No ROC margin is applied to these metrics.

\subsection{Supervised fine-tuning reference}
\label{app:ft}
Recipe (probe variant \texttt{ft}, fixed in D27 before its runs, post hoc): a linear head on the encoder output
and the whole encoder trained on the labels (BIOT 3,186,688 parameters; CBraMod 4,883,600 of 4,924,000, since
\texttt{mask\_encoding} and \texttt{proj\_out} get no gradient at the encoder tap), head learning rate
$4\times10^{-4}$ and encoder $4\times10^{-5}$, AdamW, weight decay 0.05, cosine schedule to $10^{-5}$, 10
epochs, batch 256, binary cross-entropy, epoch selected on validation ROC-AUC, encoder in eval mode throughout.
Arms: \lparm{} (pretrained checkpoint) and \lpr{} (random init), no SSL step; DL's split with 5 seeds (P1FT),
CHB-MIT and Siena rotations with 3 seeds (S1FT, S3FT). The encoder was trained in every run (at the last epoch
its training loss is below that of the linear probe on the same frozen features in 140 of 140 records). The one
pre-stated contrast, logged as ``pretraining helps fine-tuning'', is met if ft(\lparm)$-$ft(\lpr) shows a consistent difference in favour of pretraining (F); everything else here
is descriptive.

\paragraph{Fine-tuning results.} The pre-stated contrast ft(\lparm)$-$ft(\lpr) met the descriptive consistency
criterion in three of the six settings. For CBraMod it is +11.8 (95\% $t$ [+9.3, +14.2]) on Siena, where the
interval excludes 0, and +7.2 ([$-$0.8, +15.2]) on the CHB-MIT rotations, where the criterion is met although
the interval includes 0; on DL's split it is $-0.8$, no consistent difference. For BIOT it is +3.2
([$-$2.7, +9.2]) on the CHB-MIT rotations, again meeting the criterion with an interval that includes 0, +0.3 on
Siena (no consistent difference), and $-3.1$ ([$-$6.1, $-$0.1] over seeds) on DL's split, where random
initialisation scores higher (descriptive). The BIOT CHB-MIT classification
is fragile (descriptive): it survives 2 of 5 single-rotation drops, and read one seed per rotation it is
unchanged at
seeds $S_r$ (+4.28) and $1000+r$ (+4.31) but not at $2000+r$ (+1.06, no consistent difference); the CBraMod
classifications survive every drop
and are unchanged at every seed index (CHB-MIT +5.55, +6.26, +9.76; Siena +11.52, +11.66, +12.10). Fine-tuning lifts
pretrained BIOT far above its frozen probe (+32.5 on DL's split, +17.7 on the CHB-MIT rotations) and lifts
random-init BIOT by 0.5 to 2.2 points on DL's split and 12 to 22 on the rotations, whereas on DL's split the
frozen linear probe on pretrained CBraMod (92.0) scores above CBraMod fine-tuned with this recipe (85.0).
Our BIOT ft(\lparm) AUROC (0.8753) matches BIOT's own number for the checkpoint we use (0.8752), so the
pretrained BIOT checkpoint gives its published fine-tuning result in our pipeline; our CBraMod ft(\lparm) is
below CBraMod's (0.8498 against 0.8892). The ``6 EEG datasets'' BIOT row, which CBraMod's Table~8 cites as
BIOT, comes from a model further trained on the CHB-MIT training set (BIOT Section~3.6), not from unsupervised
pretraining alone, and our balanced accuracy uses a fixed threshold at logit 0 where BIOT's code uses a
validation-derived one, so balanced accuracy does not compare across these recipes.

The fine-tuning tables are Tables~S21 to~S24: fine-tuned ROC-AUC per arm with the pre-stated contrast, its
interval and the number of positive units (S21); the frozen arms next to the fine-tuned model (S22); the
published CHB-MIT fine-tuning numbers of \citet{yang2023biot} and \citet{wang2025cbramod} next to our P1FT runs
(S23); and the full recipe comparison against those papers and their public code (S24).

\FloatBarrier
\subsection{SSL learning rate and data exposure}
\label{app:e29}
E29 varies learning rate $\{3\times10^{-5},3\times10^{-4},10^{-3}\}$ and either 4,174 SSL windows or the
full eligible pool, with three epochs and one seed per rotation. The exploratory criterion is met only
for BIOT on Siena; fixed raw-probe gains reach +21.1 in cell 2 ($10^{-3}$, 4,174 windows) and +21.5 in
cell 5 ($10^{-3}$, full pool). The original dose and one-seed noise results remain in Tables~S17 and~S13.

D32 re-evaluates these two recipes with the D31 probe search on LP, EEG-adapted and matched noise-adapted
features, using three seeds within each of five rotations (90 records). The protocol was frozen before
execution and is additional to the 18-comparison family. EEG/noise arms share initialization, learning rate,
update count and all other adaptation settings. The subsample recipe uses 51 updates; the full-pool recipe uses 96--153 across rotations.

\begin{table}[!htbp]
\caption{BIOT on Siena with two larger-dose SSL recipes (learning rate $10^{-3}$, 4,174 windows in cell 2; full SSL pool in cell 5), the same 30-configuration probe search for LP, EEG and matched noise adaptation, and five rotation means of three seeds. ROC-AUC $\times100$; NB = dependence-adjusted interval. Only cell 5 with standardized features, EEG-minus-LP, supports corrected equivalence within $\pm3$; the other seven contrasts are unresolved.}
\label{tab:d32}
\centering\footnotesize
\setlength{\tabcolsep}{3pt}
\begin{tabular}{llrrrr}
\toprule
Cell & Features & EEG$-$LP & 95\% NB & EEG$-$noise & 95\% NB\\
\midrule
Cell 2 & raw & +2.4 & [-5.3, +10.0] & +0.8 & [-6.5, +8.2]\\
Cell 2 & std & -1.0 & [-6.8, +4.9] & +0.4 & [-4.0, +4.8]\\
Cell 5 & raw & +3.1 & [-5.1, +11.2] & +1.1 & [-3.7, +5.9]\\
Cell 5 & std & +0.8 & [-1.2, +2.8] & +1.5 & [-2.1, +5.1]\\
\bottomrule
\end{tabular}
\end{table}

The large raw-feature gains shrink to +2.4 and +3.1; corrected intervals are $[-5.3,+10.0]$ and
$[-5.1,+11.2]$. All eight ROC contrasts' ordinary and corrected intervals include zero; only standardized
cell 5 versus LP establishes corrected equivalence within $\pm3$. Smaller benefits remain unresolved.
The change indicates sensitivity to probe evaluation, and E29's one-seed coverage differs from D32's three.
All selected D32 configurations stopped before 500 epochs; 263 of 5,400 candidates reached the cap,
in 54 of the 180 searches. Early stopping does not establish an optimum. Full ROC/AP/trapezoidal PR-AUC
contrasts are Tables~S40--S42, with their CPU reader and records included in the supplement. The AP intervals
also include zero; no ROC margin is used for AP.

\subsection{Event-level scoring}
\label{app:e12}
Events: moving-average smoothing $k\in\{1,3,5\}$, alarms closer than 90~s merged, tolerance 30~s before and
60~s after a seizure, events longer than 5~min split. Primary metric: partial AUC up to 2 false alarms per
hour (normalised), D18 paired rule with margin 0.05 (fixed in D25 before computing).

None of the encoder-level comparisons met the descriptive event-level gain criterion. The exception is the
replication runs that use the released-code probe, where CBraMod's difference is large and meets it. An unmet
criterion is not evidence that the event-level effect is zero.
Table~\ref{tab:e12} gives every comparison; per-unit values are included in the supplementary material.

\begin{table}[!htbp]
\caption{E12, \lastarm$-$\lparm{} in normalised pAUC at $k=1/3/5$, with the rule output (N = no consistent difference
at every $k$; F = consistent difference). P1E = encoder-level probe; W1 = replication runs, released-code probe.
Per-unit values: \ours.}
\label{tab:e12}
\centering\footnotesize
\begin{tabular}{llll}
\toprule
set & BIOT (linear) & CBraMod (linear) & rule outputs\\
\midrule
CHB-MIT rotations & +0.011 / +0.017 / +0.022 & +0.002 / +0.000 / +0.001 & N (all)\\
Siena rotations & +0.034 / +0.029 / +0.020 & +0.011 / +0.019 / +0.001 & N (all)\\
Siena filt.\ rotations & +0.039 / $-$0.006 / $-$0.014 & +0.011 / +0.007 / +0.006 & N (all)\\
DL's split, P1E & +0.040 / $-$0.011 / $-$0.003 & +0.005 / +0.003 / +0.005 & N (all)\\
DL's split, W1 & +0.032 / $-$0.029 / $-$0.020 & +0.121 / +0.199 / +0.289 & CBraMod: F\\
\bottomrule
\end{tabular}
\end{table}
Other contrasts (descriptive here): at the event level, random features score consistently above frozen pretrained
BIOT on CHB-MIT (\lparm$-$\lpr{} $-0.254$, $-0.222$, $-0.201$, S) and on DL's split, and
frozen CBraMod scores consistently above random features on filtered Siena at every $k$.

\FloatBarrier
\subsection{Subject-level uncertainty}
Two-stage bootstrap (subjects within rotation, then seeds) and per-subject paired differences. CHB-MIT BIOT
\lastarm$-$\lparm{}: two-stage 95\% CI [+0.7, +3.3], positive in 15 of 20 subjects (Wilcoxon $p=0.006$),
rotation-block sign-flip $p=0.1250$, equivalence within $\pm3$ not shown. CBraMod \lastarm$-$\lparm{} is
equivalent within $\pm3$ under both intervals on all three datasets; D17 is equivalent within $\pm3$ for
both models in every setting under the two-stage interval. On DL's split most of BIOT's random-init
advantage is chb23 (\lastr$-$\lastarm{} +56.7 on chb23, +23.0 on chb24). The CHB-MIT rotation-3 outlier
(+24.4) is likewise carried by chb23 (+64.2), and filtered-Siena rotation 1 (+37.1) by PN03 and PN16.

\FloatBarrier
\subsection{Encoder parameter changes after adaptation}
\label{app:e14}
None of the 220 checked \lparm/\lastarm{} and \lpr/\lastr{} score pairs met the no-op criterion ($r=1$ with
$\max|\Delta|<10^{-6}$). The adapted block moves by a median 1.74\% (BIOT block 3, DL's split) and 1.41\%
(CBraMod block 11); score correlations with the unadapted twin range from 0.8380 to 0.9994 in median by
setting. 236 identical adaptation specifications run under several run tags reproduce epoch losses and
displacements within the stated tolerance. The fixed-recipe probe used in this check does not cache encoder
features; the tuned probes of Appendix~\ref{app:d28} cache them deliberately.

\FloatBarrier
\subsection{Representation source, labels and subject split}
\label{app:e18}
E17 on rotations (seed $S_r$, 1 seed per rotation, descriptive): CBraMod \lastarm$-$\lparm{} is +0.05
(encoder, eval), +1.41 (\texttt{proj\_out}, eval), +0.07 (encoder, train) and +8.61 (\texttt{proj\_out},
train, F).
E18 trains the probe on overlap labels on DL's split (P1F) and uses the start/end labels on the rotations
(S1U). The label change on DL's split removes no consistent difference: BIOT's random-init advantage goes from +35.8 to
+30.9. The split change (rotations, start/end labels) removes BIOT's random-init advantage
(\lastr$-$\lastarm{} +35.8 to $-2.8$) and CBraMod's \lparm$-$\lpr{} consistent difference (F).
The waterfall from the released path to the revised protocol is Table~\ref{tab:waterfall}.

\begin{table}[!htbp]
\caption{From the released SA path to the revised protocol, one change at a time (E13(d)/E18; linear head,
ROC $\times100$, mean~$\pm$~SEM; seeds and rotations resampled jointly for the residual intervals;
descriptive). The first two changes are those of our title (pretrained weights, then the encoder-level
probe). Rows are levels of the SA arm, not gains; the gains are in Table~\ref{tab:tapmode}
and Section~\ref{sec:p1e}.}
\label{tab:waterfall}
\centering\footnotesize
\begin{tabular}{lcc}
\toprule
step & BIOT SA level & CBraMod SA level\\
\midrule
DL Table~1 & 83.08 & 83.67\\
released path (random init, released-code probe) & 90.1 $\pm$ 0.2 & 86.8 $\pm$ 0.5\\
+ pretrained init & 52.5 $\pm$ 1.1 & 74.0 $\pm$ 0.3\\
+ encoder-level probe & 54.4 $\pm$ 1.1 & 92.0 $\pm$ 0.1\\
+ test scoring (grid windows, overlap labels; post hoc) & 55.9 $\pm$ 1.1 & 90.9 $\pm$ 0.1\\
+ training labels (P1F) & 60.2 $\pm$ 0.9 & 90.9 $\pm$ 0.1\\
+ split (five rotations, D24-A) & 67.0 $\pm$ 8.3 & 81.6 $\pm$ 3.4\\
unreproduced residual (DL $-$ released path) & $-$7.1 [$-$7.3, $-$6.8] & $-$3.2 [$-$4.0, $-$2.3]\\
\bottomrule
\end{tabular}
\end{table}

\FloatBarrier
\subsection{Hand-crafted feature baseline}
\label{app:e20}
Log band power (5 bands), log line length and log variance per channel (112 features), standardized logistic
regression with $C$ chosen on validation: CHB-MIT rotations 79.5~$\pm$~6.0, Siena 83.4~$\pm$~4.7, Siena
filtered 84.0~$\pm$~3.2; DL's split 90.3 (start/end labels) and 90.5 (overlap labels). Frozen CBraMod reaches
a higher mean than this baseline on filtered Siena in 5 of 5 rotations; frozen pretrained BIOT with the
fixed-recipe raw linear probe is below it in the mean in every evaluated setting. Re-scoring the
revised-protocol contrasts with the last-epoch scores or with $\ge50\%$-overlap labels changes one
descriptive classification (Siena BIOT \lparm$-$\lpr{} with $\ge50\%$ labels); average precision and PR
lift are reported without classifications.

\FloatBarrier
\subsection{Data checks: the 4,174 count, Siena annotations and QC}
\label{app:datachecks}
The annotation and QC checks reproduce their sources; the exceptions are the 4,174 SSL-window count, which no
tested rule reproduces, 4 of 47 Siena seizures (each explained) and one rotation set that the QC exclusion
empties. Detail, with the full searches in the archive (\texttt{artifacts/analysis/n4174/n4174.md},
\texttt{siena\_annot/siena\_annot.md}, \texttt{qc/qc\_sensitivity.md}):
\textbf{E17-free.} 6,249 candidate rules (1,307 primary) over our SSL store, released-script variants and
cumulative sums: no exact match for 4,174 (or for 86,592 and 216,480); the nearest primary hits are chb10
with $\max|x|<550\,\mu$V (4,150) and a chb01--chb24 variant (86,587).
\textbf{CHB-MIT seizures.} Of the 198 seizures in the PhysioNet summaries, 142 lie in files that the shared
evaluation preprocessing reads (chb12, chb13, chb17 and chb02\_16+ are not processed).
\textbf{E18-free.} Against epilepsy2bids 0.0.7, 43 of 47 Siena seizures are identical and the 4 differences
are explained (clock origin, two times on one annotation line); 198 of 198 CHB-MIT seizures are identical.
Descriptively, 21 of 47 Siena seizures show no onset-locked rise in either line length or 1--30~Hz
power.
\textbf{E19-free.} Dropping the test windows of 62 CHB-MIT files with a QC flag changes no D24-A CHB-MIT
classification (largest change in a mean 0.53). On Siena the exclusion empties at least one rotation's test set, so
the five-rotation rule cannot be applied; on the remaining four rotations three BIOT classifications change
(descriptive).

\FloatBarrier
\subsection{Earlier revised-protocol runs}
\label{app:earlier}
Before D24, the revised protocol ran with one seed per rotation, BIOT in train mode and CBraMod with the
released-code probe, then CBraMod with the encoder-level probe. Three rule outputs changed when BIOT moved to
eval mode and to three seeds: CHB-MIT BIOT \lastr$-$\lastarm{} $-4.2$ S to +0.2 N; filtered Siena BIOT
$-2.7$ N to $-4.7$ S; and CHB-MIT CBraMod \lastr$-$\lastarm{} $-2.9$ N to $-4.4$ S (letters:
Table~\ref{tab:vocab}). The BIOT SA outputs did not change (+2.9, +7.0, +3.9 with one seed). With the
released-code probe CBraMod \lastarm$-$\lparm{} read +8.6 (CHB-MIT, F) and +12.5 (filtered Siena), against
+0.1, +0.6 and +0.2, N, with the encoder-level probe.
The rest of this history --- the one-seed D17/D18/D19 outcome, the D23 re-run outcome, the frozen-CBraMod
levels before and after the probe change, and the dropped early-layer analysis, whose one positive case was
matched by the same update driven by noise inputs --- is in the decision log included in the supplementary material
(entries ``D17/D18/D19 outcome'', ``D21 robustness row outcome'', ``D23 re-run outcome''); it has no table of
its own.

\FloatBarrier
\subsection{Post hoc re-scoring of DL's split}
\label{app:rescore}
The replication runs on DL's split (backbone in train mode; CBraMod with the released-code probe) were
re-scored without retraining under three variants: inherited (252 positives in 16,964 windows), dedup
(duplicates removed, start/end labels; 46 of 16,758) and overlap (duplicates removed, overlap labels; 114 of
16,758). PR-AUC for BIOT \lastr{} is 37.7~$\pm$~0.4 / 4.7~$\pm$~0.1 / 37.9~$\pm$~0.8 with the linear head
and 36.9~$\pm$~0.9 / 4.3~$\pm$~0.2 / 37.4~$\pm$~1.2 with the shipped head (mean~$\pm$~SEM over 5 seeds):
removing the duplicate seizure windows under the inherited labels collapses PR-AUC, and scoring the same
windows with the overlap labels restores it, so in the aggregate the two effects roughly cancel. ROC moves
much less (BIOT \lastr, linear: 90.1 / 86.1 / 91.3). The full re-scoring table (both heads, four arms, both models, ROC and PR-AUC under all three scorings) is
Table~S25, with average precision and lift over prevalence for the same runs. The supplement provides the
saved analyses; full recomputation requires regenerating the omitted per-window scores
(Appendix~\ref{app:repro}).

A similarly large relative PR-AUC improvement to DL's ``20$\times$ AUCPR'' for BIOT (1.53 to 31.52)
occurred only under the random-initialisation path: PR-AUC divided by prevalence is 25.40 for BIOT \lastr{}
against 1.20 for \lparm{} (inherited scoring, linear head), and 34.68 against 3.32 on the revised protocol,
with PR-AUC 9.56 against 0.89. These are not reproductions of DL's values.

\section{Statistics, analysis timing and deviations}
\label{app:stats}
The interval, equivalence and power analyses and the provenance checks are post hoc; sensitivity
analyses are in Appendix~\ref{app:robust}. Effect estimates and confidence intervals provide the
primary evidence, with dependence-adjusted inference for subject rotations. The fixed-recipe rule
outputs are reported as a descriptive consistency criterion.

\paragraph{Intervals and equivalence.} Table~\ref{tab:intervals} gives, for the SA contrast and D17 (and on the rotations for \lastr$-$\lastarm{} and \lparm$-$\lpr), the rule
output next to the t-interval over the five rotation means (df 4), the Nadeau--Bengio corrected interval
\citep{nadeau2003inference} (test/train size ratio 0.338 on CHB-MIT, 0.347 on Siena), the D24-D
bootstrap, and the TOST $p$ for equivalence within $\pm3$. On DL's split the interval is over the five seeds:
it measures seed uncertainty on one split, not test-set sampling, and the Nadeau--Bengio correction is not
defined. A 95\% percentile bootstrap of a mean of five
normal values covers 0.850 in simulation, against 0.954 for the t-interval.
The other contrast tables of Appendix~\ref{app:robust} give the same 95\% $t$-interval (df 4) next to each rule
output; where the reading scripts did not print it, a post hoc script recomputes it from their saved per-unit
values and first reproduces every interval they did print.

\begin{table}[!htbp]
\caption{Rule output, 95\% intervals and TOST ($\pm3$, t) for \lastarm$-$\lparm{} and D17 and (bottom block,
rotations) for \lastr$-$\lastarm{} and \lparm$-$\lpr, linear head (letters:
Table~\ref{tab:vocab}). Rotations: intervals over the five rotation means. DL's split: intervals over the five seeds
(seed uncertainty on one split); D24-B registers a rule only for \lastarm$-$\lparm{} (``--'': descriptive). Equivalence
within $\pm3$ only where TOST $p<0.05$.}
\label{tab:intervals}
\centering\footnotesize
\setlength{\tabcolsep}{3pt}
\begin{tabular}{llllcccc}
\toprule
setting & model & contrast & rule & t 95\% & NB 95\% & bootstrap & TOST $p$\\
\midrule
CHB-MIT & BIOT & \lastarm$-$\lparm{} (+1.9) & N & [$-$1.0, +4.8] & [$-$2.8, +6.6] & [+0.1, +3.9] & 0.175\\
CHB-MIT & BIOT & D17 (+0.1) & R & [$-$0.1, +0.3] & [$-$0.3, +0.5] & [$-$0.1, +0.4] & 0.000\\
CHB-MIT & CBraMod & \lastarm$-$\lparm{} ($-$0.0) & N & [$-$0.4, +0.4] & [$-$0.7, +0.7] & [$-$0.3, +0.3] & 0.000\\
CHB-MIT & CBraMod & D17 (+1.0) & R & [$-$0.9, +3.0] & [$-$2.1, +4.2] & [$-$0.5, +2.6] & 0.024\\
Siena & BIOT & \lastarm$-$\lparm{} (+9.2) & F & [+2.3, +16.2] & [$-$2.3, +20.7] & [+4.6, +15.7] & 0.966\\
Siena & BIOT & D17 (+0.3) & R & [$-$0.6, +1.3] & [$-$1.2, +1.9] & [$-$0.4, +1.0] & 0.001\\
Siena & CBraMod & \lastarm$-$\lparm{} (+0.3) & N & [$-$0.4, +0.9] & [$-$0.8, +1.4] & [$-$0.6, +0.8] & 0.000\\
Siena & CBraMod & D17 ($-$0.4) & R & [$-$3.1, +2.3] & [$-$4.9, +4.1] & [$-$2.3, +1.7] & 0.028\\
Siena filt. & BIOT & \lastarm$-$\lparm{} (+4.1) & F & [+0.1, +8.2] & [$-$2.6, +10.8] & [+1.4, +6.9] & 0.755\\
Siena filt. & BIOT & D17 (+0.4) & R & [$-$0.1, +0.9] & [$-$0.5, +1.3] & [$-$0.1, +0.9] & 0.000\\
Siena filt. & CBraMod & \lastarm$-$\lparm{} (+0.1) & N & [$-$0.0, +0.3] & [$-$0.2, +0.5] & [+0.0, +0.3] & 0.000\\
Siena filt. & CBraMod & D17 ($-$1.1) & R & [$-$3.7, +1.5] & [$-$5.5, +3.2] & [$-$3.9, +0.7] & 0.058\\
DL's split & BIOT & \lastarm$-$\lparm{} ($-$0.6) & N & [$-$3.8, +2.6] & n/a & [$-$2.6, +1.4] & 0.053\\
DL's split & BIOT & D17 (+0.0) & -- & [$-$0.1, +0.1] & n/a & [$-$0.0, +0.1] & 0.000\\
DL's split & CBraMod & \lastarm$-$\lparm{} (+0.0) & N & [$-$0.0, +0.0] & n/a & [$-$0.0, +0.0] & 0.000\\
DL's split & CBraMod & D17 ($-$2.1) & -- & [$-$4.1, $-$0.1] & n/a & [$-$3.4, $-$0.9] & 0.130\\
\midrule
CHB-MIT & BIOT & \lastr$-$\lastarm{} (+0.2) & N & [$-$16.8, +17.1] & [$-$27.6, +27.9] & [$-$8.9, +12.4] & 0.332\\
CHB-MIT & BIOT & \lparm$-$\lpr{} ($-$2.0) & N & [$-$17.0, +13.1] & [$-$26.6, +22.7] & [$-$12.9, +6.1] & 0.428\\
CHB-MIT & CBraMod & \lastr$-$\lastarm{} ($-$4.4) & S & [$-$7.4, $-$1.4] & [$-$9.3, +0.5] & [$-$6.9, $-$2.2] & 0.871\\
CHB-MIT & CBraMod & \lparm$-$\lpr{} (+5.5) & F & [+2.5, +8.4] & [+0.7, +10.3] & [+3.5, +7.7] & 0.961\\
Siena & BIOT & \lastr$-$\lastarm{} (+7.2) & N & [$-$11.5, +25.9] & [$-$23.8, +38.2] & [$-$4.4, +19.6] & 0.716\\
Siena & BIOT & \lparm$-$\lpr{} ($-$16.1) & S & [$-$40.7, +8.6] & [$-$56.8, +24.7] & [$-$33.2, $-$1.7] & 0.893\\
Siena & CBraMod & \lastr$-$\lastarm{} ($-$7.6) & S & [$-$14.3, $-$0.9] & [$-$18.7, +3.4] & [$-$12.8, $-$3.0] & 0.937\\
Siena & CBraMod & \lparm$-$\lpr{} (+7.0) & F & [$-$2.4, +16.3] & [$-$8.5, +22.4] & [+0.9, +13.7] & 0.849\\
Siena filt. & BIOT & \lastr$-$\lastarm{} ($-$4.7) & S & [$-$34.5, +25.1] & [$-$54.0, +44.6] & [$-$19.2, +17.0] & 0.559\\
Siena filt. & BIOT & \lparm$-$\lpr{} (+1.0) & N & [$-$32.6, +34.6] & [$-$54.6, +56.5] & [$-$23.8, +17.4] & 0.438\\
Siena filt. & CBraMod & \lastr$-$\lastarm{} ($-$10.7) & S & [$-$17.0, $-$4.4] & [$-$21.1, $-$0.2] & [$-$15.7, $-$6.2] & 0.986\\
Siena filt. & CBraMod & \lparm$-$\lpr{} (+9.4) & F & [+3.6, +15.3] & [$-$0.2, +19.1] & [+5.4, +13.7] & 0.981\\
\bottomrule
\end{tabular}
\end{table}
Against DL's gains (H$_0$: effect $\ge$ +36.2 for BIOT, +29.5 for CBraMod; one-sided t), $p=2.5\times
10^{-6}$ for BIOT on CHB-MIT and $p=2.9\times10^{-6}$ on DL's split (linear head).

\paragraph{Power and false positives of the rule.} Simulating five units with each contrast's own SD: for
CHB-MIT CBraMod \lastarm$-$\lparm{} (SD 0.3) the rule has false-positive rate 0.000 and 80\% power at an effect
of 3.1; for CHB-MIT BIOT (SD 2.3) 0.004 and 3.9; for Siena BIOT (SD 5.6) 0.175 and 5.9. For the high-variance
BIOT contrasts \lastr$-$\lastarm{} and \lparm$-$\lpr{} (SD 12.1 to 27.1) the false-positive rate is 0.313 to
0.352 and the 80\%-power effect 11.8 to 26.2 (normal errors), so these BIOT classifications carry little
information either way.

\paragraph{Multiplicity.} With five units the smallest exact two-sided sign-flip $p$ is 2/32 = 0.0625, so no
contrast can reach 0.05, before or after Holm correction within the families C1, Issue~1 (\lastr$-$\lastarm), features and
D17.

\paragraph{Variance components.} Rotations carry most of the variance of the arm means (rotation share 0.56 to 1.00
on CHB-MIT); seed variation is small for pretrained arms (BIOT \lparm{} SD 0.86) and large for random-init BIOT
(SD 8.71 to 9.04). At subject level, in a nested decomposition of the per-subject ROC-AUC (subjects within rotations,
ANOVA-type moments), the SD between subjects (12 to 25 ROC points for most arms) exceeds the rotation
component that remains once the subject variance it contains is removed. These SDs describe heterogeneity;
the rules use the pooled-fold ROC-AUC, not per-subject ROC-AUC.

\FloatBarrier
\subsection{Analysis timing and deviations}
\label{app:hygiene}
\paragraph{CBraMod without the reused runs.} D24-A reuses CBraMod's $S_r$ run from the D23 re-run, which had
been read before D24 was written. Reading D24-A CBraMod with only the two new seeds leaves every
classification unchanged in all three settings (12 of 12 rule outputs).
\paragraph{Deviations from the logged protocols.} The protocols below are internal, timestamped
specifications kept during the project, not a public preregistration. This appendix is the one place that
gives the full timing evidence, with dates and commits; the experiment sections carry a short post hoc tag and
refer here.
\begin{itemize}
  \item D18 logged the paired rule as the decision rule (for example ``last-layer SA helps the pretrained FM'' if
  $\lastarm-\lparm>3$ in mean and in at least 4 of 5 rotations), and D24 kept it. After all data and the external
  review we report its outputs as a descriptive consistency criterion and base conclusions on the 95\%
  $t$-intervals (post hoc; Section~\ref{sec:rules}).
  \item The first PR-AUC reading of D29 reused the ROC rule outputs and their 3-point margin. That reuse was
  withdrawn after an external review (D30, post hoc, 2026-11-05): the PR-AUC comparisons are reported as
  effects with intervals and no practical-importance classification, and no PR-AUC margin was defined in its
  place (Appendix~\ref{app:d29}).
  \item Two reading rules came after data: the reader choices that the E30 rule leaves open were fixed after
  its raw records had been read but before any E30 aggregate, and the E5 reader fix (a configuration that
  fails the screen counts as not reaching) was made after the screen results.
  \item D24-A registered ``also mean $\pm$ SEM over all 15 runs''; we report the SEM over the 5 rotation means
  because seeds within a rotation share a split. The 15-run SEMs are printed by the reading script (for
  example, CHB-MIT BIOT \lparm{} $65.1\pm4.2$ over runs against $65.1\pm7.9$ over rotations).
  \item D18 registered one seed per rotation; D24 raised it to three after every one-seed result had been
  read; three classifications changed (Appendix~\ref{app:earlier}).
  \item D24-D's bootstrap is two-stage (rotations, then seeds), a refinement fixed after the D24 review; it
  under-covers with five rotations.
  \item The D24 \lparm$-$\lpr{} rule was fixed after the D23 re-run had shown CBraMod \lparm{} above \lpr{} in 14
  of 15 rotation-settings (tag P*).
  \item BIOT moved from train-mode to eval-mode probing in D24 after all train-mode BIOT results had been seen
  (registered in D24 before its runs).
  \item The D12 rule's reference value (88.5) was seen before the rule was written; D12 jobs were logged as
  submitted at about 01:05 UTC (2026-10-16) and the rule at 01:10 UTC, before any result (P*). D13 likewise precedes every
  result but not the submission.
  \item The completeness audit adopted by D25 quotes statistics computed in memory before D25; E10 and E11 cite
  script output only and are post hoc.
  \item The decision-log entries up to D27 were written without version control; timing rests on file times, logged submission times
  and result-object times; E22 checked the entries of Appendix~\ref{app:timeline} from D4 to D25 except D14 and D16;
  the D25 released-code reading choices and every later entry up to D27 rest on the decision log and our unreleased work log alone
  (D26 and D27 also on the creation times of their first jobs). The D28 and D29 protocols, both post hoc (designed after
  all earlier data and an external review), were committed to a local version-controlled repository before any of
  their runs, smoke tests included: protocols, searches and rules in commit \texttt{c0283f4} (2026-11-04 23:17
  UTC) and code and run specifications in commit \texttt{55b580d} (2026-11-05 00:51 UTC). Their
  rule outputs are pre-specified for their own data, but both are post hoc checks of existing analyses, not
  independent confirmatory studies: the motivation is retrospective, while each protocol was frozen before its
  own runs.
  \item D31 and D32 were additional analyses motivated by D28 and E29. Their protocols, including
  designation of D31 as the primary tuned analysis, were frozen in commit \texttt{222d090} before their runs;
  expanded-grid code and specifications are in \texttt{d268b37}. D31 replaces the narrower search in the
  primary results regardless of outcome. D32 is reported separately. Neither is an independent confirmatory
  study or a public preregistration.
  \item The converged linear-probe analysis was motivated by the expanded search and is post hoc.
  Its initial protocol was committed as \texttt{cb531a6}; pre-run amendments
  \texttt{145113b}, \texttt{d45a705} and \texttt{bb16e84} corrected the model-class description,
  diagnostics and objective normalization before execution. Implementation is in \texttt{cf20ede}.
  The analysis is separate from the primary 18-comparison family and is not an external replication
  or public preregistration.
  \item D28's Q3 asks whether the pretrained-against-random contrasts ``keep their direction'' without defining
  direction; the reader choice we made is stated once, with the reading itself, in Appendix~\ref{app:d28}. The D28
  outlier list, the count of selected epochs $\le2$ and the per-model grouping of Q3 were added after
  the first output and are descriptive.
  \item D27-A registers D19 for each head separately; calling A1 head-dependent borrows the both-heads rule of
  D27-B and is our descriptive reading. Read literally, the D27 addition registers A7; our reading script had
  treated it as descriptive (same outcome).
  \item The released-code reader was fixed once after the E5 follow-up results had been seen (five-seed results
  were NumPy booleans and failed its identity test for False); no rule or per-configuration result changed.
  \item The clock times first written in the decision log for D26 and D27 were estimates and too late; the
  timeline uses the creation times of their first jobs, which both entries precede.
  \item Records do not store the data variant: S3FT is unfiltered Siena despite its prefix (identified by its
  evaluation hash); the \texttt{ft} records store a label-smoothing value (0.1) that is not applied to binary
  tasks; and the documented one-seed E26 reading command no longer runs now that the cells hold three seeds (the
  D27-B script re-derives the E26 strings, 24 of 24).
  \item Smaller items: QC exclusions cannot be read under the five-rotation rule on Siena; the SSL dose sweep
  (E29) used the raw head only (the D27 noise controls add the standardized head at one seed); 5 of 296 early score files have 0.1--1.8\% of test windows tied (D20);
  evaluation inputs are FFT-resampled from float32 (relative difference about $10^{-7}$), and adaptation
  uses activation checkpointing (gradients identical within $10^{-6}$).
\end{itemize}

\section{Per-rotation results and DL's split}
\label{app:perrot}
\begin{table}[!htbp]
\caption{DL's split with the start/end labels and the encoder-level probe (D24-B; E26 for \texttt{linear\_z}). ROC-AUC
$\times100$, mean~$\pm$~SEM over 5 seeds (42, 123, 2021, 3407, 8888). SA gain = \lastarm$-$\lparm, paired
by seed, with its 95\% $t$-interval over seeds (seed uncertainty on one split, not test-set sampling; $^\dagger$
bootstrap, post hoc). The consistency criterion (D24-B) needs a mean above 3 and 4 of 5 seeds positive; every row
shows no consistent difference (N). \texttt{linear\_z} = linear head on features standardized with training-split
statistics (the standardized probe).}
\label{tab:p1e}
\centering\footnotesize
\setlength{\tabcolsep}{3pt}
\begin{tabular}{llccccll}
\toprule
model & head & \lparm{} & \lastarm{} & \lpr{} & \lastr{} & SA gain [95\% CI] & DL LP / SA\\
\midrule
CBraMod & linear & 92.00 $\pm$ 0.04 & 92.02 $\pm$ 0.05 & 85.6 $\pm$ 0.7 & 83.5 $\pm$ 0.3 & 0.01 $\pm$ 0.01 [$-$0.0, +0.0] & 54.14 / 83.67\\
CBraMod & mlp\_fullft & 91.7 $\pm$ 0.3 & 91.7 $\pm$ 0.2 & 89.5 $\pm$ 0.4 & 89.7 $\pm$ 0.9 & 0.0 $\pm$ 0.1 [$-$0.4, +0.4] & \\
CBraMod & linear\_z & 88.5 $\pm$ 0.4 & 88.6 $\pm$ 0.4 & 85.1 $\pm$ 1.2 & 85.0 $\pm$ 1.2 & 0.2 $\pm$ 0.1 [+0.0, +0.3]$^\dagger$ & \\
BIOT & linear & 55.0 $\pm$ 0.4 & 54.4 $\pm$ 1.1 & 90.1 $\pm$ 0.2 & 90.2 $\pm$ 0.2 & $-$0.6 $\pm$ 1.2 [$-$3.8, +2.6] & 46.93 / 83.08\\
BIOT & mlp\_fullft & 56.1 $\pm$ 0.1 & 54.9 $\pm$ 1.2 & 88.5 $\pm$ 0.2 & 88.5 $\pm$ 0.2 & $-$1.2 $\pm$ 1.2 [$-$4.6, +2.2] & \\
BIOT & linear\_z & 79.1 $\pm$ 1.8 & 79.3 $\pm$ 1.9 & 88.4 $\pm$ 0.4 & 88.4 $\pm$ 0.5 & 0.2 $\pm$ 0.8 [$-$0.8, +1.8]$^\dagger$ & \\
\bottomrule
\end{tabular}
\end{table}

The per-rotation results of the revised protocol --- per-rotation means over the 3 D24 seeds and the
mean~$\pm$~SEM over the 5 rotation means, for all five arms of both models in all three settings --- are
Table~S26, the per-rotation paired differences behind each rule output are included with it, and Figure~S3
plots every rotation. The contrasts those tables produce, with their intervals and rule outputs, are in
Table~\ref{tab:d24a}; the \allarm$-$\lastarm{} capacity reference (no rule
output) is +9.8 / +1.6 on CHB-MIT, +10.6 / +2.0 on Siena and +0.4 / $-0.8$ on filtered Siena (BIOT / CBraMod).

\FloatBarrier
\section{The replication attempt with the released-code probe}
\label{app:replication}
These runs use the released-code probe defined in Section~\ref{sec:probes}.

\begin{table}[!htbp]
\caption{Replication runs, DL's split, protocol of Section~\ref{sec:setup}, start/end labels and the evaluation windows of the shared preprocessing (\texttt{process2.py}), unfiltered; mean~$\pm$~SEM over
5 seeds. bal / PR / ROC $\times100$.}
\label{tab:wave1}
\centering\footnotesize
\setlength{\tabcolsep}{3pt}
\begin{tabular}{lllcccl}
\toprule
model & arm & head & bal & PR & ROC & DL bal / PR / ROC\\
\midrule
BIOT & lp & mlp\_fullft & 52.19 $\pm$ 0.07 & 2.42 $\pm$ 0.07 & 52.44 $\pm$ 0.16 & 50.74 / 1.53 / 46.93\\
BIOT & lp & linear & 50.46 $\pm$ 0.19 & 1.79 $\pm$ 0.04 & 52.05 $\pm$ 0.23 & \\
BIOT & last & mlp\_fullft & 50.21 $\pm$ 0.17 & 1.71 $\pm$ 0.09 & 51.99 $\pm$ 1.30 & 52.71 / 31.52 / 83.08\\
BIOT & last & linear & 49.88 $\pm$ 0.08 & 1.60 $\pm$ 0.06 & 52.55 $\pm$ 1.13 & \\
BIOT & last\_randinit & mlp\_fullft & 60.82 $\pm$ 0.91 & 36.86 $\pm$ 0.90 & 88.48 $\pm$ 0.19 & \\
BIOT & last\_randinit & linear & 58.30 $\pm$ 0.78 & 37.74 $\pm$ 0.36 & 90.15 $\pm$ 0.15 & \\
CBraMod & lp & mlp\_fullft & 56.62 $\pm$ 0.33 & 34.82 $\pm$ 0.54 & 85.32 $\pm$ 0.68 & 57.87 / 17.35 / 54.14\\
CBraMod & lp & linear & 57.56 $\pm$ 0.35 & 18.68 $\pm$ 0.37 & 65.36 $\pm$ 0.52 & \\
CBraMod & last & mlp\_fullft & 56.22 $\pm$ 0.70 & 35.72 $\pm$ 0.55 & 86.53 $\pm$ 1.05 & 61.37 / 24.62 / 83.67\\
CBraMod & last & linear & 58.23 $\pm$ 0.17 & 23.15 $\pm$ 0.43 & 74.04 $\pm$ 0.33 & \\
CBraMod & last\_randinit & mlp\_fullft & 54.64 $\pm$ 2.47 & 36.67 $\pm$ 2.35 & 89.45 $\pm$ 0.93 & \\
CBraMod & last\_randinit & linear & 56.23 $\pm$ 1.83 & 26.96 $\pm$ 0.71 & 86.83 $\pm$ 0.48 & \\
\bottomrule
\end{tabular}
\end{table}
With the released-code probe, DL's CBraMod SA cell (bal / PR / ROC 61.37 / 24.62 / 83.67) is matched by random
init with the linear head at batch 128 (59.73~$\pm$~1.57 / 23.86~$\pm$~2.17 / 85.32~$\pm$~0.59; ``consistent''
under D13; Section~\ref{sec:a1}).
\paragraph{Screen (D12, 2 seeds, $\pm$ = half-range; rules before any screen result).} ``Random init explains BIOT SA'':
supported (random-init last / last2 / all 88.48 / 88.51 / 88.49; pretrained \lastarm{} on the full 81,993-window
pool 55.25 $<$ 70). ``The data budget explains it'': not supported for BIOT (55.25 $<$ 78); for CBraMod the full
pool lifts the linear-head \lastarm{} from 74.0 to 86.7. Batch 128 moves BIOT \lparm{} and \lastarm{} by +6.6 and
+6.8 (to 59.0 and 58.8): toward DL for \lastarm, away for \lparm, nothing near 83. The DL-described
(filtered) evaluation moves BIOT \lparm{} 52.4 to 61.5 and \lastarm{} 52.0 to 58.6; no configuration setting
reproduces the CBraMod LP with either head.
\paragraph{Confirmation (D13, 5 seeds, linear head, batch 128, \lastr{} as the SA arm).} Gaps in DL-SEM
units (bal / PR / ROC), unfiltered (shared) evaluation: BIOT LP 0.2 / 2.7 / 11.6 (partial), BIOT SA (random) 5.0 / 6.2 / 3.6
(not), CBraMod LP 0.3 / 2.6 / 10.6 (partial), CBraMod SA (random) 0.5 / 0.3 / 1.2 (consistent); filtered
evaluation: 0.0 / 2.4 / 19.4, 5.2 / 4.0 / 3.3, 0.2 / 0.8 / 4.1, 1.2 / 0.6 / 2.4. Pretrained BIOT \lastarm{}: 12.4 and
9.8 SEM on ROC. In ROC, random-init BIOT SA gave 89.93~$\pm$~0.29 and 89.24~$\pm$~0.28 under the two
evaluation variants (DL: 83.08), and pretrained BIOT \lastarm{} 59.70~$\pm$~0.88 and 64.48~$\pm$~0.60. Verdict:
not consistent with DL Table~1 as a whole.

\FloatBarrier
\section{Analysis provenance}
\label{app:timeline}
Table~S27 records the timing of analysis decisions relative to the data they use, including the
version-controlled protocols for the probe searches, larger-dose reassessment and converged linear probes.
The decision log and saved timing audit accompany that table. Appendix~\ref{app:hygiene} distinguishes
choices made before their own runs from post hoc motivation and documents protocol deviations. A
label-free diagnostic of where to adapt was specified but not run; early-layer adaptation is retained
only as an exploratory result in Appendix~\ref{app:earlier}.

\label{sec:contact}
We first contacted the original authors around 10 September 2026 and followed up on 28 October 2026
with questions 1--6 of Appendix~\ref{app:questions}, including the setting of
\texttt{foundation\_dir} for DL Tables~1 and~2. Questions 7 and 8 were added later.
No response had been received as of 16 November 2026.

\FloatBarrier
\section{Unverified experimental settings}
\label{app:questions}
\begin{enumerate}
  \item Was \texttt{cfg.model.foundation\_dir} set, for example on the command line, for DL Tables~1 and~2?
  If not, the SSL step starts from a randomly initialised encoder whose weights then replace the pretrained
  backbone (Section~\ref{sec:a1}).
  \item Which value of \texttt{spt\_anchor\_l2sp\_lambda} was used, given that no shipped configuration
  defines it?
  \item For the CBraMod LP rows of DL Table~1, was \texttt{backbone.proj\_out} replaced by
  \texttt{nn.Identity()}, and was the frozen backbone in train or eval mode while the probe trained? (For
  the SA rows, the released \texttt{step\_up} fails with \texttt{nn.Identity()} unless the state-dict load
  is also changed; Section~\ref{sec:a2}.)
  \item Was \texttt{training.full\_ft} False (a single linear layer) for DL Table~1? The shipped configuration
  sets it True.
  \item Which batch size produced DL Table~1: 128, as in DL's text, or 256, as in
  \relcode{src/conf/config.yaml\#L16}{src/conf/config.yaml:16}?
  \item Which recordings or filter produced the 4,174 CHB-MIT SSL windows, given that the documented pipeline
  yields 81,993, the text mentions $N$ = 86K at 40\% of the data, and the data-amount configuration sets
  \texttt{compute\_steps: 2706}
  (\relcode{src/ablation_diversityN/conf_diversity/dataset/chbmit.yaml\#L14}{chbmit.yaml:14})?
  \item The README states that the five reported seeds are 42, 123, 2021, 3407 and 8888; which of them produced
  the one-seed DL Table~2?
  \item Did DL Table~2 use the same adaptation and probe protocol as DL Table~1? Its BIOT last-layer
  ROC-AUC (88.67) is a single run, while DL Table~1's BIOT SA (83.08) is a mean over 5 seeds; the released
  default SA configuration gives per-seed ROC-AUC from 78.92 to 88.67 (E4; Appendix~\ref{app:e5}).
\end{enumerate}

\FloatBarrier
\section{Compute and cost}
\label{app:compute}
Every job ran on one NVIDIA A10G (24~GB) unless stated, and costs are cloud list prices per instance hour
(ml.g5.8xlarge \$3.06, ml.g5.16xlarge \$5.12, ml.g6e.2xlarge \$2.80). The total project cost is about \$4,856. The initial probe search and
fixed-compute controls cost \$289.68 across 71 jobs (94.67 GPU hours); the expanded probe search cost
\$225.65, the larger-dose reassessment \$16.70, and the converged linear-probe analysis \$7.05
across six production jobs (2.30 GPU hours). Smoke tests and earlier implementation audits are included
in the project total and itemized in Table~S28. Released-code runs take
longer on CHB-MIT because the released data loaders resample in the main process (no worker processes are
configured). CPU analyses (E1, E2, E11--E15, E17-free to E20-free) ran on a separate CPU machine.
Table~S28 gives job counts, GPU hours and costs by analysis, with the complete run log in the supplement.

\FloatBarrier
\section{Reproducibility}
\label{app:checklist}
\label{app:repro}
The anonymized supplement contains our MIT-licensed code and tests, result records, window indices,
analysis scripts and saved outputs, all supplementary tables and figures, checkpoint hashes, and copies
of the decision, experiment and run logs with identifying information redacted. Experimental decisions,
rules, dates and numerical results are preserved. The released GPL-3.0 code is not redistributed;
our harness clones the audited commit \texttt{d056cdf}. Full instructions and commands are in
\texttt{docs/REPRODUCE.md} and the archive README.

\paragraph{Verification from saved records (CPU).} \texttt{make\_main\_tables.py} checks the original
record-derived tables. The primary tuned readings are regenerated with
\texttt{scripts/analysis/d28\_read.py --d31}, followed by \texttt{tuned\_metrics.py} for expanded-search precision--recall
and larger-dose ROC/PR results; \texttt{d29\_read.py} reads the fixed-compute controls. Restore the window indices
using the archive README before running readers that compute train/test ratios. The original-search sensitivity results and
complete configurations remain available. \texttt{d33\_read.py} regenerates the converged-probe table
from the supplied manifest and records.

\paragraph{Retraining and omitted inputs.} Raw EDFs are public on PhysioNet (CHB-MIT 1.0.0, ODC-By;
Siena 1.0.0, CC BY 4.0). Window stores, checkpoints and most per-window scores are excluded for size or
third-party distribution; preprocessing and recorded commands regenerate them. Score-level analyses
therefore require those regenerated inputs. \texttt{reproduce\_headline.sh} supplies a local-GPU route;
cloud submission needs the excluded private wrapper or a replacement integration. The job specifications
and per-record commands identify all settings, including both probe searches and the converged-probe analysis.

\paragraph{Environment and determinism.} Our pipeline uses Python 3.12 and PyTorch 2.6.0+cu126; released-code
runs use torch 2.4.1+cu121 and the documented additional dependencies. Each job used one A10G (24 GB),
except five released-code memory reruns on L40S and nine host-memory reruns with the same A10G on a larger
instance. Existing 484/484 rerun pairs agree within $10^{-6}$, not bitwise; the released code is not
reproducible at a fixed seed. Exact seeds, split hashes, checkpoint hashes and hardware are in the records
and README.

\section{Additional figures}
\label{app:figs}

\begin{figure}[!htbp]
\centering
\includegraphics[width=\linewidth]{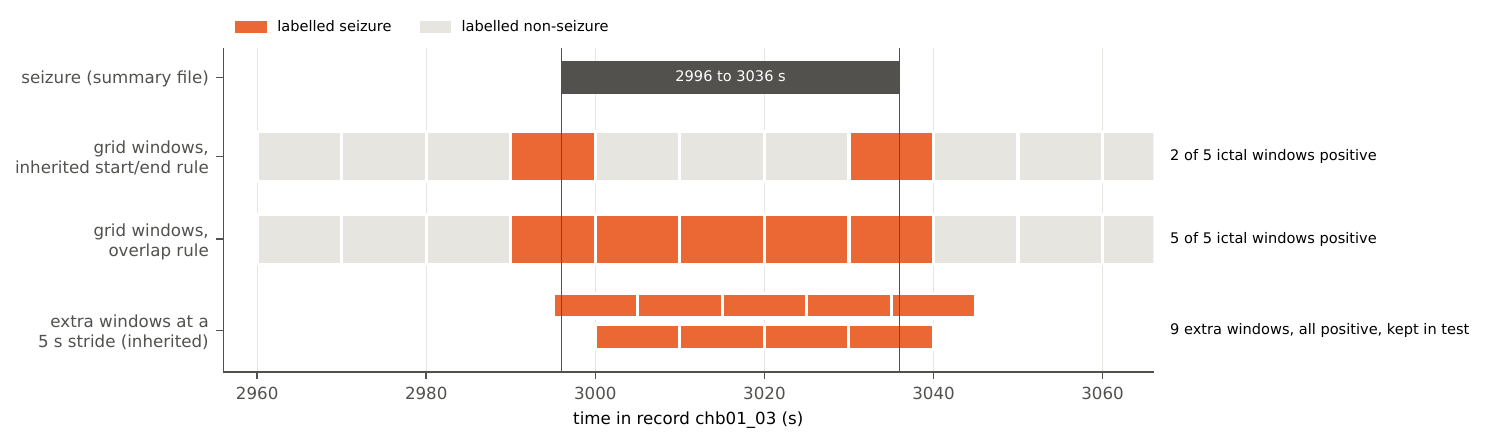}
\caption{The CHB-MIT protocol on record chb01\_03 (seizure 2996 to 3036~s): 10~s grid windows labelled by
the inherited start/end rule (\texttt{process2.py}; positive only if a seizure start or end falls inside the
window; the 40~s seizure spans five grid windows and two are positive) and by the overlap rule, and the extra 10~s windows at a 5~s stride that the shared preprocessing adds around every seizure
in every split, test included.}
\label{fig:timeline}
\end{figure}

Two further figures are in the supplementary document: Figure~S2 sets DL's four CHB-MIT cells next to our runs
on DL's split (both heads, released-code probe and encoder-level probe, pretrained and random init), and
Figure~S3 plots every D24-A contrast per rotation (both models, all three settings). The numbers they plot are
in Tables~\ref{tab:repro-summary}, \ref{tab:p1e} and~\ref{tab:d24a}.

\begin{figure}[!htbp]
\centering
\includegraphics[width=\linewidth]{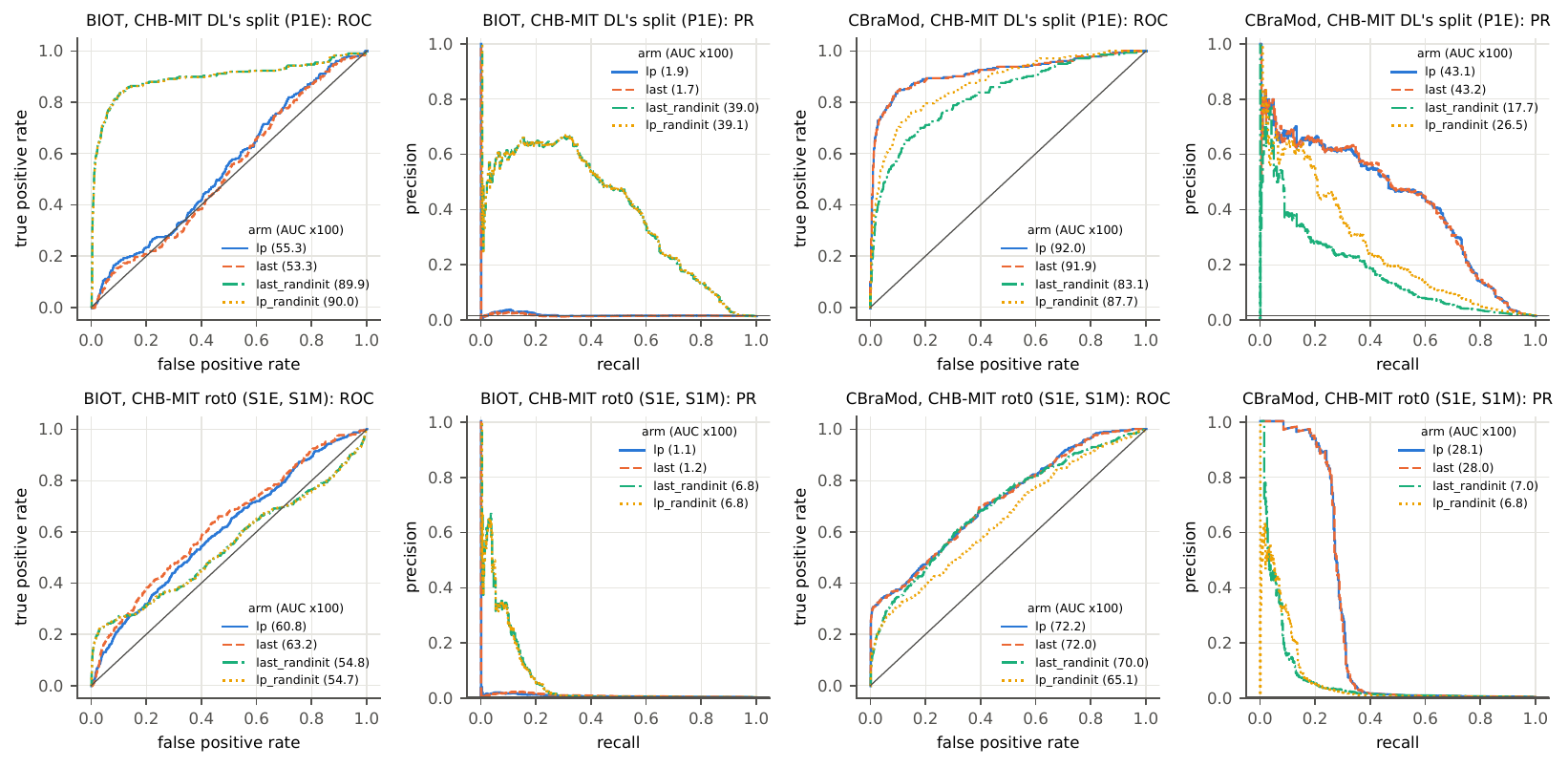}
\caption{Test ROC and precision--recall curves at the validation-selected epoch (seed 42, linear head,
encoder-level probe) for \lparm, \lastarm, \lastr{} and \lpr, on DL's split (P1E) and on CHB-MIT rotation 0 of the revised
protocol. Legend values are ROC-AUC or PR-AUC $\times100$; the grey line in the PR panels is the test
prevalence.}
\label{fig:rocpr}
\end{figure}

\begin{figure}[!htbp]
\centering
\includegraphics[width=0.8\linewidth]{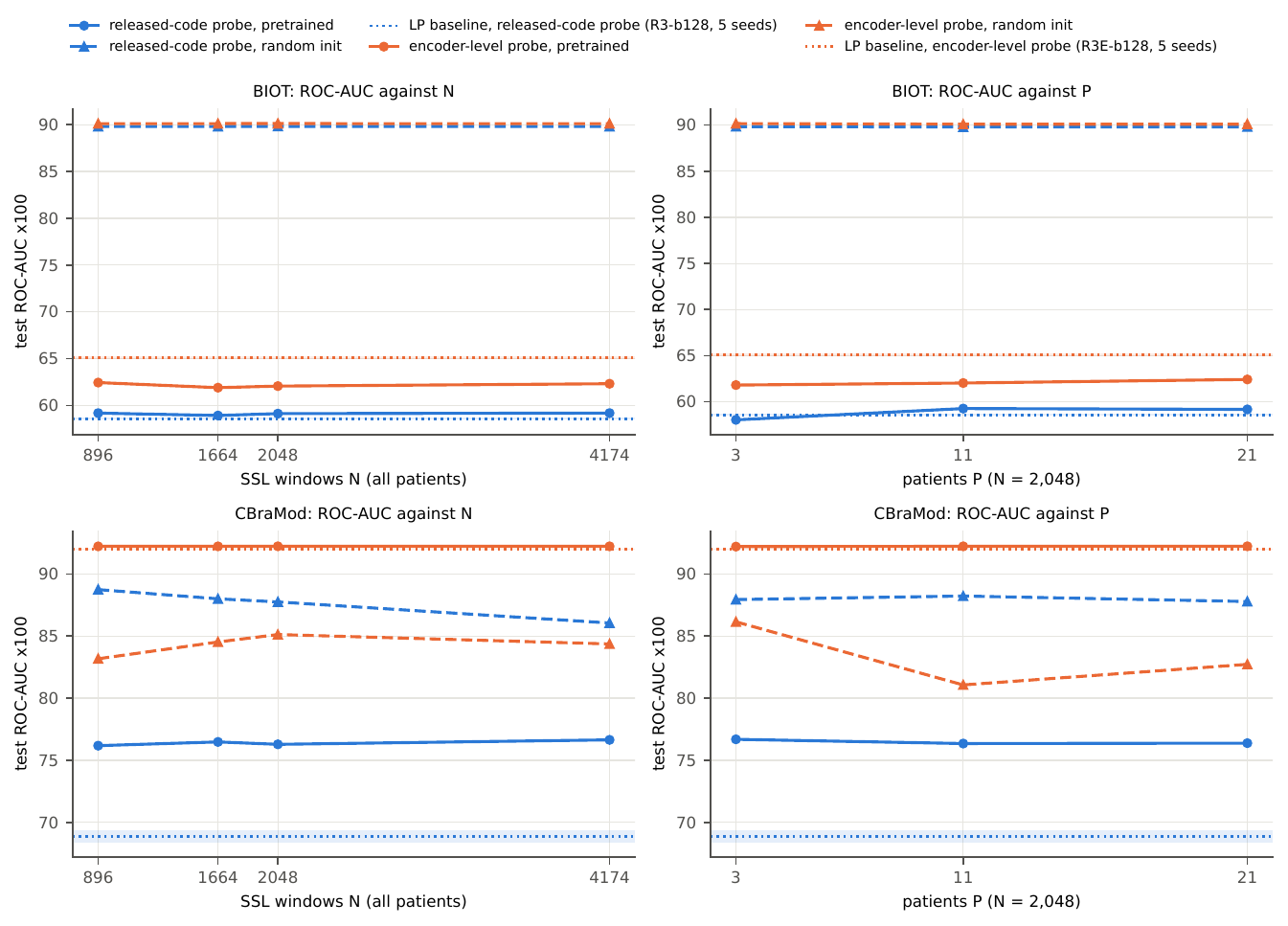}
\caption{Data-amount (C2, ROC-AUC against the number of SSL windows $N$) and patient-count (C3, against the
number of patients $P$ at $N=2{,}048$) tests on DL's split, linear head at batch 128, fixed-compute SSL,
\emph{one seed per point} (seed 42, the D24-C runs; for BIOT seed 42 is the lowest of the five seeds, and the
five-seed BIOT curve is in Table~\ref{tab:c2c3}): released-code probe (C2-*, C3-*) against the encoder-level probe
(C2E-*, C3E-*), pretrained and random init, with the LP baselines (R3-b128, R3E-b128; mean, band = SEM over
seeds) as dotted lines.}
\label{fig:c2c3}
\end{figure}

The SSL adaptation-loss figures are Figures~S4 (per pass) and~S5 (per optimizer step): the loss of the adapted
block from the pretrained checkpoint and from random init, with the restored pass marked, median and min--max
band over seeds and rotations, for both models on DL's split and on the five rotations of CHB-MIT, Siena and
filtered Siena.

\begin{figure}[!htbp]
\centering
\includegraphics[width=\linewidth]{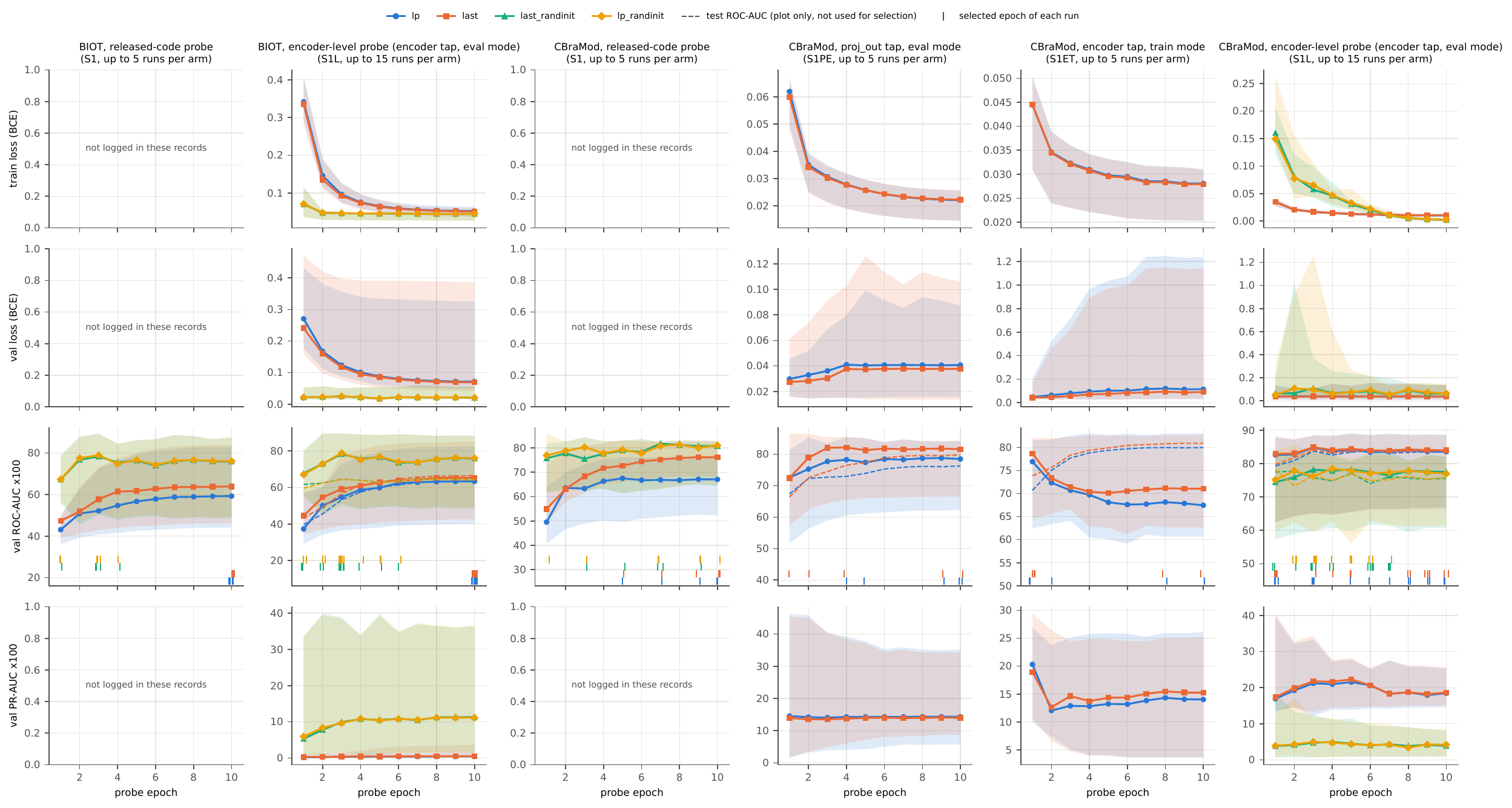}
\caption{Linear-probe learning curves per epoch on CHB-MIT, five rotations: train loss, validation loss,
validation ROC-AUC and validation PR-AUC; median line and band over seeds and rotations per arm. Ticks at the
bottom of the ROC panels mark each run's selected epoch (best validation ROC-AUC); the dashed test ROC-AUC is
plot only. With the encoder-level probe, pretrained BIOT (\lparm, \lastarm) is still improving at epoch 10 and its
runs mostly select the last epoch: the raw head is under-trained (its best epoch is its last in 58 of 90
fits, E15). The released-code-probe runs (first and
third columns) logged validation ROC-AUC only. Columns: BIOT released-code probe (S1); BIOT encoder-level probe (S1L);
CBraMod released-code probe (S1); CBraMod \texttt{proj\_out} tap, eval mode (S1PE); CBraMod encoder tap,
train mode (S1ET); CBraMod encoder-level probe (S1L). The same curves on DL's split and on Siena are in \ours.}
\label{fig:curvesrot}
\end{figure}

\end{document}